\documentclass{article}
\usepackage{float}

\usepackage{arxiv}

\usepackage{float}

\usepackage[table,dvipsnames]{xcolor}

\usepackage{geometry}
\usepackage{booktabs}
\usepackage{multirow}
\usepackage{graphicx}
\usepackage{caption}
\usepackage{subcaption}
\usepackage{threeparttable}
\usepackage{longtable}
\usepackage{makecell}
\usepackage{microtype}
\usepackage{float}
\usepackage[authoryear,round]{natbib}
\usepackage[utf8]{inputenc} % allow utf-8 input
\usepackage[T1]{fontenc}    % use 8-bit T1 fonts
\usepackage{hyperref}       % hyperlinks
\usepackage{url}            % simple URL typesetting
\usepackage{booktabs}       % professional-quality tables
\usepackage{amsmath}        % math notation
\usepackage{amsfonts}       % blackboard math symbols
\usepackage{bm}             % bold math symbols
\usepackage{enumitem}       % customized lists
\usepackage{textgreek}     % allows direct use of Greek letters in text
\usepackage{nicefrac}       % compact symbols for 1/2, etc.
\usepackage{microtype}      % microtypography
\usepackage{lipsum}		    % Can be removed after putting your text content
\usepackage{graphicx}
\usepackage{doi}
\usepackage{xcolor}
\usepackage{placeins}
\usepackage{siunitx}
\usepackage{booktabs}
\usepackage{multirow}
\usepackage{makecell}

\usepackage{booktabs, colortbl, xcolor, multirow, array, makecell, caption}
\definecolor{Drose}{RGB}{210,100,100}
\definecolor{Lrose}{RGB}{250,200,195}
\definecolor{Ngray}{RGB}{232,232,232}
\definecolor{Lsage}{RGB}{195,225,195}
\definecolor{Dsage}{RGB}{100,170,110}

\usepackage{amsmath, amssymb} % For mathematical symbols and environments
\usepackage{graphicx}        % For including graphics (if needed)
\usepackage{enumitem}        % For customized lists
\usepackage{bm}         
\usepackage{graphicx} % Required for \includegraphics if used elsewhere
\usepackage{geometry} 
\usepackage{appendix}% Example: Adjust page margins if needed
\usepackage{etoolbox}
\usepackage{graphicx}
\usepackage{multirow}
\usepackage{microtype}
\usepackage{graphicx}
\usepackage{float}
\begin{document}
%--------------------------------------- Header ------------------------------------
\title{Geopolitics, Geoeconomics, and Sovereign Risk: \\ Different Shocks, Different Channels\thanks{The authors thank, without implicating, Stephen Hansen (UCL), Daniel J. Lewis (UCL), Andreas Joseph (Bank of England), Juri Marcucci (Banca d'Italia), Santiago E. Alvarez-Blaser (Bank of Spain), Marina Diakonova (Bank of Spain), Javier Perez (Bank of Spain), Martin Saldias (Bank of Portugal) and participants in the internal seminar of Bank of Spain and BBVA research as well as participants 13th ECB Conference on Forecasting Techniques and ECB Internal DGE Seminar . We especially thank Buket Begun Boga, Patricia Soroa  and  Ismael Frutos for their contribution to build the database. }}
%\date{} 					% Uncomment to remove the date

\author{
    {\hspace{1mm}Alvaro Ortiz}\\
	BBVA Research \& CRIW(NBER) \\ 
    \texttt{alvaro.ortiz@bbva.com} \\
    \And
	{\hspace{1mm}Tomasa Rodrigo} \\
	BBVA Research \\
	\texttt{tomasa.rodrigo@bbva.com} 
    \And
    {\hspace{1mm}Pablo Saborido} \\
	BBVA Research\\
    \texttt{pablo.saborido@bbva.com} \\
}
\newcommand{\undertitle}{Version of March 20th, 2026} \date{} 
\maketitle

% Uncomment to override  the `A preprint' in the header
%\renewcommand{\headeright}{Technical Report}

\renewcommand{\shorttitle}{\textit{Geopolitics, Geoeconomics, and Sovereign Risk: Different Shocks, Different Channels}}

%%% Add PDF metadata to help others organize their library
%%% Once the PDF is generated, you can check the metadata with
%%% $ pdfinfo template.pdf

%%%------------------------------------- Abstract ------------------------------------
\begin{abstract}
Geopolitical and geoeconomic shocks reprice sovereign credit risk
through different transmission channels. Using a daily panel of 42
advanced and emerging economies over 2018--2025, we show that
geopolitical shocks raise sovereign CDS spreads primarily through
direct sovereign repricing, while the Global Financial Cycle (GFC)
channel moves in the opposite direction and partly offsets that
increase---a ``scissors pattern.'' Geoeconomic shocks, by contrast,
transmit mainly through financial conditions, policy uncertainty, and
domestic amplification, with only a limited direct repricing
component. A semistructural framework provides sign benchmarks for
four transmission channels, and a Shapley--Taylor decomposition of
nonlinear machine-learning predictions partitions each observation's
spread into Direct, GFC, Uncertainty, and Local components. Narrative
local projections around four dated crisis events recover the scissors
pattern for Russia--Ukraine and support the broader channel taxonomy
in the remaining episodes. Additional scorecard, placebo, and
sign-restricted SVAR evidence corroborates the taxonomy beyond the
baseline ML decomposition. Geopolitical direct effects decay with
distance from the conflict zone in a gravity-style pattern
($R^2 = 0.35$ for Russia--Ukraine), while policy-uncertainty shocks
activate the Uncertainty channel more globally. The taxonomy implies
that liquidity provision can mitigate GFC-driven spread widening, but
not direct geopolitical sovereign repricing.
\end{abstract}

% keywords can be removed
\keywords{Geopolitics \and Geoeconomics \and Sovereign Credit Risk \and 
Global Financial Cycle \and Transmission Channels \and 
Shapley--Taylor Decomposition \and Narrative Identification \and 
Machine Learning \and Text-as-Data}

\section{Introduction}
\label{sec:introduction}

Geopolitical and geoeconomic shocks both widen sovereign credit
spreads, but they do not do so through the same transmission channels.
Using a daily panel of 42 advanced and emerging economies over
2018--2025, we show that geopolitical shocks raise sovereign CDS
spreads primarily through direct sovereign repricing, while the Global
Financial Cycle (GFC) channel moves in the opposite direction and
partly offsets that increase---a ``scissors pattern.'' Geoeconomic
shocks, by contrast, load only weakly on direct repricing and transmit
mainly through financial conditions, policy uncertainty, and domestic
amplification.

This distinction matters for both theory and policy. When spreads
widen because the sovereign itself is repriced, the effect is
persistent, concentrated among exposed countries, and decays with
distance from the conflict zone in a gravity-style pattern. When
widening is mediated by common financial conditions or policy
uncertainty, it is more diffuse and potentially more reversible.
In policy terms, liquidity tools, swap lines, and backstop facilities
can address GFC-driven widening, but they do not eliminate the
persistent component of direct geopolitical sovereign repricing.
An originator penalty also emerges: geoeconomic-shock originators face
spread widening through the Local channel even as monetary easing
compresses spreads abroad.

The paper combines a semistructural benchmark with a three-layer
empirical design. The semistructural framework maps four standard
mechanisms---sovereign default, global pricing, policy-regime
uncertainty, and domestic amplification---into four empirical
channels: Direct, GFC, Uncertainty, and Local. It provides sign and
dominance benchmarks for short-horizon, panel-mean responses to
geopolitical and geoeconomic shocks. The empirical strategy then
proceeds in three layers.

First, we estimate a nonlinear forecasting model for sovereign CDS
spreads using high-frequency news-based indicators of geopolitical
risk, economic-policy uncertainty, trade-policy uncertainty, and
domestic political sentiment. A systematic model-comparison exercise
shows that these predictors add substantial forecast content only when
nonlinear interactions and state dependence are allowed, with
gradient-boosted trees outperforming linear alternatives. This first
layer establishes that the mapping from news and exposures to
sovereign spreads is not well approximated by a globally linear
specification.

Second, we use a Shapley--Taylor decomposition of the nonlinear
predictions to recover channel-level contributions for each
country-date observation. Guided by the semistructural sign
benchmarks, we partition predicted spreads into Direct, GFC,
Uncertainty, and Local components. The resulting taxonomy is
economically interpretable: geopolitical episodes are characterized by
strong direct repricing and offsetting GFC compression, whereas
geoeconomic episodes transmit primarily through GFC, Uncertainty, and
Local channels, with only limited direct loading.

Third, we validate this taxonomy using identification strategies that
do not rely on the baseline ML decomposition. Narrative local
projections around four dated crisis events---Russia--Ukraine,
Hamas--Israel, the U.S.\ Presidential Election, and Liberation Day
tariffs---recover the scissors pattern at the 1\% significance level
for Russia--Ukraine and confirm 15 of 16 event--channel sign
predictions. Evaluated against the semistructural benchmarks, the
point-estimate scorecard matches all 16 predictions, with a cluster
bootstrap mean of 15.9/16 (sd = 0.3) and 90\% confidence interval
$[15,\,16]$. A placebo falsification based on driver-specific
decompositions shows that all four episodes exceed at least 83\% of
random non-event dates, with the channels exiting the placebo envelope
differing across shock types in the direction the taxonomy predicts.
A complementary narrative sign-restricted SVAR estimated on raw
observable composites---entirely bypassing the Shapley--Taylor
decomposition---supports all eight sign and dominance predictions.
Because this exercise does not rely on the ML decomposition, it
provides independent evidence that the channel assignments reflect
properties of transmission rather than artifacts of model fitting.

The empirical design is intentionally layered. The semistructural
framework generates ex ante channel predictions, the nonlinear ML
stage captures threshold-dependent transmission, the Shapley--Taylor
decomposition constructs the channel objects, and the local projection
and narrative SVAR exercises provide complementary validation. Because
the channel predictions are stated before estimation, the scorecard
evaluates pre-specified benchmarks rather than ex post pattern
matching.

The paper proceeds as follows. Section~\ref{sec:thedata} describes
the data. Section~\ref{sec:benchmarks} develops the semistructural
framework and dominance benchmarks. Section~\ref{sec:Methodology}
presents the ML framework, the Shapley--Taylor channel decomposition,
and the local-projection design. Section~\ref{sec:results} evaluates
predictive performance and Shapley-based results.
Section~\ref{sec:transmission} develops the four-channel taxonomy
across the four crisis episodes. Section~\ref{sec:Causal_LP} provides
econometric validation via local projections.
Section~\ref{sec:taxonomy_test} evaluates the taxonomy against the
semistructural benchmarks. Section~\ref{sec:conclusion} concludes.

\subsection{Related Literature}

This paper connects to four main literatures.

First, we contribute to the literature that uses news-based indicators to measure geopolitical risk, economic-policy uncertainty, and trade-policy uncertainty \citep{baker2016, CaldaraEtAl2020TPU, CaldaraIacoviello2022, Bondarenko2024, AlonsoAlvarez2025, Gentzkow2019}. We show that the predictive content of these measures for sovereign risk is strongly nonlinear and depends on interactions with country exposures and domestic vulnerabilities.

Second, we contribute to the literature on sovereign risk, geopolitics,
and geoeconomic statecraft. Existing work shows that geopolitical risk
widens spreads \citep{FVFragmented2024, Boubaker2023}, that trade
tensions amplify uncertainty \citep{Ahn2020, AiyarPresbiteroRuta2023,
FVDarkShipping2025}, and that sovereign CDS spreads reflect both
default-related and global risk-premium components
\citep{Longstaff2011, Augustin2020}. We extend this literature by
showing that geopolitical and geoeconomic shocks operate through
qualitatively different transmission channels, providing a
sovereign-risk counterpart to recent work on geoeconomic vulnerability
and leverage \citep{ClaytonMaggioriSchreger2026}.

Third, we connect to the literature on the Global Financial Cycle and international asset pricing \citep{rey2013, BrunoShin2015, MirandaAgrippinoRey2020}. In sovereign markets, global financial conditions shape the risk-premium component of CDS spreads and interact with dollar funding conditions and safe-haven flows \citep{Longstaff2011, Du2018, BahajReis2022, Maggiori2020}. We show that the GFC channel does not respond uniformly across shock families: under geopolitical shocks it generates the scissors pattern, whereas under geoeconomic shocks it becomes a primary transmission margin.

Fourth, we speak to the growing literature combining machine learning
with economic identification \citep{Athey2019, Mullainathan2017,
Gu2020, ChernozhukovEtAl2024}. Our contribution is not to replace
causal designs with machine learning, but to assign each tool a clear
role: the nonlinear predictor flexibly captures threshold effects and
interactions; the Shapley--Taylor decomposition restores
interpretability at the observation level; and the narrative local
projections and sign-restricted SVAR provide external validation of
the resulting channel taxonomy.

\section{Data and Measurement}
\label{sec:thedata}

We assemble a daily panel of sovereign credit risk and its potential drivers for 42 advanced and emerging economies over January 2018 to July 2025.\footnote{The sample includes Argentina, Australia, Austria, Belgium, Brazil, Canada, Chile, China, Colombia, Czech Republic, Denmark, Egypt, Finland, France, Germany, Hungary, India, Indonesia, Israel, Italy, Japan, Jordan, Malaysia, Mexico, Morocco, Netherlands, Norway, Peru, Philippines, Poland, Qatar, Russia, Saudi Arabia, Sweden, Spain, Thailand, Turkey, Ukraine, United Kingdom, United States, Uruguay, and Vietnam.} The dependent variable is the 5-year sovereign credit default swap (CDS) spread---the annualized premium a protection buyer pays to insure against a credit event on sovereign debt.\footnote{The 5-year tenor is selected because it is the most liquid and widely traded segment of the CDS curve, establishing it as the standard benchmark for pricing default probability. Unlike government bond yield spreads, CDS spreads isolate default expectations from funding costs, interest rate risk, and bond-specific supply dynamics.} A large literature documents that sovereign spreads respond to both 
macroeconomic fundamentals and shifts in global investor sentiment 
\citep{Longstaff2011,PanSingleton2008,Aizenman2016,Augustin2020}; 
our channel decomposition formalizes this distinction. 

All variables are smoothed using a 28-day geometrically weighted moving average:
\begin{equation}
\tilde{X}_t = \sum_{k=0}^{27} w_k \, X_{t-k}\,,
\end{equation}
where the weight vector $\mathbf{w} = (w_0, \dots, w_{27})$ follows a discrete geometric distribution with $\sum_{k} w_k = 1$, assigning higher importance to more recent observations. The scheme is strongly front-loaded: approximately 69\% of the total weight falls within the most recent week and over 90\% within the last two weeks, ensuring high responsiveness to recent changes while spanning four complete weeks to equalize day-of-week composition \footnote{The exact weight vector is provided upon request in the replication package.}.
Once the weighted moving average is applied, the data is standardized to zero mean and unit variance within each country. We do not perform outlier treatment, as extreme events are informative signals of exceptional shifts in risk perception. The panel is unbalanced; country-specific data availability and percentile distributions are detailed in Appendix~A.

%% --- News-based indicators: GDELT and local sources ---

Our main contribution on the measurement side is the construction of daily, country-level news-based indicators from the \textit{Global Database of Events, Language, and Tone} (GDELT), an open-source platform that monitors broadcast, print, and online media in more than 100 languages \citep{Leetaru2013}.\footnote{An alternative source is the Dow Jones Factiva database, a premium archive widely used in academic research. The historical component of the Economic Policy Uncertainty index, for example, is constructed from Factiva news archives.} Following \citet{Bondarenko2024} and \citet{AlonsoAlvarez2025}, we construct all indicators from \textit{local-language} media rather than foreign or global outlets, because local newspapers more accurately reflect how geopolitical risks and policy uncertainties are perceived domestically.\footnote{\citet{Bondarenko2024} show that local-source measures embed heterogeneity in national perspectives and geographic proximity to conflict that global coverage smooths away. \citet{AlonsoAlvarez2025} formalize \emph{bilateral geopolitical risk} and document that local-source shocks have significant effects on domestic financial markets, whereas Anglosphere-media indicators systematically understate local impacts.}

%% --- Two types of indicators ---

We construct two types of news-based indicators. \textit{Coverage-based} indicators measure uncertainty solely through the volume of relevant news. For Economic Policy Uncertainty (EPU) and Trade Policy Uncertainty (TPU), we compute the daily share of articles containing predefined uncertainty-related keywords following \citet{baker2016}, divided by total daily news volume. Complete keyword lists are provided in Appendix~A. \textit{Tone-weighted} indicators combine volume with sentiment. For Economic Sentiment (ECO), Interest Rate Sentiment (INT), Geopolitical Risk (GPR), and Political Tensions (POL), we score relevant articles using over 40 GDELT sentiment dictionaries, yielding a tone score from $-10$ to $+10$, and multiply it by normalized coverage. The product is inverted so that higher values consistently correspond to greater risk.\footnote{Detailed construction formulas for all indicators are provided in Appendix~A. The complete set of daily and weekly indicators is publicly available at \url{https://bigdata.bbvaresearch.com/en/}.}

%% --- Variable summary table (replaces bullet lists) ---

Table~\ref{tab:variables} summarizes the full set of variables.\footnote{Our focus on high-frequency news-based indicators complements existing work emphasizing structural determinants of external debt dynamics.} The explanatory variables fall into three groups: global financial conditions, domestic macroeconomic sentiment, and political--geopolitical risk.

\begin{table}[ht!]
\centering
\caption{Variable, Sources, and Economic Interpretation }
\label{tab:variables}
\footnotesize
\begin{tabular}{@{}llll@{}}
\toprule
\textbf{Variable} & \textbf{Abbr.} & \textbf{Type / Source} & \textbf{Description} \\
\midrule
\multicolumn{4}{@{}l}{\textit{Dependent variable}} \\[2pt]
Sovereign CDS spread & CDS & Market (5Y) & Annualized premium on sovereign default insurance \\[6pt]
\multicolumn{4}{@{}l}{\textit{Global financial variables}} \\[2pt]
U.S.\ 2-year Treasury yield & FED & Market & Proxy for the monetary policy stance \citep{Swanson2021} \\
CBOE Volatility Index & VIX & Market & Implied S\&P\,500 volatility; global risk aversion \\[6pt]
\multicolumn{4}{@{}l}{\textit{Domestic macroeconomic sentiment (GDELT, local-language media)}} \\[2pt]
Economic Sentiment & ECO & Tone $\times$ coverage & Narrative framing of the domestic economy \\
Interest Rate Sentiment & INT & Tone $\times$ coverage & Expectations on monetary policy and borrowing costs \\
Econ.\ Policy Uncertainty & EPU & Coverage-based & Ambiguity regarding economic policy decisions \\
Trade Policy Uncertainty & TPU & Coverage-based & Uncertainty on trade rules, tariffs, and disputes \\[6pt]
\multicolumn{4}{@{}l}{\textit{Political and geopolitical risk (GDELT, local-language media)}} \\[2pt]
Geopolitical Risk & GPR & Tone $\times$ coverage & International conflict, military disputes, terrorism \\
Political Tensions & POL & Tone $\times$ coverage & Domestic instability, unrest, political contestation \\
\bottomrule
\end{tabular}
\end{table}

The use of local-language sources is not merely a data-collection 
choice; it reflects a substantive measurement decision. 
English-language indices such as the global GPR of 
\citet{CaldaraIacoviello2022}---constructed from ten Anglosphere 
newspapers---measure when a country attracts international media 
attention, not how geopolitical risk is perceived domestically. 
\citet{Bondarenko2024} demonstrate this distinction sharply: 
geopolitical risk shocks identified from local Russian-language 
newspapers have significant adverse effects on the Russian 
economy, whereas shocks from English-language sources do not. 
\citet{AlonsoAlvarez2025} formalize bilateral geopolitical risk 
and show that Anglosphere-media indicators systematically 
understate local financial-market impacts. Because our 
decomposition attributes sovereign spread movements to 
country-specific news indicators, the accuracy of these 
indicators at the country level is a first-order concern: 
mismeasured local risk would contaminate the Direct and Local 
channels and bias the transmission taxonomy.
============================================================
\section{Geopolitics \& Geoeconomics Transmission Channels and Dominance Benchmarks}\label{sec:gfc}
\label{sec:benchmarks}

This section develops a semistructural framework to distinguish how geopolitical and geoeconomic shocks transmit to sovereign risk. Rather than imposing a fully structural model, the framework isolates four economically interpretable channels (Direct, GFC, Uncertainty, and Local) and derives sign and dominance predictions for their short-run effects on sovereign spreads. The key implication is that geopolitical shocks should load primarily on direct sovereign repricing, with an offsetting role for global financial conditions, whereas geoeconomic shocks should transmit mainly through financial conditions, policy uncertainty, and domestic amplification. The five subsections that follow develop these mechanisms and summarize the corresponding benchmarks.

\subsection{Block 1: Sovereign Default and the Direct Channel}

Following \citet{Arellano2008}, the direct component of the spread rises with the perceived probability of sovereign distress:
\begin{equation}
  s^{\textsc{dir}}_{i,t} \;\propto\; \Pr\!\Big(y_{i,t+1} < \underline{y}_i(b_i) \;\Big|\; \mathcal{F}_t\Big),
\end{equation}
where $y_{i,t}$ denotes fiscal capacity or repayment capacity, $b_i$ is outstanding debt, and $\underline{y}_i(b_i)$ is a debt-dependent distress threshold.

The key distinction is how the two shock families affect repayment capacity on impact.  Geopolitical shocks can reduce expected fiscal capacity through destruction, trade disruption, sanctions on state capacity, regional spillovers, or capital flight from exposed sovereigns:
\begin{equation}
  \frac{\partial y_{i,t}}{\partial \varepsilon^G_t} < 0 \quad \text{for sufficiently exposed countries.}
\end{equation}

%% REVISED — old: "≈ 0 on impact and at the panel mean"
%% new: explicit dominance statement
By contrast, geoeconomic shocks typically do not impair repayment capacity as directly at short horizons.  They may change growth expectations, fiscal expectations, or discount rates, but their impact on sovereign default risk is usually more indirect and weaker than under geopolitical shocks:
\begin{equation}
  \bigg|\frac{\partial y_{i,t}}{\partial \varepsilon^E_t}\bigg|
  \;\ll\;
  \bigg|\frac{\partial y_{i,t}}{\partial \varepsilon^G_t}\bigg|
  \quad \text{on impact and at the panel mean.}
\end{equation}
Election-type or fiscal-regime episodes may still generate modest direct repricing---particularly for the originator sovereign, where policy uncertainty feeds back into fiscal credibility through the Local channel---but not the large and systematic activation expected under geopolitical shocks.

%% REVISED prediction
\begin{quote}
\textbf{Prediction D.}\; $\partial s^{\textsc{dir}}/\partial\varepsilon^G > 0$;\;
$|\partial s^{\textsc{dir}}/\partial\varepsilon^E| \ll |\partial s^{\textsc{dir}}/\partial\varepsilon^G|$
at short horizons and at the panel mean.
\end{quote}

% --------------------------------------------------
\subsection{Block 2: Global Pricing and the GFC Channel}

Let the global-financial component of sovereign spreads load on a common discount-rate factor:
\begin{equation}
  s^{\textsc{gfc}}_{i,t} = \beta^{\textsc{gfc}}_i \, q_t,
\end{equation}
where $q_t$ summarizes global financing conditions or the common risk-price component embedded in sovereign spreads, and $\beta^{\textsc{gfc}}_i$ measures country exposure to that common factor.  This reduced-form representation is consistent with intermediary-asset-pricing environments such as \citet{HeKrishnamurthy2013} and international-finance frameworks such as \citet{GabaixMaggiori2015}.

For geopolitical shocks, the benchmark sign comes from safe-haven reallocation and benchmark-rate compression.  When conflict risk rises, capital tends to move toward safe assets, lowering the safe-rate component relevant for the pricing of many sovereign claims.  For most non-targeted borrowers, this compresses the global discount-rate component of spreads even as the targeted sovereigns experience a rise in the Direct channel.  At the panel mean, this generates:
\begin{equation}
  \frac{\partial q_t}{\partial \varepsilon^G_t} < 0.
\end{equation}
The coexistence of $\partial s^{\textsc{dir}}/\partial\varepsilon^G > 0$ and $\partial s^{\textsc{gfc}}/\partial\varepsilon^G < 0$---sovereign-specific repricing moving in the opposite direction from the common financial factor---is what we term the \emph{scissors condition}.  It requires that the portfolio rebalancing effect dominates any general risk-aversion increase for most panel members, which is satisfied when the geopolitical shock triggers safe-haven flows without impairing the balance sheet of global intermediaries directly.\footnote{This mechanism connects to the distinction between vulnerability and leverage in the geoeconomic framework of \citet{ClaytonMaggioriSchreger2026}: a geopolitical shock creates vulnerability for the targeted sovereign but generates a partial financial offset for bystander sovereigns through the intermediary's portfolio adjustment.  Our scissors condition is the sovereign-risk counterpart of their structural insight.}

For geoeconomic shocks, the sign is similar but the mechanism is different.  Tariffs, policy-regime changes, and election outcomes can alter expected growth and the expected path of monetary policy.  At short horizons, those shocks often lower expected policy rates or otherwise compress the common discount-rate factor:
\begin{equation}
  \frac{\partial q_t}{\partial \varepsilon^E_t} < 0.
\end{equation}

%% REVISED — new paragraph explaining that GFC discriminates by pattern, not by sign alone
Thus both shock families can load negatively on the GFC channel, but for different reasons: geopolitical shocks through safe-haven and benchmark-rate effects, geoeconomic shocks through expected policy and discount-rate revaluation.  The GFC channel therefore discriminates between shock types not through its own sign---which is negative in both cases---but through its \emph{relative role in the channel constellation}: under geopolitical shocks, the GFC is the secondary, offsetting channel (the scissors); under geoeconomic shocks, it is the dominant transmission channel.  This benchmark is intentionally local to the event windows studied here; a sufficiently systemic financial crisis need not satisfy it.

\begin{quote}
\textbf{Prediction G.}\; $\partial s^{\textsc{gfc}}/\partial\varepsilon^G < 0$;\;
$\partial s^{\textsc{gfc}}/\partial\varepsilon^E < 0$.
The distinguishing feature is relative dominance: $|\varphi^{\textsc{gfc}}| < |\varphi^{\textsc{dir}}|$ under geopolitical shocks, $|\varphi^{\textsc{gfc}}| > |\varphi^{\textsc{dir}}|$ under geoeconomic shocks.
\end{quote}

% --------------------------------------------------
\subsection{Block 3: Policy-Regime Uncertainty and the Uncertainty Channel}

Following \citet{PastorVeronesi2012,PastorVeronesi2013}, let the uncertainty component depend on beliefs over domestic economic-policy regimes.  Let $\pi_t$ denote the perceived probability of one policy regime relative to another; then a convenient reduced-form representation is
\begin{equation}
  s^{\textsc{unc}}_{i,t} = \alpha^{\textsc{unc}}_i \, H(\pi_t),
\end{equation}
where $H(\pi_t)$ is an uncertainty index, such as Shannon entropy, and $\alpha^{\textsc{unc}}_i$ measures country sensitivity to policy-regime ambiguity.

The relevant distinction is between economic-policy uncertainty and generic uncertainty.  EPU and TPU are designed to capture uncertainty about taxes, trade barriers, regulation, fiscal priorities, and related policy choices.  Geoeconomic shocks naturally activate this margin because they directly alter beliefs about policy regimes:
\begin{equation}
  \frac{\partial H(\pi_t)}{\partial \varepsilon^E_t} > 0.
\end{equation}

%% REVISED — old: "≈ 0" → new: dominance + temporal precision
Geopolitical shocks need not do so.  A military conflict, territorial attack, or security shock may sharply raise aggregate uncertainty, but it does not automatically change the distribution over domestic economic-policy regimes.  At the short horizons of our event windows (1--3 months), geopolitical shocks have typically not yet cascaded into domestic policy-regime uncertainty---that cascade may materialise at longer horizons as sanctions and fiscal responses unfold.  Accordingly, the short-horizon benchmark for the Uncertainty channel under geopolitical shocks is substantially smaller than under geoeconomic shocks:
\begin{equation}
  \bigg|\frac{\partial H(\pi_t)}{\partial \varepsilon^G_t}\bigg|
  \;\ll\;
  \bigg|\frac{\partial H(\pi_t)}{\partial \varepsilon^E_t}\bigg|.
\end{equation}

Within geoeconomic shocks, the activation of the Uncertainty channel is expected to be strongest for policy-regime shocks such as elections and tariff-policy episodes.  Sanctions or energy-policy episodes may load more weakly on this channel and more strongly on the GFC or Local channels.

%% REVISED prediction
\begin{quote}
\textbf{Prediction U.}\;
$|\partial s^{\textsc{unc}}/\partial\varepsilon^G| \ll |\partial s^{\textsc{unc}}/\partial\varepsilon^E|$
at short horizons, especially for EPU/TPU-type shocks.
\end{quote}

% --------------------------------------------------
\subsection{Block 4: Domestic Amplification and the Local Channel}

The Local channel captures the interaction between external shocks 
and domestic vulnerabilities.  Let
\begin{equation}
  s^{\textsc{loc}}_{i,t} = \psi_i(z_{i,t})\,\varepsilon_t, \quad 
  \psi'_i(z_{i,t}) > 0,
\end{equation}
where $z_{i,t}$ collects domestic state variables and $\psi_i(\cdot)$ 
is increasing in vulnerability.  Two components of $z_{i,t}$ are 
particularly relevant for sovereign risk.  Local economic-activity 
sentiment captures shifts in perceived growth and fiscal capacity: a 
deterioration lowers expected primary surpluses and moves the 
sovereign closer to its debt-sustainability boundary.  Local 
interest-rate sentiment captures shifts in domestic borrowing costs 
and rollover conditions: a tightening raises the effective cost of 
servicing outstanding debt and amplifies the fiscal impact of any 
external shock.  Together, these two margins determine how much a 
given external impulse translates into sovereign-spread widening 
through the domestic channel.  The remaining elements of 
$z_{i,t}$---trade exposure, external dependence, domestic political 
conditions, and regional sensitivity---modulate the strength of this 
amplification across countries.

Geoeconomic shocks have a natural affinity with domestic amplification 
because they operate through heterogeneous bilateral exposures.  
Tariffs affect countries differently according to trade structure; 
election shocks affect countries differently according to policy 
alignment and financial exposure; sanctions and energy measures affect 
countries differently according to import dependence, sectoral links, 
and domestic politics.  Crucially, these shocks also shift local 
activity and interest-rate sentiment differentially: a tariff shock 
depresses growth expectations in export-dependent economies while 
tightening borrowing conditions in fiscally exposed ones, activating 
both margins of $z_{i,t}$ simultaneously.  These heterogeneous effects 
can survive aggregation and therefore appear at the panel mean:
\begin{equation}
  \mathbb{E}_i\!\left[\frac{\partial s^{\textsc{loc}}_{i,t}}
  {\partial \varepsilon^E_t}\right] > 0.
\end{equation}

Geopolitical shocks can also generate substantial local responses for 
individually exposed countries (e.g., Poland's 
$\varphi^{\textsc{loc}} = +0.06$ during Russia--Ukraine).  However, 
at the panel mean these effects are secondary to the Direct and GFC 
channels, because geopolitical shocks simultaneously produce a broad 
common repricing that dominates the cross-sectional average.  By 
contrast, the bilateral heterogeneity of geoeconomic shocks means 
that local amplification---operating through activity and 
interest-rate sentiment---\emph{is} the primary differentiator across 
countries:
\begin{equation}
  \mathbb{E}_i\!\left[\frac{|\partial s^{\textsc{loc}}_{i,t}/
  \partial \varepsilon^G_t|}{|\partial s^{\textsc{dir}}_{i,t}/
  \partial \varepsilon^G_t|}\right]
  \;\ll\;
  \mathbb{E}_i\!\left[\frac{|\partial s^{\textsc{loc}}_{i,t}/
  \partial \varepsilon^E_t|}{|\partial s^{\textsc{dir}}_{i,t}/
  \partial \varepsilon^E_t|}\right].
\end{equation}

\begin{quote}
\textbf{Prediction L.}\;
Under geopolitical shocks, 
$\mathbb{E}_i[|\varphi^{\textsc{loc}}|] \ll 
\mathbb{E}_i[|\varphi^{\textsc{dir}}|]$;\;
under geoeconomic shocks, 
$\mathbb{E}_i[\partial s^{\textsc{loc}}/\partial\varepsilon^E] > 0$
and $\varphi^{\textsc{loc}}$ is a primary transmission channel.
\end{quote}

% --------------------------------------------------

\subsection{Dominance Benchmarks and Nonlinearities}

Table~\ref{tab:benchmarks} summarizes the resulting benchmarks.  
The framework implies a \emph{scissors pattern} for geopolitical 
shocks: the Direct channel rises while the GFC channel falls, with 
the Uncertainty and Local channels playing a secondary role at the 
panel mean.  Geoeconomic shocks display a different constellation: 
the Direct channel is small relative to its geopolitical counterpart, 
and the response operates primarily through the GFC, Uncertainty, and 
Local channels.  The benchmarks are stated as sign predictions for 
strong channels and \emph{dominance inequalities} for secondary 
channels---a formulation that is directly testable by comparing 
magnitudes across episodes.

Three features deserve emphasis.  First, the four blocks are modular: 
each channel is tied to a distinct mechanism, so a miss in one margin 
does not invalidate the others.  Second, the scissors pattern is not 
imposed mechanically; it follows from the coexistence of 
sovereign-specific repricing and common-factor compression under 
geopolitical shocks.  Third, the dominance inequalities yield a clear 
pass/fail criterion: comparing $|\varphi^{\textsc{dir}}|$ or 
$|\varphi^{\textsc{unc}}|$ across geopolitical and geoeconomic 
episodes is sufficient.  The scorecard in 
Section~\ref{sec:taxonomy_test} evaluates these benchmarks across all 
crisis episodes, counting both sign confirmations ($\uparrow$, 
$\downarrow$) and dominance confirmations ($\ll$).

The framework also motivates the empirical specification.  Default 
risk is threshold-based, policy-regime uncertainty is nonlinear in 
beliefs, and domestic amplification arises from interactions between 
external shocks and country-specific vulnerabilities.  The 
reduced-form mapping from observables to sovereign spreads is 
therefore unlikely to be well captured by a globally linear model, 
motivating the boosted-tree ensemble, Shapley--Taylor decomposition, 
and panel local projections that follow.
\begin{table}[t]
\centering
\caption{Dominance Benchmarks by Shock Family}
\label{tab:benchmarks}
\small
\begin{tabular}{l @{\hskip 12pt} c @{\hskip 10pt} c @{\hskip 10pt} c @{\hskip 10pt} c}
\toprule
 & \multicolumn{4}{c}{Channel response} \\
\cmidrule(lr){2-5}
Shock family
  & $\varphi^{\textsc{dir}}$ & $\varphi^{\textsc{gfc}}$ & $\varphi^{\textsc{unc}}$ & $\varphi^{\textsc{loc}}$ \\
\midrule
Geopolitical ($\varepsilon^G$: GPR)
  & $\uparrow$
  & $\downarrow$\,\textsuperscript{(a)}
  & $\ll\!\varphi^{\textsc{unc}}_{\textsc{geoeco}}$
  & $\ll\!\varphi^{\textsc{dir}}$ \\[4pt]
Geoeconomic ($\varepsilon^E$: EPU/TPU)
  & $\ll\!\varphi^{\textsc{dir}}_{\textsc{geo}}$
  & $\downarrow$\,\textsuperscript{(b)}
  & $\uparrow$
  & $\uparrow$ \\
\midrule
\multicolumn{5}{l}{\footnotesize Mechanism} \\
  & \footnotesize Default risk
  & \footnotesize Global pricing
  & \footnotesize Policy-regime
  & \footnotesize Domestic \\
  & \footnotesize
  & \footnotesize (scissors)
  & \footnotesize uncertainty
  & \footnotesize amplification \\[2pt]
\multicolumn{5}{l}{\footnotesize Canonical reference} \\
  & \footnotesize Arellano
  & \footnotesize He--K\,/\,G--M
  & \footnotesize P--V
  & \footnotesize Threshold\,/\,exposure \\
\bottomrule
\end{tabular}

\vspace{4pt}
\begin{minipage}{0.92\textwidth}\footnotesize
\emph{Notes:} Signs refer to short-horizon, panel-mean responses around identified events.
$\ll$ denotes a dominance inequality: the channel response is predicted to be substantially smaller in magnitude than the comparator.
The framework is a benchmark for signs and relative activation, not a fully estimated structural model.
``Geoeconomic'' encompasses policy-regime shocks (elections, tariff episodes), which are expected to load most strongly on the Uncertainty channel, and sanctions- or energy-related episodes, which may load relatively more on the GFC and Local channels.
\textsuperscript{(a)}~Under geopolitical shocks, $|\varphi^{\textsc{gfc}}| < |\varphi^{\textsc{dir}}|$ (the GFC is the offsetting, secondary channel).
\textsuperscript{(b)}~Under geoeconomic shocks, $|\varphi^{\textsc{gfc}}| > |\varphi^{\textsc{dir}}|$ (the GFC is the dominant channel).
Arellano: \citet{Arellano2008}; He--K: \citet{HeKrishnamurthy2013}; G--M: \citet{GabaixMaggiori2015}; P--V: \citet{PastorVeronesi2012,PastorVeronesi2013}.
The scissors condition refers to the coexistence of $\partial s^{\textsc{dir}}/\partial\varepsilon^G > 0$ and $\partial s^{\textsc{gfc}}/\partial\varepsilon^G < 0$ under geopolitical shocks (see Section~\ref{sec:gfc}).
\end{minipage}
\end{table}

\section{Econometric Framework}
\label{sec:Methodology}

Our empirical framework has three steps. First, we compare a broad set of forecasting models and select the specification with the best pseudo-real-time out-of-sample performance (Section~\ref{sec:ML_approach}). Second, holding that specification fixed, we use Shapley--Taylor interaction values to decompose fitted forecasts into four economically interpretable transmission channels (Section~\ref{sec:shapley_interpretability}). Third, we assess whether this taxonomy receives independent support in panel local projections under two identification strategies (Section~\ref{sec:causal_framework}).

%% ============================================================
%% 3.1 FORECASTING MODEL COMPARISON
%% ============================================================
\subsection{Forecasting Model Comparison}
\label{sec:ML_approach}

We model sovereign risk as a potentially nonlinear function of market and news-based predictors. For country~$i$ on date~$t$, model class~$m$ delivers the one-step-ahead forecast
\begin{equation} \label{eq:canonical}
y_{i,t+1} = f_m(\bm{X}_{i,t}, \bm{\lambda}_i) + \varepsilon_{i,t+1},
\end{equation}
where $y_{i,t+1}$ is the standardized sovereign CDS spread, $\bm{X}_{i,t}$ is the information set available at date~$t$, and $\bm{\lambda}_i$ denotes country fixed effects. We compare fifteen model classes spanning linear regression, penalized linear models \citep{Hoerl1970, Tibshirani1996, ZouHastie2005}, tree-based ensembles \citep{Breiman2001, Friedman2001}, and neural networks \citep{LeCun2015}.

Forecasts are generated recursively in pseudo-real time using an expanding estimation window and a 28-day exclusion buffer around each train--test split to prevent leakage from overlapping moving-average windows.\footnote{Hyperparameters are selected by cross-validation within the training sample and held fixed during out-of-sample evaluation. The evaluation period runs from February 2021 to July 2025, so all four crisis episodes are assessed out of sample.} Model selection is based exclusively on pooled out-of-sample MAE and RMSE. Once the preferred architecture and hyperparameters are selected, they are held fixed. We then re-estimate that specification on the full estimation sample and use the fitted model only for the interpretability exercise in Section~\ref{sec:shapley_interpretability}. This separation ensures that model selection is not influenced by the decomposition stage.

To quantify the incremental predictive value of news-based indicators, we compare two predictor sets. The \emph{Markets-Only} benchmark contains the VIX and the U.S.\ two-year Treasury yield, providing a parsimonious baseline centered on the Global Financial Cycle \citep{Calvo1996, rey2013, MirandaAgrippinoRey2020}. The augmented specification adds six news-based indicators: geopolitical risk (GPR), economic policy uncertainty (EPU), trade policy uncertainty (TPU), and three domestic macro-political indicators (ECO, INT, POL). For each model class, the change in out-of-sample loss between the two predictor sets measures the marginal predictive contribution of news and reveals which forecasting technologies are most effective in exploiting it. Section~\ref{sec:results} reports results; the core finding is that news variables improve all model classes, but gains are substantially larger for nonlinear methods (15--19\%) than for linear specifications (5--9\%), implying that the predictive content of news operates through interactions and threshold effects \citep{Varian2014, Gu2020}.\footnote{The preferred specification is the Multilayer Random Forest (Two Stages), which combines near-frontier predictive performance with a substantive architectural advantage: the first stage separates advanced-economy from emerging-market panels, and the second refines predictions with region-specific dynamics}

A methodological caveat applies: the full-sample re-estimation means that the Shapley--Taylor channels have observed the crisis episodes used in the narrative LP\@. The LP validation therefore tests whether the ML-constructed channels respond to identified shocks in theory-consistent ways that the model was not optimised to produce. The narrative sign-restricted SVAR, which uses raw observables, provides fully independent cross-validation.
%% ============================================================
%% 3.2 SHAPLEY-BASED CHANNEL DECOMPOSITION
%% ============================================================
\subsection{Shapley-Based Channel Decomposition}
\label{sec:anatomy}
\label{sec:shapley_interpretability}

After selecting the preferred forecasting specification, we freeze its architecture and use it solely for interpretability---the Shapley analysis attributes predictions, not causal effects. Although the target variable is $y_{i,t+1}$, the decomposition is evaluated on the predictor vector observed at date~$t$. Accordingly, the object being decomposed is the fitted forecast $\hat{y}_{i,t+1\mid t} = f^\star(\bm{X}_{i,t})$, and all channel contributions are indexed by the information date~$t$.

\paragraph{Economic logic of the four channels.}
We partition the predictors into four blocks based on their economic 
role: a \emph{global financial cycle} block 
$\mathcal{G} = \{\text{VIX}, \text{US2Y}\}$ containing the two 
variables common to all 42 countries; an \emph{economic and trade 
uncertainty} block $\mathcal{U} = \{\text{EPU}, \text{TPU}\}$; a 
\emph{domestic macro-political} block 
$\mathcal{L} = \{\text{ECO}, \text{INT}, \text{POL}\}$; and a 
\emph{regional structure} block 
$\mathcal{R} = \{\text{REG}\}$, where REG denotes regional group 
indicators that encode geographic, institutional, and income-group 
heterogeneity.\footnote{Regional indicators enter the forecasting 
model as categorical features and generate Shapley interactions with 
all other predictors.} We absorb $\mathcal{R}$ into the Local 
channel rather than introducing a fifth channel because regional 
structure operates as a conditioning factor that modulates the 
domestic amplification of external shocks, not as an independent 
transmission mechanism.\footnote{Figure~\ref{fig:gravity} supports this design choice: under the
log-distance specification of equation~(\ref{eq:gravity_ols}), the Direct
channel decays with distance to the conflict epicenter ($R^2 = 0.35$ for
Russia--Ukraine, $p < 0.001$; $R^2 = 0.22$ for Hamas--Israel, $p < 0.01$),
while the regional component of the Local channel shows no systematic
relationship with distance ($R^2 = 0.003$, $p = 0.72$;
Appendix Figure~\ref{fig:gravity_direct_vs_REG}).}

These blocks motivate four channels, each isolating a distinct transmission mechanism. The \emph{Direct channel} measures a driver's own contribution orthogonal to all other predictors. The \emph{GFC channel} aggregates interactions with VIX and US2Y, capturing transmission through the global risk-taking and monetary environment \citep{rey2013, MirandaAgrippinoRey2020}. The \emph{Uncertainty channel} captures cross-reinforcement among policy-uncertainty measures \citep{Bloom2009, baker2016, CaldaraEtAl2020TPU}. The \emph{Local channel} collects interactions with domestic sentiment, political tensions, and regional structure---daily-frequency proxies for the core determinants of sovereign creditworthiness (growth prospects, interest rate--growth differential, institutional quality)---capturing the degree to which local vulnerabilities amplify or absorb external shocks.

\paragraph{Formal decomposition.}
For each observation we compute pairwise Shapley--Taylor interaction 
values \citep{Shapley1953, lundberg2017}. These attributions satisfy 
local accuracy, so they decompose the fitted forecast as
\begin{equation}
\hat{y}_{i,t+1\mid t}
\;=\; \varphi_{0}
\;+\; \sum_{j=1}^{M} \varphi_{jj,i,t}
\;+\; \sum_{j<k} \varphi_{jk,i,t}\,,
\label{eq:taylor_shap}
\end{equation}
where $\varphi_{0}$ is the baseline prediction, $\varphi_{jj,i,t}$ 
the main effect of predictor~$j$, and $\varphi_{jk,i,t}$ the 
pairwise interaction between predictors~$j$ and 
$k$.\footnote{Shapley values are the unique attribution satisfying 
local accuracy (attributions sum to the prediction) and consistency 
(if a feature's marginal impact increases, its attribution cannot 
decrease). In practice we compute pairwise interaction values using 
TreeSHAP. The algorithm produces a negligible numerical residual 
$\eta_{i,t}$ with median magnitude below $10^{-4}\sigma$; we treat 
the decomposition as exact throughout.} The pairwise structure is 
essential for the channel decomposition that follows. Standard 
first-order Shapley values assign a single aggregate attribution to 
each predictor, conflating a variable's own contribution with its 
interactions with other predictors. The Shapley--Taylor extension 
separates the main effect $\varphi_{jj}$ from each pairwise 
interaction $\varphi_{jk}$, which is what allows us to distinguish a 
driver's \emph{direct} effect on sovereign risk from its 
\emph{indirect} transmission through the global financial cycle, 
uncertainty, or domestic conditions. Without this second-order 
decomposition, the four-channel partition would not be identified. 
For any driver~$Z$, we partition its total contribution into the four 
channels:
\begin{equation}
\underbrace{\varphi^{Z,\mathrm{tot}}_{i,t}}_{\text{total}}
\;=\;
\underbrace{\varphi_{ZZ,i,t}\vphantom{\sum_{k}}}_{\varphi^{Z,\mathrm{dir}}_{i,t}}
\;+\;
\underbrace{\displaystyle
    \sum_{k\,\in\,\mathcal{G}}\varphi_{Zk,i,t}
}_{\varphi^{Z,\mathrm{GFC}}_{i,t}}
\;+\;
\underbrace{\displaystyle
    \sum_{\substack{k\,\in\,\mathcal{U}\\[1pt]k\,\neq\,Z}}\varphi_{Zk,i,t}
}_{\varphi^{Z,\mathrm{UNC}}_{i,t}}
\;+\;
\underbrace{\displaystyle
    \sum_{\substack{k\,\in\,\mathcal{L}\,\cup\,\mathcal{R}\\[1pt]k\,\neq\,Z}}\varphi_{Zk,i,t}
}_{\varphi^{Z,\mathrm{LOC}}_{i,t}}
\label{eq:channels}
\end{equation}
By construction, these four terms sum to 
$\varphi^{Z,\mathrm{tot}}_{i,t}$ for every 
observation.\footnote{When $Z \in \mathcal{U}$, the Uncertainty 
channel contains only the single cross-interaction between EPU and 
TPU\@. When $Z \in \mathcal{L}$, the Local channel excludes $Z$'s 
own main effect, which enters the Direct channel. Regional 
indicators in $\mathcal{R}$ contribute to the Local channel for all 
drivers~$Z$.} Section~\ref{sec:transmission} uses these channel 
series to characterize how the transmission of different shock types 
varies over time, across countries, and across the cross-sectional 
distribution of sovereign risk.

%% ============================================================
%% 3.3 ECONOMETRIC VALIDATION VIA LOCAL PROJECTIONS
%% ============================================================
\subsection{Econometric Validation: Local Projections with Identified Shocks}
\label{sec:causal_framework}

The Shapley decomposition yields constructed channel series, but it does not by itself establish that these series capture economically meaningful transmission mechanisms. To assess whether the proposed taxonomy receives independent support, we estimate panel local projections under two identification strategies. Section~\ref{sec:Causal_LP} reports the results.

\subsubsection{Panel Local Projection Specification}
\label{sec:LP_spec}

For each horizon $h = 0, 1, \ldots, 90$ days, we estimate:
\begin{equation}
\label{eq:LP}
\Delta_{h} \, Y^{ch}_{i,t+h} \;=\; \alpha^{h}_{i} \;+\; \beta^{h} \, S_{i,t} \;+\; \sum_{p=1}^{5} \gamma^{h}_{p} \, Y^{ch}_{i,t-p} \;+\; {\delta^{h}}' \, \mathbf{X}_{i,t} \;+\; u^{h}_{i,t+h}\,,
\end{equation}
where $\Delta_{h} Y^{ch}_{i,t+h} \equiv Y^{ch}_{i,t+h} - Y^{ch}_{i,t-1}$ is the cumulative change in the outcome; $Y^{ch}$ denotes either the raw standardized CDS spread or one of the four channel series $\{\varphi^{\mathrm{dir}}, \varphi^{\mathrm{GFC}}, \varphi^{\mathrm{UNC}}, \varphi^{\mathrm{LOC}}\}$; $\alpha^{h}_{i}$ are country fixed effects; $\mathbf{X}_{i,t}$ collects contemporaneous controls; and $S_{i,t}$ is the identified shock. This generic notation accommodates both identification strategies: under Strategy~I, $S_{i,t}$ is country-specific; under Strategy~II, it is common across countries for a given event date. Standard errors use the \citet{DriscollKraay1998} estimator, robust to heteroskedasticity, serial correlation up to $h+1$ lags, and cross-sectional dependence. We report 68\% and 90\% confidence bands throughout. The estimation covers 75,650 country--day observations.

\subsubsection{Strategy~I: Full-Sample Innovation Identification}
\label{sec:strategy_AR5}

We identify shocks as innovations from country-specific AR(5) processes:
\begin{equation}
\label{eq:AR5}
Z_{i,t} \;=\; \mu_{i} + \sum_{p=1}^{5} \rho_{i,p} \, Z_{i,t-p} + \varepsilon^{Z}_{i,t}\,.
\end{equation}
The standardized residual $S^{Z}_{i,t} = \hat{\varepsilon}^{Z}_{i,t} \,/\, \hat{\sigma}_{\varepsilon,i}$ serves as the shock in equation~(\ref{eq:LP}).\footnote{Qualitatively we obtain similar results under panel AR(5) specifications with country fixed effects or from simple first differences $\Delta Z_{i,t}$. See Appendix~\ref{app:LP_FD}.} This strategy estimates the average marginal response to a one-standard-deviation country-level innovation, pooling across the entire sample. Its strength is broad coverage. Its limitation is that it averages over routine and crisis periods alike, which attenuates episode-specific transmission patterns---an attenuation that the scissors mechanism makes especially severe, since the offsetting Direct and GFC channels tend to cancel in aggregate.

\subsubsection{Strategy~II: Narrative Identification}
\label{sec:strategy_narrative}

We construct event dummies equal to one in a $\pm 3$-day window around each of four dated events:
\begin{equation}
\label{eq:NAR}
D^{e}_{t} \;=\; \mathbb{1}\!\big(|t - t_{e}| \leq 3\big), \qquad e \in \{E1, E2, E3, E4\}\,,
\end{equation}
where $t_{E1}$ = 24 February 2022 (Russia--Ukraine invasion), $t_{E2}$ = 7 October 2023 (Hamas--Israel attack), $t_{E3}$ = 5 November 2024 (U.S.\ presidential election), and $t_{E4}$ = 2 April 2025 (``Liberation Day'' tariff announcement). A separate local projection is estimated for each event, with $S_{i,t} = D^{e}_{t}$. Each event is paired with its primary news indicator: GPR for $E1$ and $E2$, EPU for $E3$, TPU for $E4$.

The narrative approach offers two advantages. First, event timing is plausibly exogenous to day-to-day sovereign CDS movements. Second, the taxonomy generates event-specific predictions---geopolitical events should primarily activate the Direct channel, whereas geoeconomic events should largely bypass it---that the narrative local projections evaluate directly.

\section{The Layered Structure of Sovereign Risk}
\label{sec:results}

This section establishes three facts that motivate the transmission 
analysis. First, news-based indicators add material predictive content 
beyond global financial variables, but primarily through nonlinear 
models. Second, in the selected specification, the largest average 
contributions come from global financial conditions, while domestic 
sentiment differentiates countries within that common environment. 
Third, geopolitical and policy-uncertainty indicators matter primarily 
through nonlinear interactions rather than through large average main 
effects. Together, these results motivate the four-channel 
decomposition developed in Section~\ref{sec:transmission}.

% ─────────────────────────────────────────────────────────────────────────────
\subsection{Model Selection}
\label{subsec:horse_race}

All predictors and the dependent variable are constructed as 28-day 
moving averages, so adjacent observations mechanically share underlying 
daily information. To avoid leakage, we evaluate all model classes in a 
recursive pseudo-real-time design with a 28-day exclusion buffer around 
each forecast origin. This buffer ensures that no training observation 
overlaps with the test target, so any improvement from adding 
news-based indicators can be interpreted conservatively as genuine 
incremental predictive content.\footnote{A simple AR(1) benchmark 
performs worse than the Markets-Only specification, confirming that the 
latter provides a stricter baseline.}
\begin{table}[htbp]
\centering
\caption{\textbf{News Indicators Improve Nonlinear Forecasts of Sovereign Risk}}\label{tab:forecast}\label{tab:mae_rmse}
\vspace{6pt}

\resizebox{1.0\textwidth}{!}{%
\renewcommand{\arraystretch}{1.1}
{\footnotesize
\begin{tabular}{|l|cc|cc|cc|cc|}
\hline
\multirow{3}{*}{Machine Learning Model}
  & \multicolumn{2}{c|}{Benchmark}
  & \multicolumn{2}{c|}{News Extended}
  & \multicolumn{4}{c|}{Difference (News Extended vs Benchmark)} \\
& \multicolumn{2}{c|}{Market Only}
  & \multicolumn{2}{c|}{Market + News}
  & \multicolumn{2}{c|}{RMSE}
  & \multicolumn{2}{c|}{MAE} \\ \cline{2-9}
& RMSE & MAE & RMSE & MAE & Diff & \% Var & Diff & \% Var \\
\hline
Linear Regression              & 1.09 & 0.92 & 1.03 & 0.86 & -0.06 &  -5.7\% & -0.06 &  -6.4\% \\
Lasso                          & 1.03 & 0.87 & 0.95 & 0.80 & -0.08 &  -8.0\% & -0.08 &  -8.7\% \\
Ridge                          & 1.09 & 0.92 & 1.02 & 0.85 & -0.07 &  -6.4\% & -0.07 &  -7.2\% \\
Elastic Net                    & 1.04 & 0.88 & 0.95 & 0.79 & -0.09 &  -8.8\% & -0.08 &  -9.5\% \\
Quantile Linear Regression     & 0.93 & 0.77 & 0.90 & 0.74  & -0.03 &  -3.7\% & -0.03 & -4.5\% \\
Principal Components (PCR)     & 1.09 & 0.92 & 0.99 & 0.82 & -0.09 &  -8.6\% & -0.09 & -10.3\% \\
Factor Models (FAR)            & 0.92 & 0.75 & 0.88 & 0.72 & -0.04 &  -4.4\% & -0.03 &  -3.6\% \\
Gradient Boosting              & 1.00 & 0.81 & 0.85 & 0.67 & -0.14 & -14.2\% & -0.14 & -17.9\% \\
Bagging                        & 1.03 & 0.83 & 0.84 & 0.67 & -0.19 & -18.3\% & -0.17 & -20.1\% \\
Random Forest                  & 0.99 & 0.75 & 0.82 & \cellcolor{green!20}0.65 & -0.17 & -17.1\% & -0.11 & \cellcolor{green!20}-14.3\% \\
Extremely Randomized Trees     & 0.98 & 0.74 & 0.80 & \cellcolor{green!20}0.60 & -0.18 & -18.5\% & -0.14 & \cellcolor{green!20}-19.0\% \\
Multilayer Random Forest (1S)  & 0.97 & 0.75 & 0.84 & \cellcolor{green!20}0.62 & -0.13 & -13.6\% & -0.13 & \cellcolor{green!20}-17.2\% \\
Multilayer Random Forest (2S)  & 1.01 & 0.77 & 0.85 & \cellcolor{green!20}0.65 & -0.16 & -15.9\% & -0.12 &\cellcolor{green!20} -15.5\% \\
Shallow CNN                    & 1.05 & 0.84 & 0.97 & 0.77 & -0.09 &  -8.3\% & -0.07 &  -8.8\% \\
Deep CNN                       & 1.04 & 0.77 & 0.89 & \cellcolor{green!20}0.66 & -0.15 & -14.1\% & -0.11 & \cellcolor{green!20}-14.3\% \\
\hline
\end{tabular}}
}
\begin{flushleft}\scriptsize
Notes: Out-of-sample RMSE and MAE for one-day-ahead forecasts of 
standardized sovereign CDS spreads under two information sets: 
Markets-Only (VIX and U.S.\ two-year Treasury yield) and Markets+News 
(augmented with GPR, EPU, TPU, ECO, INT, POL). All variables are 
28-day moving averages. Models are estimated recursively with a 28-day 
exclusion buffer. ``Diff'' denotes the change relative to the 
Markets-Only benchmark; negative values indicate lower forecast loss.
\end{flushleft}
\end{table}
Table~\ref{tab:mae_rmse} reports out-of-sample forecast accuracy under 
the Markets-Only and Markets+News information sets. Three findings stand out. Flexible nonlinear methods outperform linear specifications, 
with Extremely Randomized Trees achieving the lowest errors 
(MAE\,=\,0.60). News variables improve every model class, but the 
gains are substantially larger for nonlinear methods (15--19\% RMSE 
reduction) than for linear specifications (5--9\%), indicating that 
much of the predictive content of news enters through interactions and 
threshold effects that linear frameworks understate 
\citep{Varian2014, Gu2020}. The magnitude of improvement rises 
monotonically with model flexibility 
\citep{Mullainathan2017, Athey2019}.

Although Extremely Randomized Trees delivers the lowest raw forecast 
loss, the remainder of the paper uses the Multilayer Random Forest 
(Two Stages) as the baseline for interpretation. Its performance 
remains close to the frontier, while its architecture preserves the 
advanced-economy/emerging-market split and region-specific structure 
that are central to the subsequent decomposition.\footnote{The 
Multilayer Random Forest extends the standard Random Forest by 
stacking two sequential ensemble layers \citep{Breiman2001}. The 
first layer is trained separately on advanced- and emerging-economy 
panels; the second combines intermediate predictions with the full set 
of covariates.} 

\subsection{Sovereign Risk Drivers}
\label{subsec:drivers_shap}

\begin{figure}[ht!]
    \centering
    \caption{\textbf{Global Financial Conditions Dominate Sovereign 
    Risk: Shapley Feature Importance}}
    \vspace{2pt}
    \includegraphics[width=0.95\textwidth]{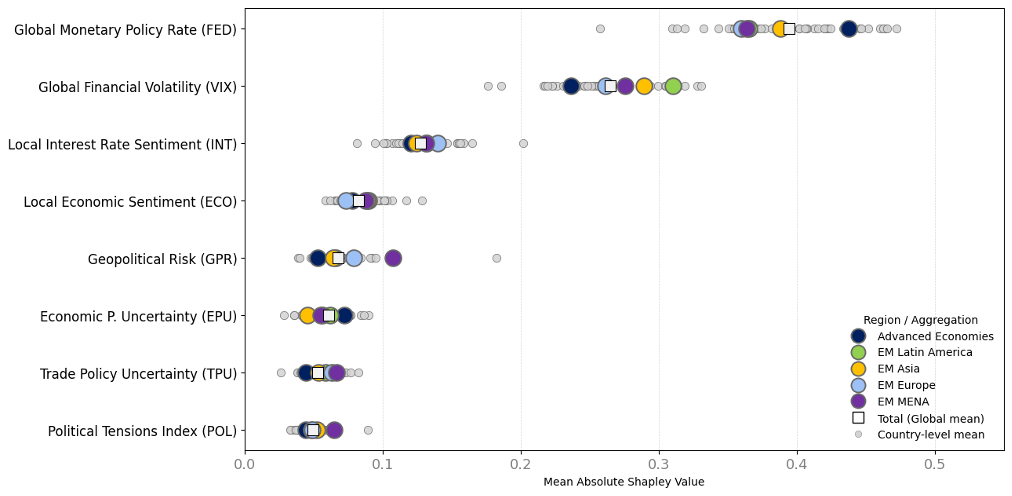}
    \label{fig:shapglobal}
    \begin{flushleft}
       \scriptsize
        Notes: Mean absolute Shapley values for all predictors over 
        2018--2025, shown at the global, regional, and country levels. 
        Country-level values are computed by averaging over all 
        observations for each country; the global value is the 
        unweighted mean across country means.
    \end{flushleft}
\end{figure}

Figure~\ref{fig:shapglobal} summarizes the fitted model's feature 
importance using mean absolute Shapley values at the global, regional, 
and country levels.

The Shapley ranking points to a layered structure. The first layer is 
global. The U.S.\ two-year yield and the VIX are the largest 
contributors in almost every region and for nearly every country, with 
mean absolute Shapley values exceeding those of any other variable---
consistent with the view that sovereign spreads are anchored by common 
financial conditions \citep{rey2013, MirandaAgrippinoRey2020}, with 
the VIX corroborating its role as a barometer of global risk aversion 
whose spikes precede capital-flow reversals and spread widening 
\citep{Gelos2011}. 

The second layer is domestic. Local interest-rate sentiment (INT) and 
economic sentiment (ECO) rank just below the global variables but 
display much greater cross-country dispersion, indicating that 
country-specific narratives are critical for differentiating sovereign 
spreads conditional on the global environment 
\citep{Eichengreen2021a}.

The third layer is conditional. 
Geopolitical and policy-uncertainty indicators---GPR, EPU, and 
TPU---are smaller on average but become much more prominent in EM 
Europe and MENA, where exposure to geopolitical and geoeconomic risk 
is greater \citep{baker2016, CaldaraEtAl2020TPU}, foreshadowing the 
state-dependent nonlinearities documented below.

% ─────────────────────────────────────────────────────────────────────────────
\subsection{Nonlinear Interactions and State Dependence}
\label{subsec:interactions}

Pairwise Shapley--Taylor interaction values show that global financial 
conditions are not only the largest main effects but also the dominant 
sources of nonlinear interaction. The largest and most pervasive 
interaction terms involve US2Y and the VIX, as well as interactions 
between US2Y and domestic interest-rate sentiment. By contrast, GPR 
and EPU rarely appear as large stand-alone drivers across the full 
sample, but they interact strongly with both global and domestic 
variables in EM Europe and MENA while remaining much weaker 
elsewhere. In the fitted model, geopolitical and policy-uncertainty 
variables matter primarily by changing how sovereign risk responds to 
the broader financial environment. This pattern provides a direct 
motivation for the channel grouping introduced in 
Section~\ref{sec:shapley_interpretability}, where the global financial 
cycle block ($\mathcal{G}$) is defined as the set of variables that 
generate the strongest and most pervasive interactions.

\begin{figure}[ht!]
    \setlength{\belowcaptionskip}{-5pt}
    \centering
    \caption{\textbf{State Dependence in Sovereign Risk: Two-Factor 
    Interaction Surfaces}}
    \vspace{2pt}
    \includegraphics[width=1\textwidth]{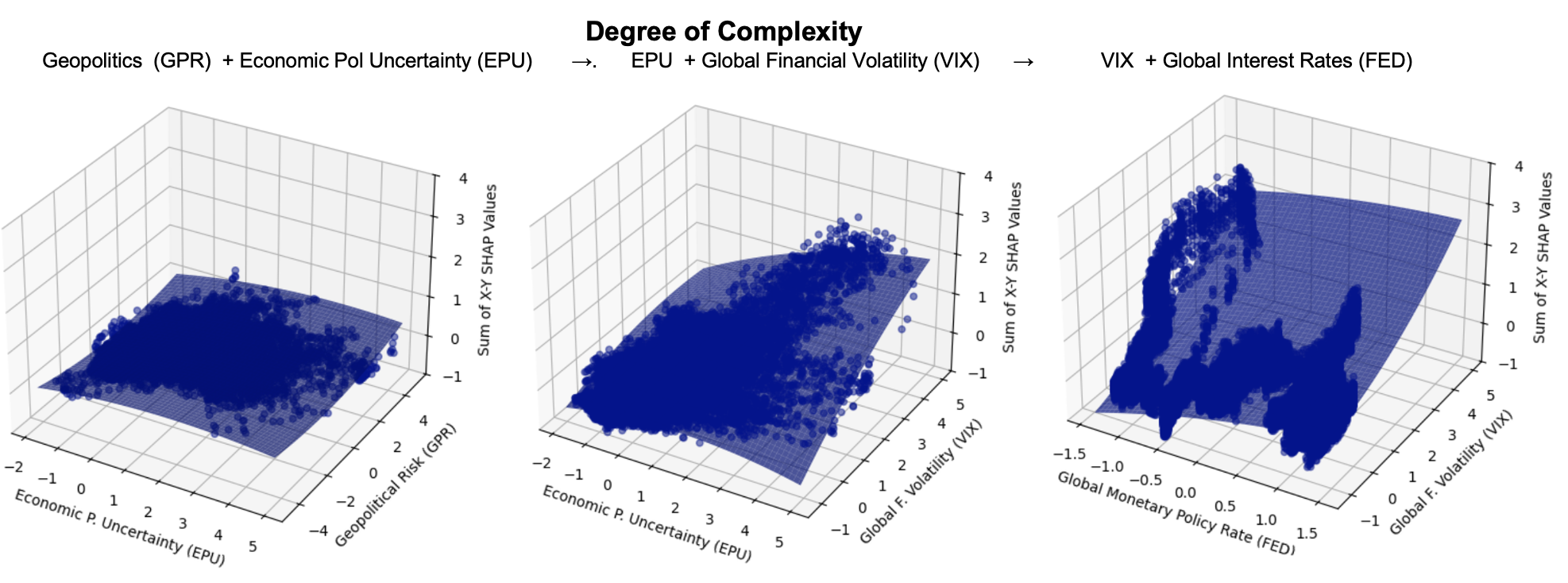}
    \label{fig:shap3d}
    \begin{flushleft}
       \scriptsize
        \vspace{-5pt}
        Notes: The figure illustrates the combined impact of key 
        predictors on sovereign CDS spreads for Advanced Economies 
        using two-factor SHAP dependence surfaces. Left panel: joint 
        influence of Geopolitical Risk (GPR) and Economic Policy 
        Uncertainty (EPU)---the surface is nearly flat, indicating 
        low sensitivity when financial conditions are uncontrolled. 
        Centre panel: joint contribution of EPU and Global Financial 
        Volatility (VIX)---high uncertainty in a high-VIX 
        environment is substantially more detrimental than either 
        shock alone \citep{Bloom2009}. Right panel: joint effect of 
        VIX and global monetary policy (US2Y)---the largest and most 
        nonlinear increases in sovereign risk arise when volatility 
        spikes coincide with tight monetary conditions 
        \citep{rey2013, BrunoShin2015}, but are partially contained 
        under accommodative policy \citep{Bekaert2013}. See 
        Section~\ref{subsec:interactions} for discussion.
    \end{flushleft}
\end{figure}

The interaction surfaces reinforce this state dependence 
(Figure~\ref{fig:shap3d}). When global financial conditions 
are benign, the joint contribution of geopolitical risk and policy 
uncertainty is relatively muted in most regions. By contrast, 
uncertainty becomes materially more important in high-volatility 
states \citep{Bloom2009}, and the strongest nonlinearities arise when 
elevated volatility coincides with tight U.S.\ rates: volatility 
shocks are powerfully amplified under tight monetary policy 
\citep{rey2013, BrunoShin2015} but partially contained under 
accommodative conditions \citep{Bekaert2013}. Geoeconomic or policy 
uncertainty, while often manageable in isolation, thus becomes a 
potent threat when it coincides with financial stress---a finding that 
motivates the four-channel decomposition developed next.

Cross-country connectedness measures applied to the panel of daily 
Shapley attributions confirm that 
global financial variables are the dominant transmitters of 
cross-country co-movement, while geopolitical shocks generate episodic 
synchronization that dissipates unless reinforced by tighter financial 
conditions.

\section{Transmission Anatomy of Geopolitical and Geoeconomic Shocks}
\label{sec:transmission}

Applying the four-channel decomposition, we begin by inspecting the raw attribution dynamics around each episode. Country-level Shapley paths around the Russia--Ukraine invasion (Appendix Figure~\ref{fig:episode_RU}) are consistent with staged propagation from geopolitical news to energy-related macro pressures and then to interactions with global financial conditions; the Hamas--Israel war appears more regionally contained; and the U.S.\ election and tariff episodes transmit less through the Direct channel and more through policy uncertainty, trade exposure, and global financial conditions.

These raw attributions are informative but combine several mechanisms at once. To separate them, we apply the four-channel Shapley--Taylor decomposition to realized episode data, feeding observations through the estimated model---held fixed at its pre-episode parameters---and tracking daily channel contributions for the 42-country panel.\footnote{Country-level Shapley attribution dynamics for each episode are reported in Appendix~\ref{app:Transmission}.} Three features organize the results. First, the configuration of channels differs systematically across geopolitical and geoeconomic episodes. Second, the cross-sectional distribution of each channel follows a distinct geographic or economic pattern. Third, the channels decay at different speeds. The remainder of this section develops these results in turn.

% ─────────────────────────────────────────────────────────────────────────────
\subsection{The Scissors Pattern}
\label{sec:scissors}

\begin{figure}[ht!]
    \setlength{\belowcaptionskip}{-5pt}
    \centering
    \caption{\textbf{Geopolitical vs.\ Geoeconomic Shocks: The Scissors Pattern}}
    \label{fig:episode_tariffs}
    \label{fig:episode_HI}
    \vspace{2pt}
    \includegraphics[width=0.9\textwidth]{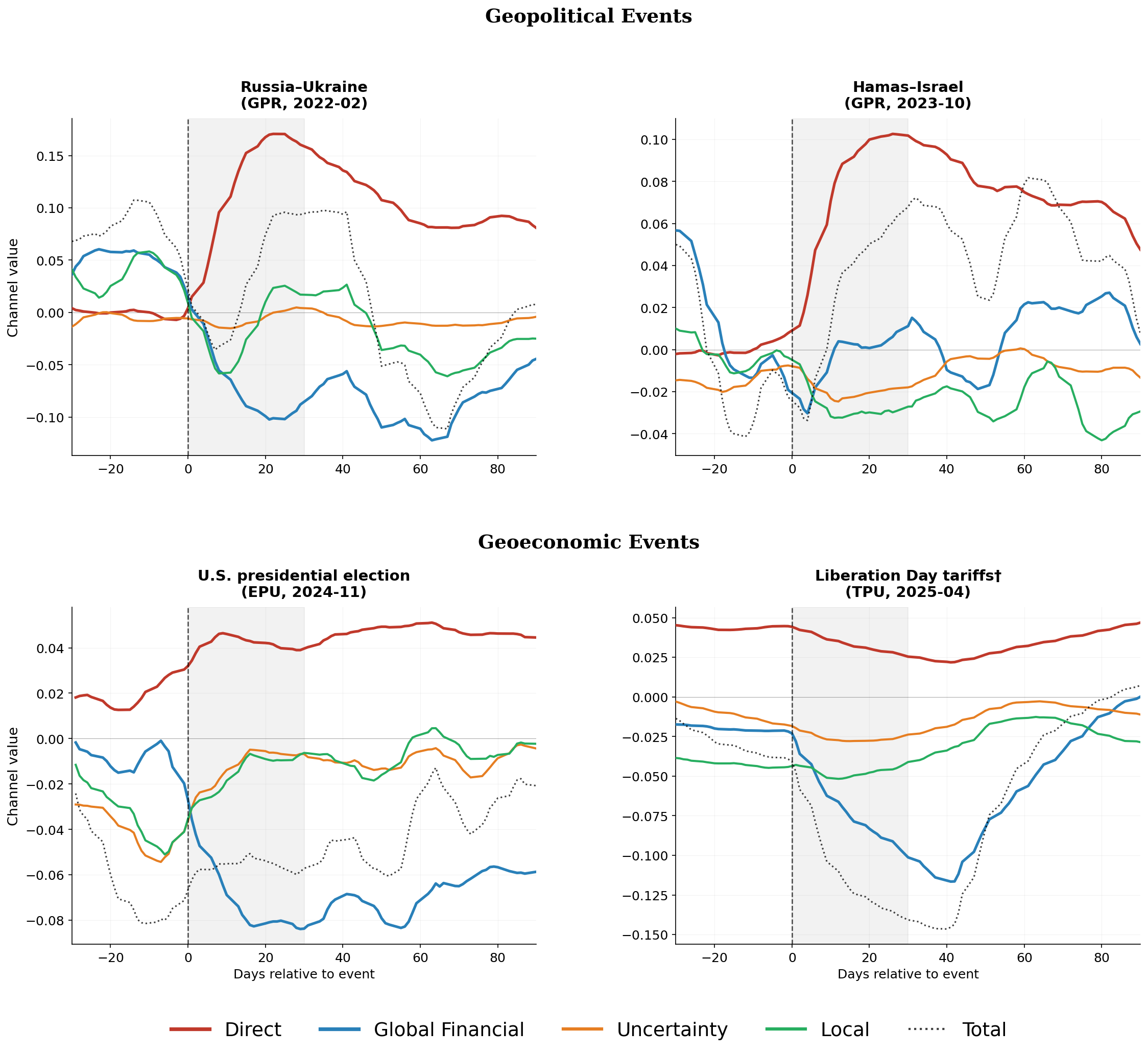}
    \label{fig:channels}
    \begin{flushleft}
       \scriptsize
        \vspace{-5pt}
        Notes: Each panel plots the cross-sectional average (42 countries) of the four Shapley–Taylor transmission channels for the indicator most directly associated with each episode: GPR for geopolitical shocks (top row); EPU/TPU for geoeconomic shocks (bottom row). Direct (red): own Shapley–Taylor term $\phi^\mathrm{dir}$. Global Financial (blue): interactions with VIX and US2Y. Uncertainty (orange): interactions with EPU and TPU. Local (green): interactions with ECO, INT, POL, and REG. Dotted black line: total Shapley contribution. Vertical dashed lines mark the event date. All series are 7-day moving averages with a trailing smoother.
    \end{flushleft}
\end{figure}

Figure~\ref{fig:channels} plots the cross-sectional average of the four Shapley--Taylor channels across all 42 countries for each episode. Geopolitical episodes generate a scissors pattern between the Direct and GFC channels, whereas geoeconomic episodes display limited direct repricing and transmit mainly through financial and uncertainty channels.

\paragraph{Geopolitical episodes.}
In panel~(a), the Russia--Ukraine invasion opens a clear ``scissors'': the Direct channel (red) jumps to roughly $+0.10$ within days as conflict exposure reprices default probability \citep{CaldaraIacoviello2022}, while the GFC channel (blue) drops to approximately $-0.05$ as the global financial system partially absorbs the shock through the risk-taking channel \citep{rey2013, BrunoShin2015}. The two move in opposite directions, so the total Shapley contribution is slightly negative at the panel mean ($\varphi^{\mathrm{tot}}_{1m} = -0.07$)---the GFC and Local channels more than offset the Direct component, implying that bystander sovereigns experience net spread compression even as conflict-exposed countries are sharply repriced.
Panel~(b) shows a similar but more muted pattern for Hamas--Israel: the Direct channel spikes, but the GFC response is much smaller in magnitude, consistent with a conflict whose spillovers remain more regionally contained. In both geopolitical episodes, the Uncertainty and Local channels are comparatively muted in cross-country mean terms at impact, even though country-level responses can be substantial for exposed sovereigns.

\paragraph{Geoeconomic episodes: transmission through financial and uncertainty channels.}
The geoeconomic episodes invert the picture. In panel~(c), the Direct channel remains economically negligible throughout---the U.S.\ election does not reprice default probability. Instead, the GFC channel begins declining three months \emph{before} the event as markets price in looser monetary policy \citep{MirandaAgrippinoRey2020}, and continues falling to roughly $-0.08$ by December, while the Uncertainty channel also contributes positively. Panel~(d) is the sharpest illustration: the GFC plunges from near zero to approximately $-0.12$ within two weeks of the April~2025 tariff announcement. The Direct channel remains near zero---the tariff shock shifts expected monetary policy and risk appetite, not default probability.

\paragraph{Quantitative summary.}
Table~\ref{tab:impact_persistence} summarizes these patterns at one- and three-month horizons. The Direct channel exceeds the GFC at the panel level in both geopolitical episodes ($\varphi^{\mathrm{dir}}_{1\mathrm{m}} = +0.11$ vs.\ $\varphi^{\mathrm{GFC}}_{1\mathrm{m}} = -0.11$ for Russia--Ukraine; $+0.08$ vs.\ $-0.01$ for Hamas--Israel); the GFC dominates the Direct channel in both geoeconomic episodes. Taken together, geopolitical shocks load primarily on the Direct channel while provoking a partial GFC offset; geoeconomic shocks largely bypass the Direct channel and transmit through financial and uncertainty channels. Appendix Tables~\ref{tab:Geo_Country_combined} and~\ref{tab:Econ_Country_combined} report the complete country-level decomposition for all 42 economies, confirming that these patterns hold broadly across the panel.

% ─────────────────────────────────────────────────────────────────────────────
\subsection{Country Heterogeneity}
\label{sec:heterogeneity}

The scissors pattern describes the average response across countries. At the country level, transmission depends on the relative importance of two additional forces: exposure to the global financial cycle ($\varphi^{\mathrm{GFC}}$) and the strength of domestic amplification ($\varphi^{\mathrm{LOC}}$). By construction, these channels depend on different sets of predictors and need not move together.

\begin{table}[htbp]
\centering
\caption{Four-Channel Decomposition across Crisis Episodes: Impact and Persistence}
\label{tab:impact_persistence}
\footnotesize
\setlength{\tabcolsep}{3pt}
\begin{tabular}{@{}l
    >{\centering\arraybackslash}p{2.3cm}
    rrrrr @{\quad}
    rrrrr@{}}
\toprule
 & &
 \multicolumn{5}{c}{\textsc{Short-Term Impact} ($\Delta_{1\mathrm{m}}$)} &
 \multicolumn{5}{c}{\textsc{Persistence} ($\Delta_{3\mathrm{m}}$)} \\
\cmidrule(lr){3-7}\cmidrule(lr){8-12}
 & \makecell[c]{\scriptsize Nature of\\\scriptsize Shock}
 & $\varphi^{\mathrm{dir}}$ & $\varphi^{\mathrm{GFC}}$ & $\varphi^{\mathrm{UNC}}$
 & $\varphi^{\mathrm{LOC}}$ & $\varphi^{\mathrm{tot}}$
 & $\varphi^{\mathrm{dir}}$ & $\varphi^{\mathrm{GFC}}$ & $\varphi^{\mathrm{UNC}}$
 & $\varphi^{\mathrm{LOC}}$ & $\varphi^{\mathrm{tot}}$ \\
\midrule
%──────────────────────────────────────────────────────────────────────────────
\multicolumn{12}{@{}l}{\textit{Panel A\,: Russia--Ukraine (GPR, Feb.\ 2022)}} \\[3pt]
\quad \textit{Panel avg.}
  & \multirow{4}{2.3cm}{\centering\scriptsize\textbf{Geopolitics}\\[2pt]\scriptsize(Panel A)}
  & \cellcolor{red!15}+.10 & \cellcolor{green!15}-.12 & \cellcolor{gray!10}\phantom{+}.00
  & \cellcolor{green!8}-.03  & \cellcolor{green!8}-.04
  & \cellcolor{red!15}+.07 & \cellcolor{green!15}-.09 & \cellcolor{gray!10}\phantom{+}.00
  & \cellcolor{green!8}-.03  & \cellcolor{green!15}-.05 \\
\quad Ukraine &
  & \cellcolor{red!15}+.13 & \cellcolor{red!15}+.66 & \cellcolor{green!8}-.02
  & \cellcolor{red!15}+.48  & \cellcolor{red!15}+1.25
  & \cellcolor{red!15}+.13 & \cellcolor{red!15}+.66 & \cellcolor{green!8}-.02
  & \cellcolor{red!15}+.48  & \cellcolor{red!15}+1.25 \\
\quad Poland &
  & \cellcolor{red!15}+.15 & \cellcolor{green!15}-.10 & \cellcolor{red!8}+.04
  & \cellcolor{red!15}+.06  & \cellcolor{red!15}+.14
  & \cellcolor{red!15}+.11 & \cellcolor{green!8}-.02 & \cellcolor{red!8}+.04
  & \cellcolor{red!15}+.06  & \cellcolor{red!15}+.20 \\
\quad Germany &
  & \cellcolor{red!15}+.25 & \cellcolor{green!15}-.24 & \cellcolor{red!8}+.02
  & \cellcolor{red!15}+.06  & \cellcolor{red!15}+.09
  & \cellcolor{red!15}+.15 & \cellcolor{green!15}-.11 & \cellcolor{red!8}+.01
  & \cellcolor{red!8}+.05   & \cellcolor{red!15}+.10 \\[6pt]
%──────────────────────────────────────────────────────────────────────────────
\multicolumn{12}{@{}l}{\textit{Panel B\,: Hamas--Israel (GPR, Oct.\ 2023)}} \\[3pt]
\quad \textit{Panel avg.}
  & \multirow{4}{2.3cm}{\centering\scriptsize\textbf{Geopolitics}\\[2pt]\scriptsize(Panel B)}
  & \cellcolor{red!15}+.08 & \cellcolor{green!8}-.01 & \cellcolor{red!8}+.01
  & \cellcolor{green!8}-.02  & \cellcolor{red!15}+.06
  & \cellcolor{red!15}+.06 & \cellcolor{green!8}-.02 & \cellcolor{gray!10}\phantom{+}.00
  & \cellcolor{green!8}-.01  & \cellcolor{red!8}+.02 \\
\quad Israel &
  & \cellcolor{red!15}+.21 & \cellcolor{red!15}+.32 & \cellcolor{red!15}+.19
  & \cellcolor{red!15}+.30  & \cellcolor{red!15}+1.02
  & \cellcolor{red!15}+.21 & \cellcolor{red!15}+.34 & \cellcolor{red!15}+.16
  & \cellcolor{red!15}+.33  & \cellcolor{red!15}+1.04 \\
\quad Egypt &
  & \cellcolor{red!15}+.27 & \cellcolor{red!15}+.08 & \cellcolor{red!15}+.08
  & \cellcolor{red!15}+.07  & \cellcolor{red!15}+.51
  & \cellcolor{red!15}+.21 & \cellcolor{green!15}-.07 & \cellcolor{red!8}+.01
  & \cellcolor{green!15}-.06 & \cellcolor{red!15}+.09 \\
\quad Germany &
  & \cellcolor{red!15}+.09 & \cellcolor{green!15}-.15 & \cellcolor{green!8}-.02
  & \cellcolor{red!8}+.02   & \cellcolor{green!15}-.06
  & \cellcolor{red!15}+.09 & \cellcolor{green!15}-.18 & \cellcolor{green!8}-.04
  & \cellcolor{gray!10}\phantom{+}.00 & \cellcolor{green!15}-.13 \\[6pt]
%──────────────────────────────────────────────────────────────────────────────
\multicolumn{12}{@{}l}{\textit{Panel C\,: U.S.\ Presidential Election (EPU, Nov.\ 2024)}} \\[3pt]
\quad \textit{Panel avg.}
  & \multirow{4}{2.3cm}{\centering\scriptsize\textbf{Economic}\\[-1pt]\scriptsize\textbf{Uncertainty}\\[2pt]\scriptsize(Panel C)}
  & \cellcolor{red!8}+.03 & \cellcolor{green!15}-.06 & \cellcolor{red!8}+.03
  & \cellcolor{red!8}+.01  & \cellcolor{red!8}+.01
  & \cellcolor{red!8}+.04 & \cellcolor{green!15}-.06 & \cellcolor{red!8}+.03
  & \cellcolor{red!8}+.01  & \cellcolor{red!8}+.01 \\
\quad USA &
  & \cellcolor{red!8}+.04 & \cellcolor{green!15}-.08 & \cellcolor{red!8}+.01
  & \cellcolor{red!8}+.02  & \cellcolor{green!8}-.01
  & \cellcolor{red!15}+.06 & \cellcolor{green!15}-.10 & \cellcolor{green!8}-.02
  & \cellcolor{red!8}+.02  & \cellcolor{green!8}-.04 \\
\quad Mexico &
  & \cellcolor{red!8}+.03 & \cellcolor{green!8}-.03 & \cellcolor{red!8}+.01
  & \cellcolor{gray!10}\phantom{+}.00 & \cellcolor{red!8}+.01
  & \cellcolor{red!8}+.05 & \cellcolor{green!15}-.06 & \cellcolor{green!8}-.03
  & \cellcolor{green!8}-.04 & \cellcolor{green!15}-.08 \\
\quad Germany &
  & \cellcolor{red!8}+.04 & \cellcolor{gray!10}\phantom{+}.00 & \cellcolor{red!15}+.06
  & \cellcolor{red!8}+.04  & \cellcolor{red!15}+.15
  & \cellcolor{red!8}+.05 & \cellcolor{green!8}-.04 & \cellcolor{red!8}+.05
  & \cellcolor{red!8}+.04  & \cellcolor{red!15}+.10 \\[6pt]
%──────────────────────────────────────────────────────────────────────────────
\multicolumn{12}{@{}l}{\textit{Panel D\,: U.S.\ Tariffs / Liberation Day (TPU, Apr.\ 2025)}} \\[3pt]
\quad \textit{Panel avg.}
  & \multirow{4}{2.3cm}{\centering\scriptsize\textbf{Trade}\\[-1pt]\scriptsize\textbf{Uncertainty}\\[2pt]\scriptsize(Panel D)}
  & \cellcolor{green!8}-.01 & \cellcolor{green!8}-.01 & \cellcolor{red!8}+.01
  & \cellcolor{red!8}+.02  & \cellcolor{red!8}+.01
  & \cellcolor{green!8}-.01 & \cellcolor{green!8}-.01 & \cellcolor{red!8}+.01
  & \cellcolor{red!8}+.02  & \cellcolor{red!8}+.01 \\
\quad USA &
  & \cellcolor{green!8}-.01 & \cellcolor{green!8}-.02 & \cellcolor{green!8}-.03
  & \cellcolor{red!15}+.10 & \cellcolor{red!8}+.03
  & \cellcolor{green!8}-.01 & \cellcolor{green!8}-.02 & \cellcolor{green!8}-.03
  & \cellcolor{red!15}+.10 & \cellcolor{red!8}+.03 \\
\quad India &
  & \cellcolor{green!8}-.02 & \cellcolor{green!15}-.08 & \cellcolor{green!8}-.03
  & \cellcolor{red!8}+.02  & \cellcolor{green!15}-.10
  & \cellcolor{green!8}-.02 & \cellcolor{green!15}-.08 & \cellcolor{green!8}-.03
  & \cellcolor{red!8}+.02  & \cellcolor{green!15}-.10 \\
\quad Germany &
  & \cellcolor{green!8}-.01 & \cellcolor{green!8}-.05 & \cellcolor{gray!10}\phantom{+}.00
  & \cellcolor{green!8}-.01 & \cellcolor{green!15}-.07
  & \cellcolor{green!8}-.01 & \cellcolor{green!8}-.05 & \cellcolor{gray!10}\phantom{+}.00
  & \cellcolor{green!8}-.01 & \cellcolor{green!15}-.07 \\
\addlinespace[3pt]
\bottomrule
\end{tabular}

\vspace{4pt}
\begin{minipage}{0.98\textwidth}
\scriptsize
\textit{Notes:} Each cell reports $\Delta \equiv \bar{\varphi}^{\mathrm{post}} - \bar{\varphi}^{\mathrm{pre}}$
in standardized CDS units.
\textsc{Short-Term Impact}: post-window = 1\,month after event onset.
\textsc{Persistence}: post-window = 3\,months after event onset.
Pre-event window = 3\,months in both cases.
Cell shading:
\colorbox{red!15}{\phantom{x}} $\Delta>+0.05$,\;
\colorbox{red!8}{\phantom{x}} $0.005<\Delta\leq+0.05$,\;
\colorbox{gray!10}{\phantom{x}} $|\Delta|\leq0.005$,\;
\colorbox{green!8}{\phantom{x}} $-0.05\leq\Delta<-0.005$,\;
\colorbox{green!15}{\phantom{x}} $\Delta<-0.05$.
Values in $\sigma$-CDS units.
\textit{Panel avg.}\ = unweighted cross-country mean (42~countries).
Germany appears in all panels as a common benchmark.
28-day moving averages applied to all series.
$\varphi^{\mathrm{LOC}}$ includes interactions with local macro-political variables and regional indicators
(ECO, INT, POL, REG; see equation~\ref{eq:channels}).
\end{minipage}
\end{table}

Table~\ref{tab:impact_persistence} illustrates three distinct transmission modes within the geopolitical episodes. \textit{Ukraine's} response is dominated by the GFC channel: $\varphi^{\mathrm{GFC}}_{1\mathrm{m}} = +0.67$, the largest single-channel value in the sample. Sanctions and market exclusion converted a geopolitical shock into a severe financial-intermediation event \citep{FVFragmented2024}, compounded by collapsing domestic sentiment ($\varphi^{\mathrm{LOC}} = +0.62$). \textit{Israel} presents a different balance: $\varphi^{\mathrm{GFC}} = +0.35$ and $\varphi^{\mathrm{LOC}} = +0.46$ contribute with broadly similar magnitudes, consistent with the joint activation of global risk-appetite sensitivity and domestic amplifiers \citep{Gorodnichenko2025}. The contrast is economically meaningful: Ukraine's crisis is primarily one of financial exclusion; Israel's reflects domestic strain in equal measure. \textit{Poland} illustrates partial GFC compensation: the Direct-channel rise of $+0.18$ is materially offset by $\varphi^{\mathrm{GFC}} = -0.15$, consistent with the idea that deep integration into European capital markets can cushion part of the initial shock.

The geoeconomic episodes reveal a different asymmetry. At the panel level, the tariff shock compresses spreads through the GFC as easier expected global financial conditions benefit most sovereigns. But the United States---the originator---is a partial exception. Its GFC contribution is a muted $-0.02$, while its Local channel registers $+0.10$: the policy uncertainty the originator created raises its own risk premia through the interaction of TPU with domestic sentiment and political tensions. The originator does not receive the same financial offset as the rest of the panel---consistent with evidence that geoeconomic instruments generate domestic risk premia \citep{Clayton2025} and with the theoretical prediction of \citet{PastorVeronesi2013} that uncertainty-generated risk premia are largest for the source country.

% ─────────────────────────────────────────────────────────────────────────────
\subsection{Differential Persistence}
\label{sec:persistence}

If the Direct channel is associated with more persistent sovereign-specific repricing, while the GFC channel reflects more transitory financial adjustment, the two should fade at different speeds. Table~\ref{tab:impact_persistence} is consistent with that prediction. The Direct channel retains roughly two-thirds of its one-month value at three months in both geopolitical episodes, while the GFC decays faster and can reverse sign \citep{Bekaert2013, Bloom2009}.

Egypt illustrates this temporal separation. At impact, all four channels contribute positively ($\varphi^{\mathrm{tot}}_{1\mathrm{m}} = +0.56$). By three months, the GFC has reversed ($+0.07 \to -0.08$) and the Local channel has followed, pulling the total to $+0.10$. But the Direct channel barely moved ($+0.27 \to +0.21$): Egypt's fundamental exposure persisted even after the financial amplifiers faded.

%% ─── TABLE ───────────────────────────────────────────────────────────────────

% ─────────────────────────────────────────────────────────────────────────────
\subsection{Cross-Sectional Fingerprints}
\label{sec:fingerprints}

The preceding subsections established that different shock types activate different channels over time. We now show that each activated channel also exhibits a distinct \textit{cross-sectional} distribution, completing the transmission taxonomy.

\begin{figure}[ht!]
    \setlength{\belowcaptionskip}{-5pt}
    \centering
    \caption{\textbf{Direct Geopolitical Effects Decline with Distance from Conflict}}
    \vspace{2pt}
    \includegraphics[width=1\textwidth]{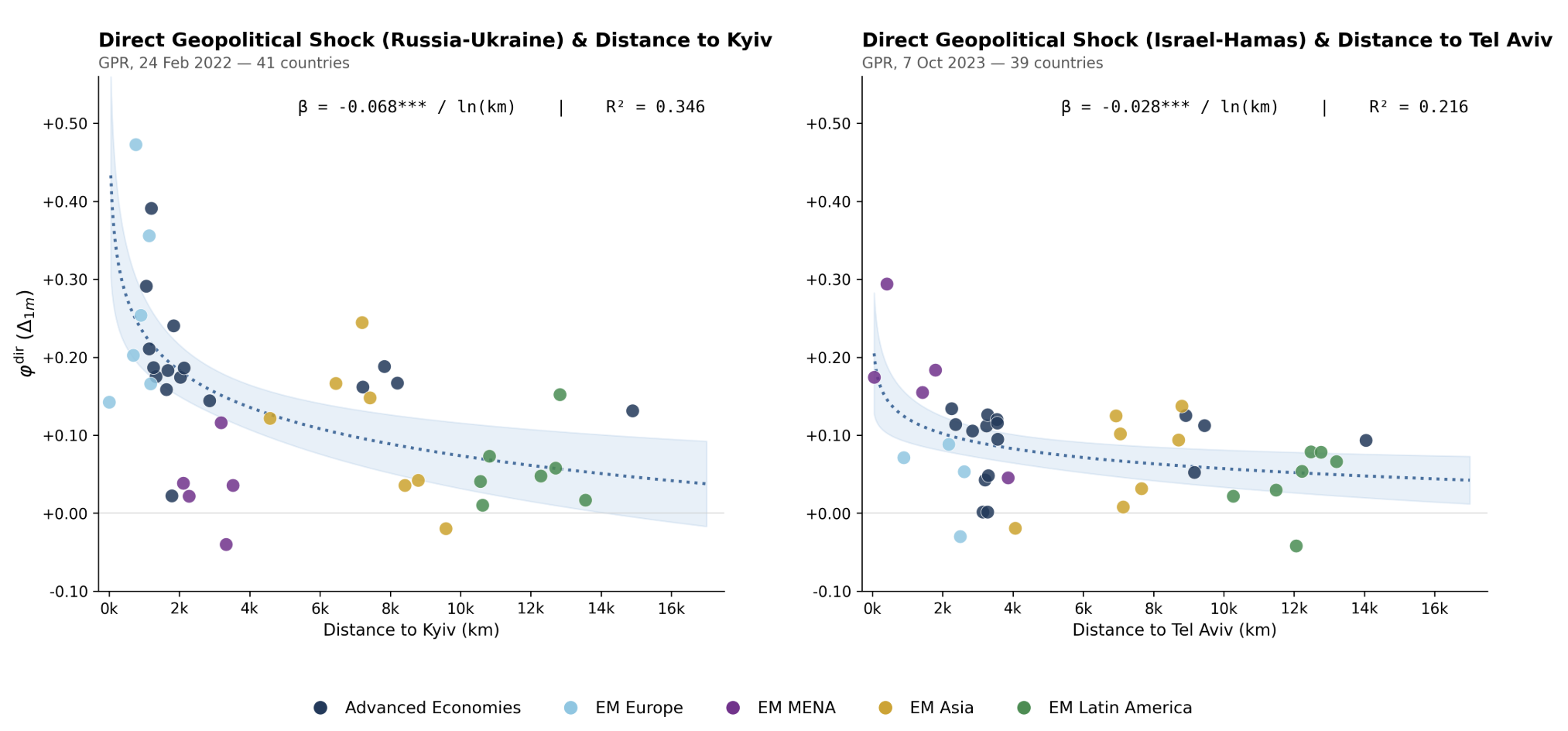}
    \label{fig:gravity}
    \begin{flushleft}
       \scriptsize
        \vspace{-5pt}
        Notes: Each panel plots the one-month Direct-channel response $\varphi^{\mathrm{dir}}$ ($\Delta_{1\mathrm{m}}$) against great-circle distance (km) from the national capital to the conflict epicenter (Kyiv, left panel; Tel~Aviv, right panel). The decomposition follows equation~(\ref{eq:channels}). Left panel: 41~sovereigns (Russia--Ukraine, 24~February~2022; Ukraine at $d=0$ plotted but excluded from log-distance OLS). Right panel: 39~sovereigns (Hamas--Israel, 7~October~2023). Dotted lines: OLS fitted values from equation~(\ref{eq:gravity_ols}); shaded bands: 95\% confidence intervals. All values in within-country standard deviations of CDS spreads ($\sigma$-CDS). Jordan excluded due to insufficient pre-2023 CDS data. $^{***}\,p<0.01$, $^{**}\,p<0.05$, $^{*}\,p<0.10$.
    \end{flushleft}
\end{figure}

\paragraph{Geopolitical shocks: gravity-like distance decay.}
If geopolitical shocks reprice default probability in proportion to conflict exposure, countries closer to the conflict should register larger Direct-channel effects---a gravity-style prediction familiar from the trade literature \citep{HeadMayer2014} and from work linking geographic proximity to financial contagion \citep{GlickRose1999}. Figure~\ref{fig:gravity} evaluates this prediction by regressing the short-term direct effect on log great-circle distance:
\begin{equation}
  \varphi^{\mathrm{dir}}_{i}
  \;=\; \alpha \;+\; \beta\,\ln d_{i} \;+\; u_{i}\,,
  \label{eq:gravity_ols}
\end{equation}
where $d_{i}$ is distance to Kyiv (left panel) or Tel~Aviv (right panel).

The Russia--Ukraine shock produces a steep, concave distance gradient
($\hat{\beta} = -0.068^{***}$ per log-km; $R^{2} = 0.35$, $n = 40$).
Countries within 1{,}200\,km of Kyiv---Czech Republic
($+0.32$), Germany ($+0.28$), Austria ($+0.23$)---register Direct-channel responses two to
three times the cross-sectional mean ($+0.11$); most Latin American
sovereigns beyond 10{,}000\,km fall below $+0.08$, though notable
exceptions such as Argentina ($+0.15$) and Australia ($+0.13$) suggest
that non-geographic linkages also play a role. The Hamas--Israel shock
reveals a different geography: the slope is roughly half as steep
($\hat{\beta} = -0.028^{***}$; $R^{2} = 0.22$, $n = 39$), with the
four largest responses all belonging to MENA sovereigns within
2{,}000\,km---Egypt ($+0.27$), Qatar ($+0.20$), Israel ($+0.19$),
Saudi Arabia ($+0.14$). Both coefficients are significant at 1\% under
the log specification, which captures the concave decay
pattern---steep attenuation near the epicenter, flattening at greater
distances---better than the linear alternative. The cross-sectional
footprint of the Direct channel is gravity-like in both cases, but
much more pronounced for the more systemic of the two geopolitical
shocks.

Appendix Figure~\ref{fig:gravity_direct_vs_REG} confirms that the regional component of the Local channel shows no distance relationship ($R^{2} = 0.003$, $p = 0.72$), validating its absorption into LOC rather than creating a fifth channel.

\paragraph{Policy-uncertainty shocks: global and simultaneous.}
The election episode leaves a different signature. Its most distinctive feature is the broad-based activation of the Uncertainty channel $\varphi^{\mathrm{UNC}}$, which turns positive for a large share of countries at roughly the same time. At the panel level it registers $+0.03$ at both horizons, with a cross-country interquartile range of $[+0.01, +0.06]$; the appendix tables confirm positive values for France, Brazil, China, Germany ($+0.06$), and most other sovereigns.\footnote{The cross-country standard deviation of $\varphi^{\mathrm{UNC}}_{1\mathrm{m}}$ is $0.02$ for the election episode---roughly one-third of the corresponding dispersion for $\varphi^{\mathrm{dir}}$ in the Russia--Ukraine episode ($0.07$). The narrow spread confirms that the Uncertainty channel activates broadly rather than selectively.} This broad-based activation is much less visible in the other three episodes. In the tariff shock, $\varphi^{\mathrm{UNC}}$ is comparatively small; in the geopolitical episodes, it is limited mainly to the most exposed countries. The election episode therefore stands out not only by which channel activates, but also by how broadly and synchronously it does so---consistent with models in which political uncertainty generates correlated risk premia across all assets simultaneously \citep{PastorVeronesi2012, baker2016}.

\paragraph{A three-way fingerprint.}
Together, these results complete the cross-sectional dimension of the taxonomy. Geopolitical shocks activate $\varphi^{\mathrm{dir}}$ with gravity-like structure. Trade shocks operate through $\varphi^{\mathrm{GFC}}$ in proportion to GFC exposure. Policy-uncertainty shocks activate $\varphi^{\mathrm{UNC}}$ broadly across countries. Each shock type has a distinct first-responder channel \textit{and} a distinct cross-sectional distribution. This combination provides a useful empirical diagnostic for distinguishing the nature of a shock in real time.

% ─────────────────────────────────────────────────────────────────────────────
% CLOSING SUMMARY
% ─────────────────────────────────────────────────────────────────────────────

In sum, the transmission anatomy rests on four findings: (i)~geopolitical shocks display a scissors pattern while geoeconomic shocks load on financial and uncertainty channels; (ii)~country-level heterogeneity depends on GFC exposure and domestic amplification; (iii)~channels decay at different speeds; and (iv)~each shock type leaves a distinct cross-sectional fingerprint.

\section{Econometric Validation via Local Projections}
\label{sec:Causal_LP}

This section asks whether the four-channel taxonomy developed in 
Section~\ref{sec:transmission} receives separate econometric support 
in panel local projections under the two identification strategies 
introduced in Section~\ref{sec:causal_framework}. The logic is 
simple. If the proposed channels capture economically meaningful 
differences in transmission, then they should respond differently to 
identified shocks even outside the machine-learning decomposition 
stage. Table~\ref{tab:causal_LP} summarizes the main evidence at the 
30-day horizon. The left panel reports the full-sample innovation LP; 
the right panel reports the narrative LP.

%% ============================================================
%% TABLE
%% ============================================================
\begin{table}[htbp]
\centering
\caption{\textbf{Local-Projection Validation of the Channel Taxonomy}}\label{tab:ET_robustness}\label{tab:lp_validation}
\label{tab:causal_LP}
\renewcommand{\arraystretch}{1.20}
\resizebox{\textwidth}{!}{%
\tiny
\begin{tabular}{@{} p{1.9cm} c  c c c  @{\qquad}  c c c @{}}
\toprule
 & & \multicolumn{3}{c}{\scriptsize\textsc{Full-Sample Innovation LP}} & \multicolumn{3}{c}{\scriptsize\textsc{Narrative LP}} \\
\cmidrule(lr){3-5} \cmidrule(l){6-8}
 & & {\scriptsize(A)} & {\scriptsize(B)} & {\scriptsize(C)} & {\scriptsize(A)} & {\scriptsize(B)} & {\scriptsize(C)} \\
 & & {Baseline} & {Global} & {Extended} & {Baseline} & {Global} & {Extended} \\
\cmidrule(lr){3-3} \cmidrule(lr){4-4} \cmidrule(lr){5-5}\cmidrule(lr){6-6} \cmidrule(lr){7-7} \cmidrule(lr){8-8}
\scriptsize\textbf{Episode} & \scriptsize\textbf{Channel} & {$\hat\beta$\;Sig.} & {$\hat\beta$\;Sig.} & {$\hat\beta$\;Sig.} & {$\hat\beta$\;Sig.} & {$\hat\beta$\;Sig.} & {$\hat\beta$\;Sig.} \\
\midrule
\multirow{4}{1.9cm}{\raggedright \scriptsize\textbf{Russia--Ukraine}\\
\tiny GPR, Feb.\,2022\\
\tiny\textit{Geopolitical}} & $\varphi^{\mathrm{dir}}$ & $+.005$\;-- & $+.005$\;-- & $+.005$\;-- & \cellcolor{blue!18}$+.122$\;$^{***}$ & \cellcolor{blue!18}$+.121$\;$^{***}$ & \cellcolor{blue!18}$+.125$\;$^{***}$ \\
 & $\varphi^{\mathrm{GFC}}$ & $-.001$\;-- & $-.001$\;-- & $-.001$\;-- & \cellcolor{blue!18}$-.102$\;$^{***}$ & \cellcolor{blue!18}$-.104$\;$^{***}$ & \cellcolor{blue!18}$-.102$\;$^{***}$ \\
 & $\varphi^{\mathrm{UNC}}$ & $-.000$\;-- & $-.000$\;-- & $-.000$\;-- & \cellcolor{blue!7}$-.009$\;$^{*}$ & \cellcolor{blue!7}$-.010$\;$^{*}$ & \cellcolor{blue!7}$-.009$\;$^{*}$ \\
 & $\varphi^{\mathrm{LOC}}$ & $-.004$\;-- & $-.004$\;-- & $-.004$\;-- & $-.018$\;-- & $-.023$\;-- & $-.023$\;-- \\[4pt]
\midrule
\multirow{4}{1.9cm}{\raggedright \scriptsize\textbf{Hamas--Israel}\\
\tiny GPR, Oct.\,2023\\
\tiny\textit{Geopolitical}} & $\varphi^{\mathrm{dir}}$ & $+.005$\;-- & $+.005$\;-- & $+.005$\;-- & \cellcolor{blue!18}$+.071$\;$^{***}$ & \cellcolor{blue!18}$+.072$\;$^{***}$ & \cellcolor{blue!18}$+.068$\;$^{***}$ \\
 & $\varphi^{\mathrm{GFC}}$ & $-.001$\;-- & $-.001$\;-- & $-.001$\;-- & $+.003$\;-- & $+.000$\;-- & $-.000$\;-- \\
 & $\varphi^{\mathrm{UNC}}$ & $-.000$\;-- & $-.000$\;-- & $-.000$\;-- & $+.006$\;-- & $+.006$\;-- & $+.006$\;-- \\
 & $\varphi^{\mathrm{LOC}}$ & $-.004$\;-- & $-.004$\;-- & $-.004$\;-- & $-.017$\;-- & $-.018$\;-- & $-.019$\;-- \\[4pt]
\midrule
\multirow{4}{1.9cm}{\raggedright \scriptsize\textbf{U.S.\ Election}\\
\tiny EPU, Nov.\,2024\\
\tiny\textit{Geoeconomic}} & $\varphi^{\mathrm{dir}}$ & $+.002$\;-- & $+.002$\;-- & $+.003$\;-- & \cellcolor{blue!18}$+.013$\;$^{***}$ & \cellcolor{blue!18}$+.013$\;$^{***}$ & \cellcolor{blue!18}$+.015$\;$^{***}$ \\
 & $\varphi^{\mathrm{GFC}}$ & $+.002$\;-- & $+.002$\;-- & $+.004$\;-- & \cellcolor{blue!18}$-.044$\;$^{***}$ & \cellcolor{blue!18}$-.046$\;$^{***}$ & \cellcolor{blue!18}$-.046$\;$^{***}$ \\
 & $\varphi^{\mathrm{UNC}}$ & $+.001$\;-- & $+.001$\;-- & $+.001$\;-- & \cellcolor{blue!18}$+.021$\;$^{***}$ & \cellcolor{blue!18}$+.021$\;$^{***}$ & \cellcolor{blue!18}$+.024$\;$^{***}$ \\
 & $\varphi^{\mathrm{LOC}}$ & $+.000$\;-- & $+.000$\;-- & $+.000$\;-- & \cellcolor{blue!18}$+.016$\;$^{***}$ & \cellcolor{blue!18}$+.015$\;$^{***}$ & \cellcolor{blue!18}$+.017$\;$^{***}$ \\[4pt]
\midrule
\multirow{4}{1.9cm}{\raggedright \scriptsize\textbf{``Liberation Day''}\\
\tiny TPU, Apr.\,2025\\
\tiny\textit{Geoeconomic}} & $\varphi^{\mathrm{dir}}$ & $-.001$\;-- & $-.001$\;-- & $-.002$\;-- & \cellcolor{blue!18}$-.008$\;$^{***}$ & \cellcolor{blue!18}$-.009$\;$^{***}$ & \cellcolor{blue!18}$-.016$\;$^{***}$ \\
 & $\varphi^{\mathrm{GFC}}$ & $-.001$\;-- & $-.001$\;-- & $-.001$\;-- & $-.015$\;-- & \cellcolor{blue!7}$-.019$\;$^{*}$ & \cellcolor{blue!18}$-.028$\;$^{***}$ \\
 & $\varphi^{\mathrm{UNC}}$ & $+.001$\;-- & $+.001$\;-- & $+.001$\;-- & \cellcolor{blue!18}$+.026$\;$^{***}$ & \cellcolor{blue!18}$+.026$\;$^{***}$ & \cellcolor{blue!18}$+.034$\;$^{***}$ \\
 & $\varphi^{\mathrm{LOC}}$ & $+.000$\;-- & $+.000$\;-- & $-.001$\;-- & \cellcolor{blue!12}$+.034$\;$^{**}$ & \cellcolor{blue!12}$+.032$\;$^{**}$ & \cellcolor{blue!7}$+.023$\;$^{*}$ \\

\addlinespace[2pt]
\bottomrule
\end{tabular}%
}% end resizebox

\vspace{6pt}
\begin{minipage}{\textwidth}
\scriptsize
\textit{Notes:} Each cell reports the cumulative impulse response at 
horizon $h = 30$~days. Coefficients rounded to three decimal places. 
\textit{Full-sample innovation LP}: panel local projection with 
country-specific AR(5) innovations; Driscoll--Kraay SE with bandwidth 
$\max(20,h)$, country and time FE, 5~lags. \textit{Narrative LP}: 
panel local projection with $\pm 3$-day event dummies; Driscoll--Kraay 
SE with country FE. Controls: (A)~none; (B)~VIX, US2Y; (C)~VIX, 
US2Y, ECO, INT, POL, EPU, TPU. $^{***}\,p<0.01$,\; 
$^{**}\,p<0.05$,\; $^{*}\,p<0.10$; -- $=$ not significant at 10\%. 
Cell shading indicates significance level of the Narrative LP 
estimate: \colorbox{blue!18}{\phantom{xx}} $p<0.01$,\; 
\colorbox{blue!12}{\phantom{xx}} $p<0.05$,\; 
\colorbox{blue!7}{\phantom{xx}} $p<0.10$. Because the full-sample 
innovation LP is estimated by shock type rather than by narrative 
episode, the GPR-based coefficients are repeated in the two 
geopolitical rows to facilitate comparison with the narrative panel. 
GPR = Geopolitical Risk, EPU = Economic Policy Uncertainty, 
TPU = Trade Policy Uncertainty, US2Y = U.S.\ two-year Treasury yield.
\end{minipage}
\end{table}

%% ============================================================
%% 6.1 — FULL-SAMPLE INNOVATION LP
%% ============================================================
\subsection{Full-Sample Innovation LP}
\label{sec:LP_AR5}
The left panel of Table~\ref{tab:causal_LP} reports the full-sample 
innovation LP. For GPR, EPU, and TPU innovations, the 30-day responses 
are small in magnitude and statistically indistinguishable from zero 
across all four channels and all three control sets. No coefficient 
exceeds $\pm 0.005\sigma$ in absolute value, and none is significant 
at the 10\% level.\footnote{Despite the aggregate null, channel-level IRFs 
produce qualitative sign patterns consistent with four of the 
taxonomy's sign conditions at the 68\% level, though none reaches 
conventional significance. See Appendix~\ref{app:LP_FD} for 
channel-level detail at $h = 5, 30, 60$.}

This null is consistent with the full-sample design averaging 
over many routine news days on which channel responses are weak or 
offsetting. Three results favour attenuation over absence of transmission: the 
narrative LP recovers large and significant channel responses for 
precisely the episodes where the taxonomy predicts they should be 
strongest; the placebo falsification 
confirms that these episodes are exceptional relative to random dates; and a narrative sign-restricted SVAR 
estimated on raw observables recovers the scissors 
pattern with theory-consistent signs (section 8.1 and further details in Appendix~\ref{app:svar}).%% ============================================================
%% 6.2 — NARRATIVE LP
%% ============================================================
\subsection{Narrative LP: Four Crisis Episodes}
\label{sec:LP_narrative}

In sharp contrast to the full-sample null, the narrative LP detects 
large and statistically significant aggregate CDS responses: 
Russia--Ukraine generates $+0.54\sigma$ at $h=30$ ($p<0.01$), rising 
to ${\sim}1.5\sigma$ by $h=90$; Hamas--Israel produces $-0.18\sigma$ 
($p<0.05$), consistent with flight-to-quality compression for 
non-MENA sovereigns; the U.S.\ election raises CDS by $+0.11\sigma$ 
($p<0.10$) with a gradual 60-day build-up; and ``Liberation Day'' 
tariffs produce a hump-shaped response peaking at $+0.15\sigma$ 
($p<0.01$) before partially reversing.\footnote{Full aggregate 
impulse-response paths are reported in Appendix~\ref{app:LP_FD}.} 
The contrast with the innovation null resolves an apparent puzzle: 
geopolitical shocks have large effects on sovereign spreads, but 
these effects are concentrated in discrete crisis episodes that are 
washed out when averaged over routine news-innovation days.

\begin{figure}[ht!]
    \setlength{\belowcaptionskip}{-5pt}
    \centering
    \caption{\textbf{Channel Impulse Responses from Narrative Local 
    Projections}}
    \vspace{2pt}
    \includegraphics[width=1\textwidth]{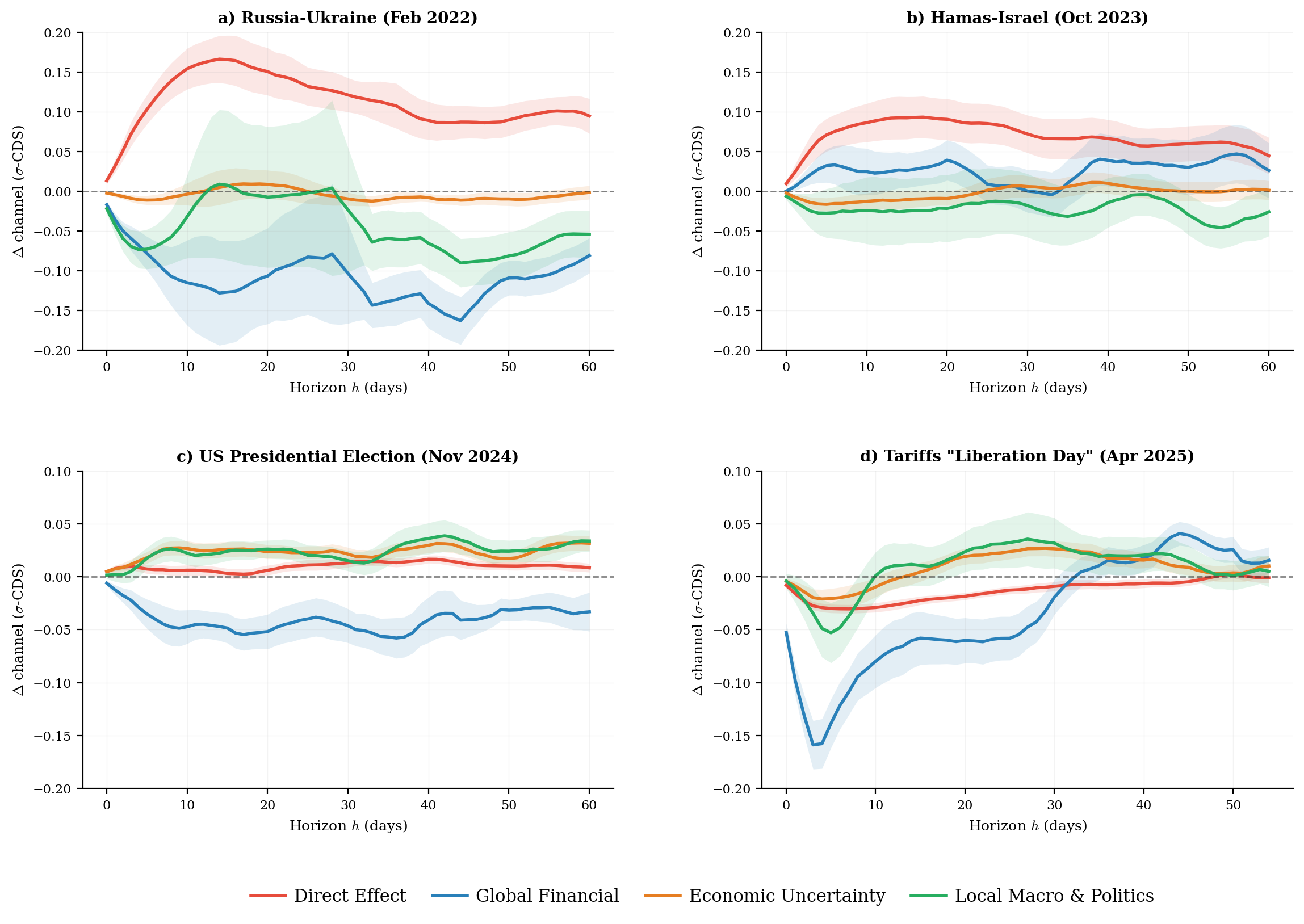}
    \label{fig:channels_taxonomy}
    \begin{flushleft}
       \scriptsize
        \vspace{-5pt}
        Notes: Each panel plots cumulative impulse responses of the four Shapley--Taylor channels---Direct ($\varphi^{\mathrm{dir}}$, red), GFC ($\varphi^{\mathrm{GFC}}$, blue), Uncertainty ($\varphi^{\mathrm{UNC}}$, orange), and Local ($\varphi^{\mathrm{LOC}}$, green)---from narrative local projections with $\pm3$-day event dummies at horizons $h = 0, \ldots, 30$ days. Top row: geopolitical episodes; bottom row: geoeconomic episodes. Specification~(B): VIX and US2Y controls; country FE; Driscoll--Kraay SE. Shaded bands: 90\% confidence intervals..
    \end{flushleft}
\end{figure}

\paragraph{Channel decomposition.}
Figure~\ref{fig:channels_taxonomy} plots the full horizon profiles for 
the four channels under the narrative design. 
Table~\ref{tab:causal_LP} reports the corresponding cumulative 
responses at $h=30$ under the three control sets. For interpretation, 
specification~(B) is our preferred baseline because it conditions on 
global financial variables while remaining parsimonious; 
specifications~(A) and~(C) show that the results are stable to 
omitting or extending the control set. Four results stand out.

\textit{E1---Russia--Ukraine (GPR shock).}\enspace The Direct channel 
rises sharply ($\hat{\beta}^{h=30} = +0.121$, $p < 0.01$) while the 
GFC channel moves in the opposite direction ($-0.104$, $p < 0.01$), 
recovering the scissors pattern at the 1\% significance level. The 
horizon profiles in Figure~\ref{fig:channels_taxonomy} suggest that 
the Direct and GFC channels decay at different speeds, with the Direct 
response more persistent.\footnote{We also estimate a 
threshold specification and a smooth-transition variant following 
\citet{AuerbachGorodnichenko2012} and \citet{RameyZubairy2018}. 
Neither recovers the scissors using full-sample innovation shocks.}

\textit{E2---Hamas--Israel (GPR shock).}\enspace The Direct channel 
activates ($+0.072$, $p < 0.01$) at roughly half the E1 magnitude, 
while the GFC channel is indistinguishable from zero ($+0.000$). The 
E1--E2 contrast supports a distinction between globally systemic 
geopolitical shocks (textbook scissors) and regionally contained ones 
(no GFC offset)---the single unconfirmed prediction among the sixteen 
event--channel tests.

\textit{E3---U.S.\ Presidential Election (EPU shock).}\enspace The 
Uncertainty channel is the largest positive response ($+0.021$, 
$p < 0.01$), the GFC is significantly negative ($-0.046$, 
$p < 0.01$), and both Direct ($+0.013$) and Local ($+0.015$) 
channels are positive and significant. The key point is not that the 
Direct channel is literally absent, but that it is secondary to the 
Uncertainty and GFC responses. This multi-channel activation---with 
Uncertainty as first responder---is the broad-based fingerprint that 
distinguishes policy-uncertainty shocks from both geopolitical and 
trade-policy episodes.

\textit{E4---``Liberation Day'' Tariffs (TPU shock).}\enspace The 
Uncertainty ($+0.026$, $p < 0.01$) and Local ($+0.032$, $p < 0.05$) 
channels are the dominant positive contributors, while the GFC drops 
on impact ($-0.019$, $p < 0.10$). The Direct channel response 
($-0.009$, $p < 0.01$) is economically small relative to the dominant 
channels---an order-of-magnitude difference from Russia--Ukraine 
($\varphi^{\mathrm{dir}} = +0.121$), establishing that geopolitical 
and geoeconomic shocks generate qualitatively different Direct-channel 
responses.

\paragraph{Stability and scorecard.}
The qualitative pattern is stable across the three control sets. 
Moving from specification~(A) to specification~(C) changes point 
estimates only modestly (e.g., the E1 Direct channel moves from 
$+0.122$ to $+0.125$; the E1 GFC from $-0.102$ to $-0.102$) and 
leaves the main sign pattern intact. Overall, the narrative LP 
supports 15 of 16 event--channel predictions: 
$\varphi^{\mathrm{dir}}$ activates for both geopolitical events 
(E1,~E2) but is economically negligible for the geoeconomic event 
(E4); $\varphi^{\mathrm{UNC}}$ dominates for the 
political-uncertainty event (E3); and the scissors is recovered at 
the 1\% significance level for E1. Because this validation exercise 
is separate from model selection and from the Shapley decomposition 
itself, it provides complementary econometric support for the 
proposed interpretation of the channels.%% 

\begin{figure}[t]
\setlength{\belowcaptionskip}{-5pt}
    \centering
    \caption{\textbf{Placebo Falsification: Actual Channel Responses vs.\
  Random-Date Envelope.}}
    \vspace{2pt}
  \centering
  \includegraphics[width=\textwidth]{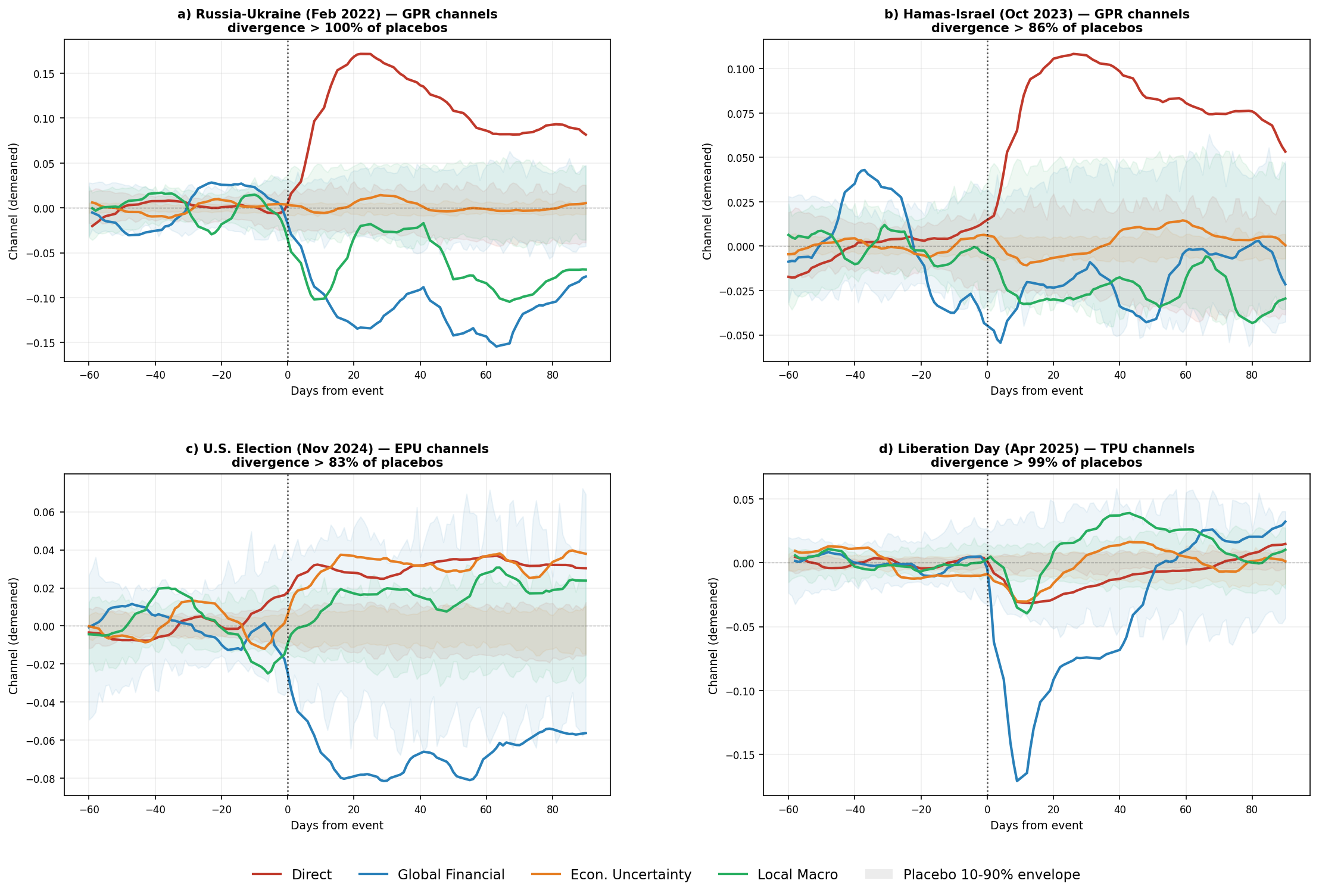}
  \begin{flushleft}
       \scriptsize
        \vspace{-5pt}
  Notes: Each panel plots the cross-country mean of the four
  demeaned channel series ($\varphi^{\mathrm{dir}}$, $\varphi^{\mathrm{gfc}}$,
  $\varphi^{\mathrm{unc}}$, $\varphi^{\mathrm{loc}}$) in a $[-60,+90]$-day
  window around the event date (solid lines) against the 10--90\% envelope
  from 100 randomly drawn non-event dates (shaded bands). Each episode uses
  its driver-specific Shapley--Taylor decomposition: GPR for Russia--Ukraine
  and Hamas--Israel; EPU for the U.S.\ Presidential Election; TPU for the
  Liberation Day tariff shock. Panel titles report the percentile of the
  placebo distribution exceeded by the actual event's maximum post-event
  channel divergence. Channel series smoothed with a 7-day trailing moving
  average and demeaned over the pre-event window.
  \label{fig:placebo_4ep}
  \end{flushleft}
\end{figure}

\paragraph{Placebo falsification.}
\label{sec:placebo}
To verify that the scissors pattern is event-specific, we draw 100 random
non-event dates and construct the 10--90\% envelope of channel responses in a
$[-60,+90]$-day window around each placebo date, using the driver-specific
Shapley--Taylor decomposition for each episode (GPR for geopolitical events;
EPU and TPU for geoeconomic events). All four episodes produce channel
divergence well outside this envelope: Russia--Ukraine exceeds 100\% of
placebos, Liberation Day 99\%, Hamas--Israel 86\%, and the U.S.\ Presidential
Election 83\%. Crucially, the channels that exit the placebo bands differ
across shock types in the direction the taxonomy predicts.
Figures~\ref{fig:placebo_RU}--\ref{fig:placebo_LD} in
Appendix~\ref{app:placebo} decompose each episode channel by channel.
For geopolitical episodes, the Direct channel exits the 90\% band upward while
the GFC channel exits downward, and the Uncertainty channel remains inside the
placebo range. For geoeconomic episodes, the pattern reverses: the GFC channel
dominates the departure, the Uncertainty channel exits the envelope (upward for
the Election, consistent with EPU-driven repricing), while the Direct channel
stays within or moves opposite to the geopolitical case.

\paragraph{Block bootstrap validation.}
Because the channel decomposition treats Shapley--Taylor values as
observed regressors, the Driscoll--Kraay standard errors in
Table~\ref{tab:causal_LP} do not account for estimation uncertainty
in the first-stage ML model. We address this with a
generated-regressor block bootstrap ($B=500$) that, in each
replication, jointly re-estimates the Multilayer Random Forest tree,
recomputes the Shapley--Taylor decomposition, and re-runs the panel
local projection. Appendix Figure~\ref{fig:bootstrap_CI} overlays
the resulting 90\% confidence intervals on the baseline
Driscoll--Kraay bands.

The graph yields three findings. First, for the raw CDS spread---a
directly observed variable unaffected by the first-stage
model---bootstrap and Driscoll--Kraay confidence intervals are
comparable in width, confirming that the two methods agree when no
generated regressor is involved. Second, for the four channel
series, bootstrap bands are substantially tighter than the
Driscoll--Kraay bands. This reflects two distinct uncertainty
sources: the channel decomposition is highly stable across
re-estimations of the random forest trained on approximately 75,000
observations, so the bootstrap correctly reports that first-stage
estimation noise is small; the Driscoll--Kraay estimator additionally
corrects for cross-sectional dependence across 42 sovereigns exposed
to common shocks---a source of uncertainty that the block bootstrap,
which resamples temporal blocks within a fixed cross-section, does
not fully replicate. The two methods therefore bound different
margins: the bootstrap captures first-stage ML instability, while
Driscoll--Kraay captures second-stage panel dependence. We use the
latter as the conservative baseline throughout. Third, no channel
that is significant under Driscoll--Kraay loses significance under
the bootstrap; if anything, several impact responses ($h=0$) that
fall short of conventional thresholds under DK become significant
once generated-regressor uncertainty \citep{Pagan1984} is properly
internalized. Together, these results confirm that the narrative LP evidence
supporting the transmission taxonomy---the scissors pattern, the
event--channel sign predictions, and the placebo rankings documented
in Sections~\ref{sec:Causal_LP}--\ref{sec:taxonomy_test}---is not an
artifact of a single ML partition.

\section{Evaluating the Channel Taxonomy}
\label{sec:taxonomy_test}

The semistructural framework in Section~\ref{sec:benchmarks} generates four dominance benchmarks---Predictions D, G, U, and L---each mapping a shock family to a specific channel configuration (Table~\ref{tab:benchmarks}).  Evaluated jointly across the four crisis episodes and four channels, the taxonomy produces $4 \times 4 = 16$ event--channel predictions.  We assess each cell as a pass if the sign and relative magnitude of the estimated $\Delta_{1m}^{c} \equiv \bar{\varphi}^{c}_{\mathrm{post},1m} - \bar{\varphi}^{c}_{\mathrm{pre}}$ conform to the predicted direction, using the calibrated thresholds from the Shapley--Taylor decomposition (Table~\ref{tab:benchmarks}).

\begin{figure}[ht!]
    \setlength{\belowcaptionskip}{-5pt}
    \centering
    \caption{\textbf{Cluster Bootstrap Validation of Taxonomy of Events}}\label{fig:bootstrap_scorecard}
     \subcaption{1000 replications, resampling 42 countries with replacement}
        \includegraphics[width=\textwidth]{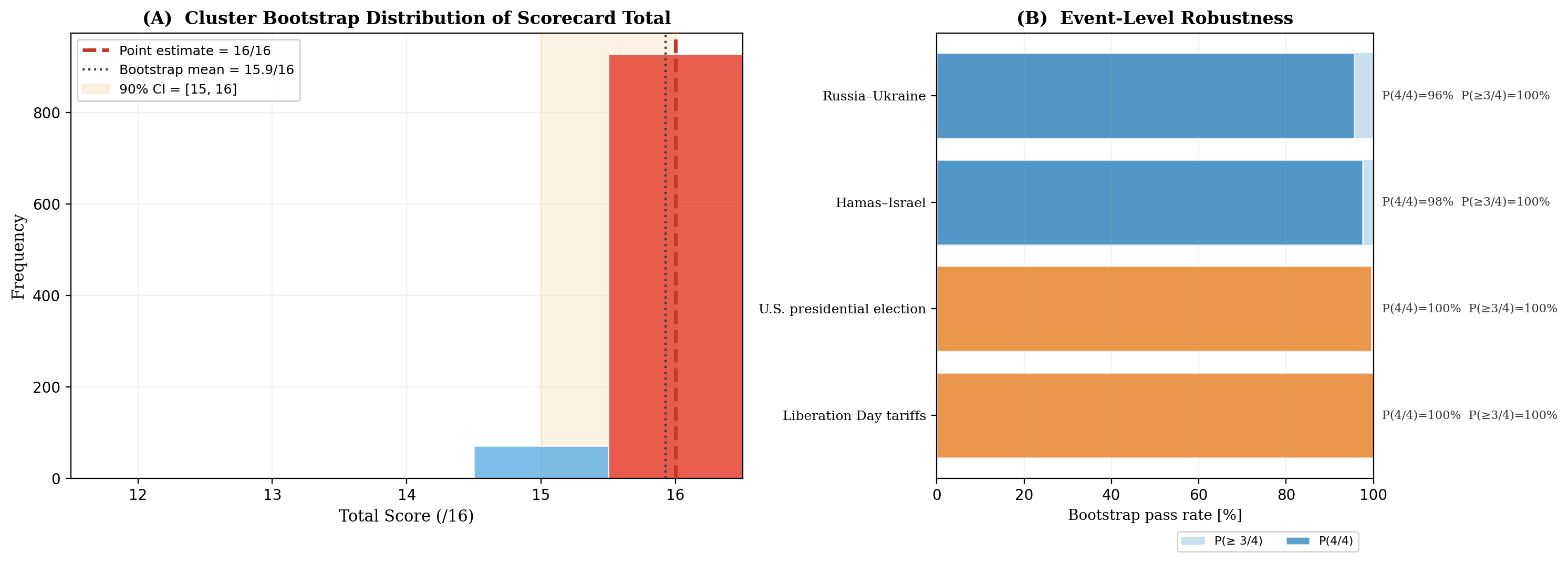}
      
    \vspace{2pt}

    \begin{flushleft}
       \scriptsize
        \vspace{-5pt}
        Notes: Each of B = 1,000 replications resamples 42 countries with replacement from the original panel, recomputes the cross-country daily panel mean of each Shapley-Taylor channel, and re-scores the 1-event taxonomy. Point estimate: 16/16. Bootstrap mean: 15.9/16 (sd = 0.3). 90\% CI: [15, 16]. Panel (B) dark bars show P(1/4); light bars show P($\geq$ 3/4). Values report P(4/4) | P($\geq$ 3/4).
    \end{flushleft}
\end{figure}

The point estimate is \textbf{16 out of 16}: every event--channel cell matches its predicted sign or dominance inequality.  The two geopolitical episodes activate the Direct channel ($\varphi^{\mathrm{dir}} > 0$, significant) while the GFC channel moves in the opposite direction, recovering the scissors condition from Prediction~G\@.  The Uncertainty and Local channels remain secondary, consistent with Predictions U and L\@.  The two geoeconomic episodes display the inverted configuration: the Direct channel is small relative to its geopolitical counterpart ($|\varphi^{\mathrm{dir}}_{\mathrm{geoeco}}| \ll |\varphi^{\mathrm{dir}}_{\mathrm{geo}}|$), while the GFC, Uncertainty, and Local channels carry the transmission, with $|\varphi^{\mathrm{GFC}}| > |\varphi^{\mathrm{dir}}|$ confirming GFC dominance under geoeconomic shocks.

To assess whether this result depends on the particular composition of the cross-sectional panel, we perform a cluster bootstrap with $B = 1{,}000$ replications, resampling 42 countries with replacement.  Figure~\ref{fig:bootstrap_scorecard} reports the results.  The bootstrap mean is $15.9/16$ (sd $= 0.3$), with a 90\% confidence interval of $[15,\, 16]$; over 95\% of replications score 16/16.  Panel~(B) decomposes robustness by episode: the two geoeconomic episodes pass all four predictions in 100\% of replications; Hamas--Israel at $P(4/4) = 98\%$; Russia--Ukraine at $P(4/4) = 95\%$, with the only residual fragility in the marginally significant Uncertainty channel---consistent with geopolitical shocks not activating policy-regime uncertainty at short horizons (Prediction~U).

\subsection{Independent Validation via Sign-Restricted SVARs}
\label{sec:svar_validation}

The scorecard and bootstrap validation above rely on the Shapley--Taylor channel decomposition.  A natural concern is that the channel assignments may reflect properties of the machine-learning model rather than properties of transmission.  To address this, we estimate narrative sign-restricted SVARs on \emph{raw observable composites}---entirely bypassing the Shapley--Taylor decomposition---following the identification logic of \citet{AntolinDiazRubioRamirez2018}.

For each country~$i$, we estimate a five-variable VAR(5) on observables that map directly into the four transmission channels: the sovereign CDS spread, the country-level geopolitical risk index (GPR, proxying the Direct channel), a global financial-conditions composite (VIX and U.S.\ two-year Treasury yield, proxying the GFC channel), an uncertainty composite (EPU and TPU, proxying the Uncertainty channel), and a domestic macro-political composite (ECO, INT, POL, proxying the Local channel).  Identification combines sign restrictions derived from the semistructural benchmarks in Table~\ref{tab:benchmarks}, relative-magnitude restrictions encoding the scissors logic ($|\Theta^{\text{GFC}}| < |\Theta^{\text{GPR}}|$ under geopolitical shocks; $|\Theta^{\text{GFC}}| > |\Theta^{\text{GPR}}|$ under geoeconomic shocks), and narrative restrictions on the four crisis dates.\footnote{Full specification details---including the restriction design, estimation algorithm, and companion SVAR on Shapley--Taylor channel series---are provided in Appendix~\ref{app:svar}.}

Figure~\ref{fig:svar_raw_irfs} reports the mean-group structural impulse responses from 639 accepted draws.  Under the geopolitical shock, GPR rises sharply (approximately $+0.10$ around $h = 10$, with the 68\% credible band above zero) while the financial-conditions composite falls (approximately $-0.06$, with the 68\% band below zero), recovering the scissors pattern with $|\Theta^{\text{GFC}}| < |\Theta^{\text{GPR}}|$.  The Uncertainty and Local composites remain secondary, consistent with Predictions~U and~L.  Under the geoeconomic shock, the GPR response is muted while the financial-conditions channel is the dominant response (approximately $-0.12$), with the Uncertainty ($+0.06$) and Local ($+0.04$) composites carrying the remaining transmission and $|\Theta^{\text{GFC}}| > |\Theta^{\text{GPR}}|$.  All eight sign and dominance predictions from Table~\ref{tab:benchmarks} are supported at the posterior median.

% --- Promoted Figure: SVAR on Raw Observables ---
\begin{figure}[ht!]
    \centering
    \caption{\textbf{Narrative Sign-Restricted SVAR: Structural Impulse Responses on Raw Observables}}
    \label{fig:svar_raw_irfs}
    \includegraphics[width=0.95\textwidth]{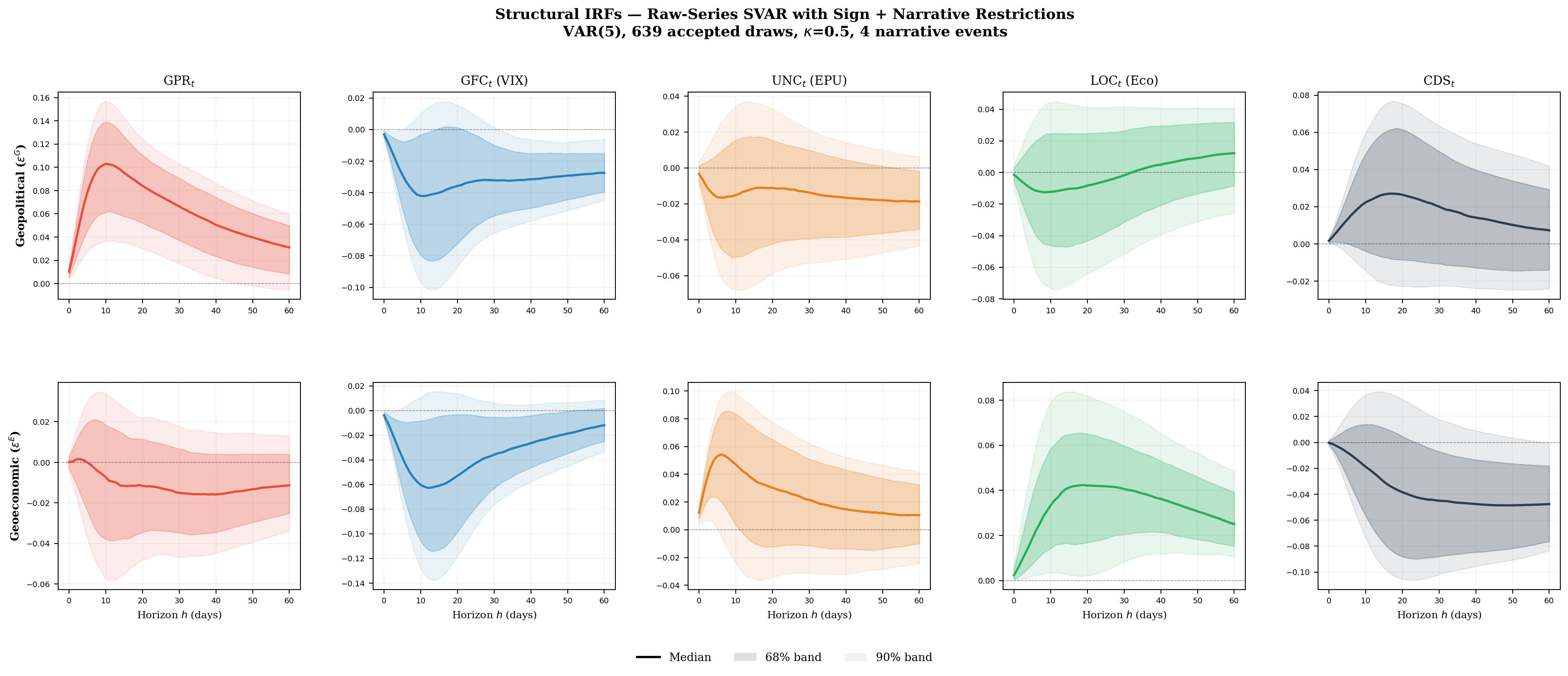}
    \begin{flushleft}
       \scriptsize
        Notes: Mean-group structural impulse responses from country-level VAR(5) models estimated on raw observable composites (equation~\ref{eq:svar_y_revised}), identified with sign restrictions (Table~\ref{tab:svar_signs}), relative-magnitude restrictions, and narrative restrictions on four crisis dates following \citet{AntolinDiazRubioRamirez2018}. Top row: geopolitical shock~($\varepsilon^G$); bottom row: geoeconomic shock~($\varepsilon^E$). Columns: GPR (Direct proxy), GFC composite (VIX-based), Uncertainty composite (EPU/TPU), Local composite (ECO/INT/POL), and sovereign CDS spread. Solid lines: posterior median across 639 accepted draws ($\kappa = 0.5$). Dark shaded bands: 68\% pointwise credible intervals; light bands: 90\%. Under the geopolitical shock, GPR rises sharply while GFC falls, recovering the scissors pattern ($|\Theta^{\textsc{gfc}}| < |\Theta^{\textsc{gpr}}|$). Under the geoeconomic shock, the GFC channel dominates ($|\Theta^{\textsc{gfc}}| > |\Theta^{\textsc{gpr}}|$), with Uncertainty and Local channels carrying the remaining transmission. All eight sign and dominance predictions from Table~\ref{tab:benchmarks} are supported at the posterior median. Because neither the estimation nor the identification uses Shapley--Taylor values, this exercise provides independent validation of the channel taxonomy.
    \end{flushleft}
\end{figure}

When the same SVAR-identified shocks are used as regressors in full-sample panel local projections, the responses attenuate sharply toward zero---economically small and statistically indistinguishable from zero across all horizons and control sets (Appendix Figures~\ref{fig:lp_svar_shocks} and~\ref{fig:svar_robustness}).  This attenuation mirrors the innovation-based null in Section~\ref{sec:LP_AR5} and is consistent with state-dependent transmission: the time-invariant SVAR weights average over states in which transmission is active and states in which it is dormant \citep{PlagborgMollerWolf2021}.

Table~\ref{tab:svar_summary} summarises how the four empirical frameworks relate to one another.  The Shapley--Taylor scorecard, the narrative LP, the sign-restricted SVAR on raw observables, and the full-sample LP with SVAR shocks form a coherent body of evidence: the taxonomy is recoverable from raw observables without the ML decomposition, and transmission is state-dependent rather than pervasive and linear.

% --- Promoted Table: Cross-Validation Summary ---
\begin{table}[h!]
\centering
\caption{Cross-Validation Summary Across Methods}
\label{tab:svar_summary}
\scriptsize
\begin{tabular}{@{}l l l c l@{}}
\toprule
Figure / section & Method & Variables & Result & Relation to baseline \\
\midrule
Main text
& Shapley $\Delta_{1m}$ + narrative LP
& ML channels
& 16/16 event--channel predictions
& Baseline \\[3pt]
Appendix Figure~\ref{fig:svar_channel_irfs}
& Sign-restricted SVAR
& ML channels
& 8/8 sign and magnitude relations
& Internal coherence \\[3pt]
Figure~\ref{fig:svar_raw_irfs}
& Sign-restricted mean-group SVAR
& Raw observables
& 8/8 sign and magnitude relations
& Independent of ML decomposition \\[3pt]
Appendix Figure~\ref{fig:lp_svar_shocks}
& Full-sample LP with identified SVAR shocks
& Raw observables
& Null average responses
& Consistent with state dependence \\
\bottomrule
\end{tabular}
\begin{flushleft}
\scriptsize
Notes: ``Result'' reports the number of predicted sign and relative-magnitude relations supported by the baseline posterior median. ``Independent of ML decomposition'' means that the raw-observable SVAR neither estimates nor identifies shocks using Shapley--Taylor values. ``Null average responses'' means that the full-sample LP responses are economically small and statistically indistinguishable from zero at conventional levels.
\end{flushleft}
\end{table}

\section{Conclusion}
\label{sec:conclusion}

This paper shows that geopolitical and geoeconomic shocks transmit to
sovereign credit risk through systematically different channels.  The
central result is a scissors pattern: geopolitical shocks raise
spreads primarily through direct sovereign repricing while the Global
Financial Cycle channel moves in the opposite direction, whereas
geoeconomic shocks bypass the Direct channel and transmit mainly
through financial conditions, policy uncertainty, and domestic
amplification.  This taxonomy is consistently recovered across three
independent empirical frameworks---Shapley--Taylor decomposition,
narrative local projections, and sign-restricted SVARs on raw
observables---providing a unified characterization of how different
shock families map into sovereign risk.

This distinction has direct implications for policy design.  When
spread widening is driven by global financial conditions, central-bank
liquidity provision and swap lines can help stabilize funding markets
\citep{BahajReis2022}.  By contrast, when spreads reflect
sovereign-specific geopolitical repricing, liquidity tools are
unlikely to address the underlying risk premium.  In such cases, the
relevant margins lie in diplomatic de-escalation, fiscal adjustment,
or institutional credibility.  The originator penalty documented for
geoeconomic shocks adds a further dimension: the country that
initiates a tariff or sanctions episode may face spread widening
through the Local channel even as monetary-easing expectations
compress spreads abroad, implying that geoeconomic instruments carry a
sovereign-risk cost for the originator that standard trade-policy
analysis does not capture.  More broadly, the results highlight a risk
of policy misdiagnosis: similar movements in spreads may require
fundamentally different responses depending on the underlying
transmission channel.  High-frequency news-based indicators are
therefore valuable not only as measures of risk, but as tools for
identifying the dominant channel in real time.

Several limitations deserve acknowledgment.  The channel measures
remain model-based decompositions, and the semistructural framework
serves as a benchmark rather than a fully estimated structural model.
The Shapley--Taylor attribution is an accounting exercise; the
narrative validation confirms theory-consistent signs across crisis
episodes but does not close the gap between accounting and structural
causation over the full sample.  The analysis is also silent on
welfare and optimal policy: we characterize \emph{how} shocks
transmit, but not \emph{what} the optimal central-bank or fiscal
response should be conditional on the active channel.

These limitations point to natural extensions.  First, embedding the
four-channel decomposition within a structural geoeconomic framework
such as \citet{ClaytonMaggioriSchreger2026} would link empirical
channel assignments to underlying policy trade-offs and move from sign
predictions to quantitative welfare analysis.  Second, augmenting the
episode set with pure global financial shocks---such as the SVB/Credit
Suisse stress of March~2023 or the yen carry-trade unwind of
August~2024---would complete a three-way taxonomy (geopolitical,
geoeconomic, financial) and test whether financial shocks generate a
third distinct channel configuration in which the GFC channel
dominates \emph{positively}, rather than through the offsetting or
transmitting roles documented here.  Third, modeling sovereign spreads
as state-dependent functions of the four channels would formalize the
threshold and interaction effects that the gradient-boosted trees
capture nonparametrically, providing a bridge between the
machine-learning and structural approaches.
\clearpage
\bibliographystyle{apalike}
\bibliography{references.bib}

\clearpage % This command forces a new page

\appendix

\section{APPENDIX:The data}
\label{app:data}

This Appendix details data availability and percentile distributions by country (Appendix A.1), the construction of news indicators (Appendix A.2) and LOESS Unconditional Scatters. 

\subsection{Data availability by country}
\label{sec:data_av}

 The following table presents country-specific data availability together with descriptive statistics summarizing the distribution of the main variables. For each country, the start and end dates of the available sample are reported. To characterize the distribution, we provide the 25th and 75th percentiles (p25 and p75), which capture the inter-quartile range where the central 50\% of observations lie, for each of the described variables in the data section.

\begin{table}[ht!]
    \setlength{\belowcaptionskip}{-5pt}
    \centering
    \caption{\textbf{Country-specific sample periods and interquartile distributions of variables}}
    \label{tab:data_table}
    \vspace{2pt}
    \includegraphics[width=\textwidth]{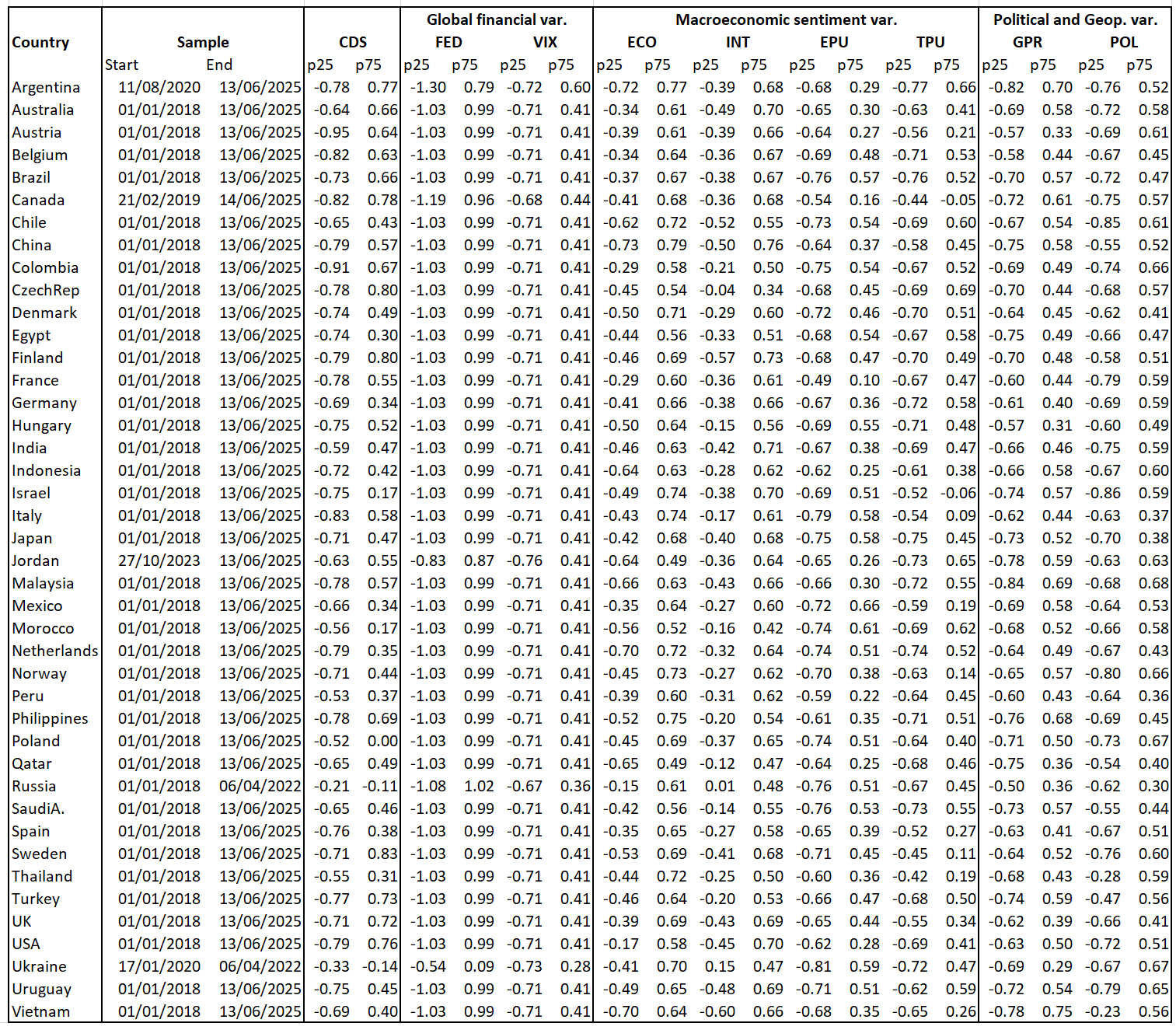}
    \vspace{-4pt}
    \parbox{\textwidth}{\scriptsize
    \textit{Notes:} The table reports country-specific sample periods and interquartile ranges (25th and 75th percentiles) for all variables used in the empirical analysis. The sample covers 42 sovereigns over January 2018--July 2025 (unbalanced panel); start dates vary by CDS data availability. All variables are expressed as 28-day moving averages, demeaned and standardized to zero mean and unit variance within each country. \textbf{CDS}: 5-year sovereign credit default swap spread (basis points, standardized). \textbf{Global financial variables}: FED = 2-year U.S.\ Treasury yield (proxy for global monetary policy); VIX = CBOE Volatility Index (proxy for global financial volatility). \textbf{Macroeconomic sentiment variables}: ECO = local economic sentiment index; INT = local interest-rate sentiment index; EPU = Economic Policy Uncertainty index; TPU = Trade Policy Uncertainty index. \textbf{Political and geopolitical variables}: GPR = Geopolitical Risk index; POL = Political Tensions index. All news-based indicators (ECO, INT, EPU, TPU, GPR, POL) are constructed by BBVA Research from GDELT data as described in Appendix~A.2. FED and VIX are global (non-country-specific) variables; their interquartile ranges differ across countries only because of variation in sample periods. Russia and Ukraine have truncated samples ending in June 2022 and January 2020, respectively, reflecting CDS market disruptions following the onset of the conflict. Jordan enters the sample in October 2023.}
\end{table}
\clearpage

\subsection{Media sentiment indicators development process}
\label{sec:indicators}

All news-based indicators are constructed by BBVA Research using the Global Database of Events, Language, and Tone \citep{Leetaru2013}. Each index is computed at the country--day level, smoothed with a 28-day moving average, and standardized to zero mean and unit variance over the country-specific sample period. No outlier treatment is applied, as extreme events are informative signals of shifts in risk perception. Four indices are used.\footnote{Complete country-specific keyword lists and GDELT taxonomy codes for all 42 countries are available in the replication package at \url{https://bigdata.bbvaresearch.com/en/}.}

\medskip
\noindent\textbf{Economic Policy Uncertainty (EPU).}
Following \citet{baker2016}, the EPU index measures the relative coverage of news articles that jointly mention a country name, terms related to economic uncertainty, and policy-relevant institutions or instruments. The generic query template is:

\smallskip
\begin{quote}
\small\texttt{\{COUNTRY\_NAME\} (uncertain OR uncertainty) (economic OR economy) (policy OR tax OR spending OR regulation OR central bank OR budget OR deficit)}
\end{quote}
\smallskip

\noindent Country-specific queries adapt this template to local institutional vocabulary. For instance, the United States query adds references to Congress, the White House, and the Federal Reserve; Mexico includes Banxico, BdeM, and the Chamber of Deputies; Spain references Hacienda, parliament, and specific fiscal terms.\footnote{Representative examples: \emph{United States}: ``United States (uncertainty OR uncertain) (economic OR economy) (congress OR legislation OR white house OR regulation OR federal reserve OR deficit)''; \emph{Mexico}: ``Mexico (economic OR economy) (uncertain OR uncertainty) (regulation OR deficit OR budget OR Bank OR BdeM OR Banxico OR congress OR senate OR deputies OR legislation OR taxes OR Federal Reserve)''; \emph{Spain}: ``Spain (uncertainty OR uncertain OR instability OR risk) (economic OR economy) (parliament OR government OR Hacienda OR deficit OR budget OR expenditure OR debt OR taxes OR law OR reform OR regulation OR Bank)''.} The index equals the ratio of EPU-related articles to total published articles for each country--day.

\medskip
\noindent\textbf{Trade Policy Uncertainty (TPU).}
Following \citet{CaldaraEtAl2020TPU}, the TPU index captures news coverage related to trade-policy instruments. The query requires the co-occurrence of a trade term and a policy-instrument term:

\smallskip
\begin{quote}
\small\texttt{(tariff OR tariffs OR import OR imports OR export OR exports OR trade OR dumping OR antidumping OR GATT OR WTO) (duty OR duties OR barrier OR barriers OR ban OR bans OR tax OR taxes OR subsidy OR subsidies)}
\end{quote}
\smallskip

\noindent The index equals the ratio of TPU-related articles to total published articles. Unlike the EPU, the TPU query is uniform across countries.

\medskip
\noindent\textbf{Geopolitical Risk (GPR).}
Following \citet{CaldaraIacoviello2022}, the GPR index combines tone (sentiment) and relative coverage of articles classified under two groups of GDELT conflict-related taxonomy themes. Group~1 covers military and political-violence categories (e.g., \texttt{WAR}, \texttt{CONFLICT}, \texttt{TERROR}, \texttt{MILITARY}, \texttt{REBELLION}, \texttt{PEACEKEEPING}). Group~2 covers threat and disruption categories (e.g., \texttt{SANCTIONS}, \texttt{BLOCKADE}, \texttt{ARMEDCONFLICT}, \texttt{CRISIS}, \texttt{RAID}, \texttt{KILL}). An article must match at least one theme from each group. The index is the product of tone and relative coverage, multiplied by $-1$ for interpretability (higher values $=$ greater geopolitical risk).

\medskip
\noindent\textbf{Political Tensions (POL).}
The Political Tensions Index captures tone and coverage of news classified under the GDELT taxonomy theme \texttt{USPEC\_POLITICS\_GENERAL1} (elections, government institutions, legislative activity, political scandals, civil rights, national security, among others) that additionally contain at least one of the following keywords: \emph{political instability, political uncertainty, political crisis, political polarization, political extremism, political turmoil, political conflict}. As with GPR, the index is the product of tone and relative coverage, multiplied by $-1$.\clearpage

% ============================================================
% APPENDIX — Semistructural Microfoundations
% ============================================================
% Changes from body-section version:
%   1. "Claude" removed
%   2. Framing paragraph added (skip-without-loss)
%   3. Predictions 1–4 now reference Section 3 dominance notation
%   4. Remark compressed to 2 sentences
%   5. Minor tightening throughout (~15% shorter)
% ============================================================

\section{APPENDIX: Semistructural Microfoundations}
\label{app:model}

This appendix provides micro-founded mechanisms for the dominance benchmarks
summarized in Table~\ref{tab:benchmarks}. Each channel is given minimal
internal structure---enough to show that the sign predictions, the dominance
inequalities, and the key empirical patterns (gravity decay, the scissors,
broad uncertainty activation, domestic amplification) arise from primitive
economic forces. Readers who accept the benchmarks of Section~\ref{sec:benchmarks}
on economic grounds may skip this appendix without loss of continuity.

\paragraph{Setup.}

Let $s_{i,t}$ denote the sovereign CDS spread of country~$i$ at time~$t$.
We write
\begin{equation}
  s_{i,t}
  \;=\;
  \mu_i + \tau_t
  + D_{i,t}
  + G_{i,t}
  + U_{i,t}
  + L_{i,t}
  + \varepsilon_{i,t}\,,
  \label{eq:spread_channels}
\end{equation}
where $\mu_i$ and $\tau_t$ are country and time effects, $\varepsilon_{i,t}$
is an idiosyncratic disturbance, and the four channel terms capture distinct
transmission mechanisms.

\paragraph{Channel~I: Direct sovereign repricing.}

A geopolitical shock of intensity $g_t \geq 0$ reduces expected output in
proportion to conflict proximity. Building on the endogenous-default tradition
\citep{Arellano2008}, we model the Direct channel as
\begin{equation}
  D_{i,t}
  = h\!\bigl(\alpha(d_i)\,g_t,\; F_{i,t}\bigr)\,,
  \label{eq:direct_channel}
\end{equation}
where $d_i$ is geodesic distance from country~$i$ to the conflict epicenter,
$\alpha(\cdot)$ is a decreasing and concave exposure function
($\alpha' < 0$, $\alpha'' > 0$), and $F_{i,t}$ is a fragility index
summarising the sovereign's proximity to its default boundary. The function
$h(\cdot)$ is increasing in both arguments and exhibits a \emph{threshold}:
when $F_{i,t}$ is low, moderate exposure has a small effect on spreads; when
$F_{i,t}$ exceeds a critical level $\bar{F}$, the same exposure triggers a
nonlinear jump as the sovereign approaches its default boundary. The
concavity of $\alpha(\cdot)$ in distance implies steep attenuation near the
epicenter and flattening at greater distances---the functional form captured
by the log-distance specification estimated in Section~\ref{sec:fingerprints}.

This channel provides the microfoundation for Prediction~D of
Section~\ref{sec:benchmarks}: geopolitical shocks generate large $D_{i,t}$
for exposed sovereigns, while geoeconomic shocks---which do not directly
impair repayment capacity---produce
$|D_{i,t}^{\,\text{geoeco}}| \ll |D_{i,t}^{\,\text{geo}}|$.

\paragraph{Channel~II: Global Financial Cycle.}

Sovereign CDS are priced by global financial intermediaries whose risk-bearing
capacity fluctuates with balance-sheet constraints
\citep{HeKrishnamurthy2013,GabaixMaggiori2015}. Let $W_t$ denote
intermediary-sector wealth and $\bar{W}$ its leverage constraint. The GFC
channel is
\begin{equation}
  G_{i,t}
  = \lambda_i \cdot \xi(W_t)\,,
  \qquad
  \xi' < 0\,,\;\;
  \xi'' > 0\,,
  \label{eq:global_channel}
\end{equation}
where $\xi(W_t)$ is a convex amplification function that rises steeply as the
intermediary approaches its constraint, and $\lambda_i$ is country~$i$'s
loading on the global financial factor, determined by its integration into
dollar funding markets, reliance on portfolio flows, and exchange-rate regime
\citep{rey2013,MirandaAgrippinoRey2020}. When $W_t$ is high
(accommodative conditions), $\xi(\cdot)$ is flat and a volatility spike has
moderate effects; when $W_t$ is low (tight conditions), $\xi(\cdot)$ is steep
and the same spike is powerfully amplified \citep{BrunoShin2015}.

\paragraph{The scissors condition.}

The intermediary framework delivers the scissors pattern. A large geopolitical
shock ($g_t > 0$) raises $D_{i,t}$ for exposed sovereigns. Simultaneously,
two offsetting forces act on $W_t$: mark-to-market losses on exposed debt
reduce it, while flight-to-quality flows and central-bank liquidity provision
\citep{BahajReis2022} replenish it. When intermediary capacity is sufficient
($W_t \gg \bar{W}$) and the central-bank response is aggressive, the net
effect is $\Delta W_t > 0$, so $\xi(W_t)$ \emph{falls}:
\begin{equation}
  \boxed{\;
  \underbrace{\Delta D_{i,t} > 0}_{\text{Direct rises}}
  \qquad\text{while}\qquad
  \underbrace{\Delta G_{i,t} < 0}_{\text{GFC falls}}
  \;}
  \label{eq:scissors_condition}
\end{equation}
This is the scissors. Russia--Ukraine (large shock, robust intermediary,
aggressive ECB/Fed action) produces a textbook scissors; Hamas--Israel
(smaller shock, limited global spillover) produces a muted GFC offset. When
the shock overwhelms intermediary capacity ($\Delta W_t < 0$), the GFC
\emph{reinforces} rather than offsets the Direct channel---the financial-exclusion
regime exemplified by Ukraine ($\varphi^{\text{GFC}}_{1m} = +0.66$).

This mechanism delivers the dominance component of Prediction~G: under
geopolitical shocks the GFC is the secondary, offsetting channel
($|\varphi^{\textsc{gfc}}| < |\varphi^{\textsc{dir}}|$), whereas under
geoeconomic shocks---which compress the discount-rate factor without
activating the Direct channel---the GFC becomes the dominant transmission
channel ($|\varphi^{\textsc{gfc}}| > |\varphi^{\textsc{dir}}|$).

\paragraph{Channel~III: Geoeconomic Uncertainty.}

We depart from \citet{PastorVeronesi2012,PastorVeronesi2013} in two respects.
First, in PV, political uncertainty resolves when the government announces
policy. In a geoeconomic setting, uncertainty is \emph{self-reinforcing}: a
tariff announcement generates uncertainty about retaliation, escalation, and
regime change. We model the underlying state as a hidden Markov geoeconomic
regime $\theta_t \in \{\theta_L, \theta_H\}$, with investors updating beliefs
$\pi_{i,t} = \Pr(\theta_t = \theta_H \mid \mathcal{I}_{i,t})$ from noisy
signals. Second, PV operates in a closed-economy equity setting with a
representative agent; our setting requires a multi-country sovereign credit
market with a global intermediary.

The Uncertainty channel is:
\begin{equation}
  U_{i,t}
  = \kappa_i \cdot \mathcal{H}(\pi_{i,t})
  + \delta_i \cdot \pi_{i,t}\,,
  \label{eq:uncertainty_channel}
\end{equation}
where $\mathcal{H}(\pi) = -\pi\ln\pi - (1-\pi)\ln(1-\pi)$ is the Shannon
entropy of the posterior and $\kappa_i$ is country~$i$'s sensitivity to
geoeconomic ambiguity. The first term captures the pure \emph{uncertainty
premium}, maximised at $\pi = \tfrac{1}{2}$ and vanishing at the extremes.
The second captures the \emph{level effect}: higher $\pi_{i,t}$ raises spreads
because the high-tension regime implies worse expected fundamentals.

The entropy function is concave, generating \emph{convexity in the uncertainty
premium}: marginal uncertainty increases matter more at moderate levels than at
extremes---the nonlinearity documented in the EPU$\,\times\,$VIX interaction
surface of Figure~\ref{fig:shap3d}. Crucially, when a common
geoeconomic signal arrives (an election outcome, a tariff announcement),
$\pi_{i,t}$ updates simultaneously for all countries, producing correlated
uncertainty premia with narrow cross-country dispersion---the pattern
documented for the election episode (IQR $[+0.01, +0.06]$; $\sigma = 0.02$).

This structure delivers both Prediction~U and the temporal distinction
underlying its dominance formulation: geopolitical shocks do not directly
shift the distribution over $\{\theta_L, \theta_H\}$ at short horizons,
so $|\varphi^{\textsc{unc}}_{\text{geo}}| \ll
|\varphi^{\textsc{unc}}_{\text{geoeco}}|$. The cascade from geopolitical
tension into policy-regime uncertainty (sanctions $\to$ fiscal response
$\to$ regime ambiguity) operates at longer horizons, outside our event
windows.

The channel also generates the \emph{originator penalty} of
\citet{PastorVeronesi2013}, but through a different mechanism: the originator
faces a dual burden as its posterior $\pi_{US,t}$ rises from domestic policy
uncertainty while the interaction with domestic sentiment amplifies spread
widening through the Local channel---even as the GFC compresses spreads abroad.

\paragraph{Channel~IV: Local amplification.}

Domestic fundamentals amplify or absorb external shocks through a threshold
process:
\begin{equation}
  L_{i,t}
  = \beta_L\, X^{\text{macro}}_{i,t}
  + \gamma_L\, X^{\text{macro}}_{i,t}
    \cdot \mathbf{1}\!\{F_{i,t} > \bar{F}\}\,,
  \label{eq:local_channel}
\end{equation}
where $X^{\text{macro}}_{i,t}$ stacks domestic sentiment, interest-rate
expectations, political tensions, and regional structure---daily-frequency
proxies for the core determinants of debt sustainability. Below the fragility
threshold, shocks pass through with modest amplification ($\beta_L$). Above
it, amplification jumps to $\beta_L + \gamma_L$, capturing sovereign--bank
doom loops, capital flight, or loss of market access
\citep{FarhiTirole2018}.

The threshold structure delivers Prediction~L: under geopolitical shocks,
the Direct and GFC channels dominate the panel mean, so
$\mathbb{E}_i[|\varphi^{\textsc{loc}}|] \ll
\mathbb{E}_i[|\varphi^{\textsc{dir}}|]$. Under geoeconomic shocks,
bilateral heterogeneity in $X^{\text{macro}}_{i,t}$ and the threshold
indicator make the Local channel a primary differentiator across countries.

\paragraph{The nonlinear pricing function.}

Combining the four channels, the spread is
\begin{equation}
  s_{i,t}
  = m\!\bigl(
    \underbrace{\alpha(d_i)\,g_t,\, F_{i,t}}_{\text{Direct}},\;
    \underbrace{\xi(W_t),\, \lambda_i}_{\text{GFC}},\;
    \underbrace{\mathcal{H}(\pi_{i,t}),\, \kappa_i}_{\text{Uncertainty}},\;
    \underbrace{X^{\text{macro}}_{i,t},\, \mathbf{1}\{F_{i,t}>\bar{F}\}}_{\text{Local}}
  \bigr)
  + \mu_i + \tau_t + \varepsilon_{i,t}\,,
  \label{eq:m_function}
\end{equation}
where $m(\cdot)$ is an unknown function encoding three sources of nonlinearity:
(i)~threshold effects through $\mathbf{1}\{F_{i,t} > \bar{F}\}$;
(ii)~interaction effects through the dependence of $G_{i,t}$ on $\xi(W_t)$,
which interacts with $g_t$ via intermediary losses;
(iii)~convexity through $\mathcal{H}(\pi_{i,t})$.
The illustrative expansion
\begin{equation}
  s_{i,t}
  =
  \mu_i+\tau_t
  + \beta_D\, \alpha(d_i)\,g_t
  + \beta_G\, \xi(W_t)
  + \beta_U\, \mathcal{H}(\pi_{i,t})
  + \beta_L\, X^{\text{macro}}_{i,t}
  + \gamma_1\, \alpha(d_i)\,g_t\,\mathbf{1}\{F_{i,t}>\bar{F}\}
  + \gamma_2\, g_t\,\xi(W_t)
  + \gamma_3\, \bigl[\mathcal{H}(\pi_{i,t})\bigr]^2
  + u_{i,t}
  \label{eq:illustrative_nonlinear}
\end{equation}
shows that even a parsimonious approximation generates interaction and
higher-order terms that a linear panel model cannot capture, motivating the
use of flexible ML estimators to approximate $m(\cdot)$.

\paragraph{Summary of predictions.}

The framework generates four predictions that correspond to the dominance
benchmarks of Table~\ref{tab:benchmarks}:

\begin{enumerate}[label=\textbf{P\arabic*.}, leftmargin=2em, itemsep=4pt]

\item \textbf{Gravity decay} (Prediction~D).
  The concavity of $\alpha(d_i)$ implies that $\varphi^{\textsc{dir}}_{i,t}$
  decays with log-distance to the conflict epicenter.
  \emph{Test}: Figure~\ref{fig:gravity}.

\item \textbf{Scissors and GFC dominance} (Predictions D and G).
  Equation~\eqref{eq:scissors_condition} delivers
  $\varphi^{\textsc{dir}} > 0$ and $\varphi^{\textsc{gfc}} < 0$ under
  geopolitical shocks, with
  $|\varphi^{\textsc{gfc}}| < |\varphi^{\textsc{dir}}|$.
  Geoeconomic shocks bypass the Direct channel and transmit primarily through
  the GFC, so $|\varphi^{\textsc{gfc}}| > |\varphi^{\textsc{dir}}|$.
  \emph{Test}: Table~\ref{tab:causal_LP}, Figure~\ref{fig:channels_taxonomy}.

\item \textbf{Broad-based uncertainty activation} (Prediction~U).
  Because $\pi_{i,t}$ updates on a common geoeconomic signal,
  $\varphi^{\textsc{unc}}$ activates broadly under geoeconomic shocks. Under
  geopolitical shocks, $|\varphi^{\textsc{unc}}_{\text{geo}}| \ll
  |\varphi^{\textsc{unc}}_{\text{geoeco}}|$ at short horizons.
  \emph{Test}: cross-country dispersion across episodes.

\item \textbf{Nonlinear amplification and originator penalty} (Prediction~L).
  The threshold $\bar{F}$ produces qualitatively different transmission for
  high- vs.\ low-fragility sovereigns. Under geopolitical shocks,
  $\mathbb{E}_i[|\varphi^{\textsc{loc}}|] \ll
  \mathbb{E}_i[|\varphi^{\textsc{dir}}|]$; under geoeconomic shocks,
  $\varphi^{\textsc{loc}}$ is a primary channel, and the originator faces
  widening spreads through it even as the GFC compresses spreads abroad.
  \emph{Test}: Table~\ref{tab:impact_persistence}.

\end{enumerate}

\noindent
The framework nests \citet{PastorVeronesi2012,PastorVeronesi2013} as special
cases---our Direct, Uncertainty, and GFC channels correspond to PV's impact
component, political uncertainty premium, and the intermediary response absent
in their representative-agent setting---and connects to
\citet{ClaytonMaggioriSchreger2026}, whose vulnerability--leverage model
provides the structural foundation for the strategic interaction sustaining
$\theta_H$ in Channel~III.

%% ============================================================
%% APPENDIX B — Shapley Dependence and Interaction Analysis
%% ============================================================
\clearpage

\section{Appendix. Country-Level Transmission Dynamics}\phantomsection\label{app:Figures}

\begin{figure}[ht!]
    \setlength{\belowcaptionskip}{-5pt}
    \centering
    \caption{\textbf{Russia–Ukraine Invasion: Country-Level Shapley Dynamics}}
    \vspace{2pt}
    \includegraphics[width=0.7\textwidth]{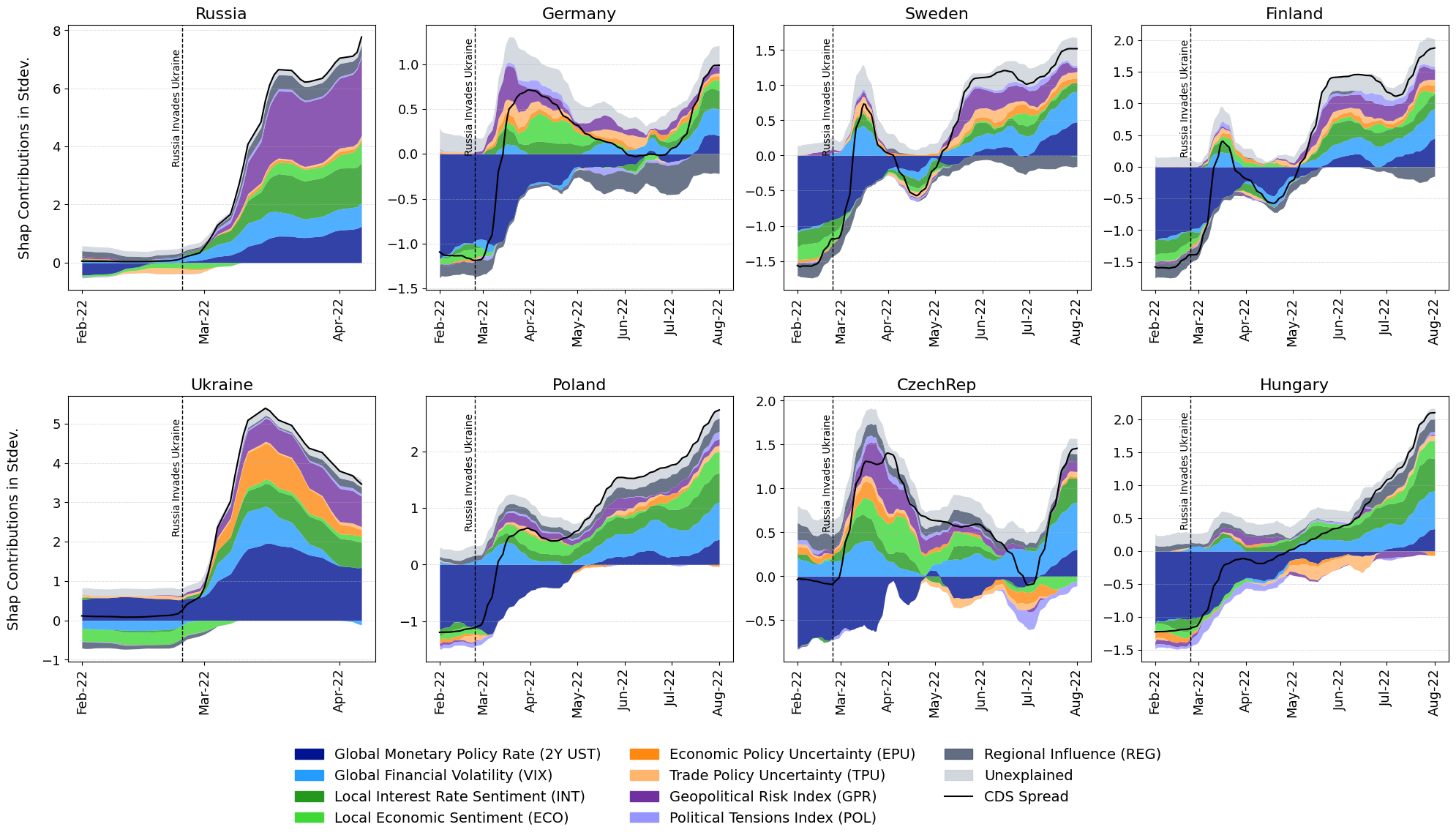}
    \label{fig:episode_RU}
    \begin{flushleft}
       \scriptsize
        \vspace{-5pt}
        Notes: Each panel plots the evolution of Shapley contributions to sovereign CDS spreads for selected countries before and after February~2022. Stacked areas represent individual predictor contributions; black line is the observed CDS spread. For Russia and Ukraine, contributions from GPR, global financial volatility, and domestic conditions surge simultaneously. For European economies, the initial geopolitical impulse is followed by a cascade through domestic inflation, local rate tightening, and global financial tightening. All series are 28-day moving averages.
    \end{flushleft}
\end{figure}

Country-level Shapley attribution dynamics for the remaining three episodes (Hamas--Israel, U.S.\ Election, Liberation Day) are available in the replication package.

\begin{figure}[ht!]
    \centering
    \caption{\textbf{Gravity Validation: Direct Channel vs.\ Regional Component — Russia--Ukraine (Feb.\ 2022)}}
   
    \includegraphics[width=0.8\textwidth]{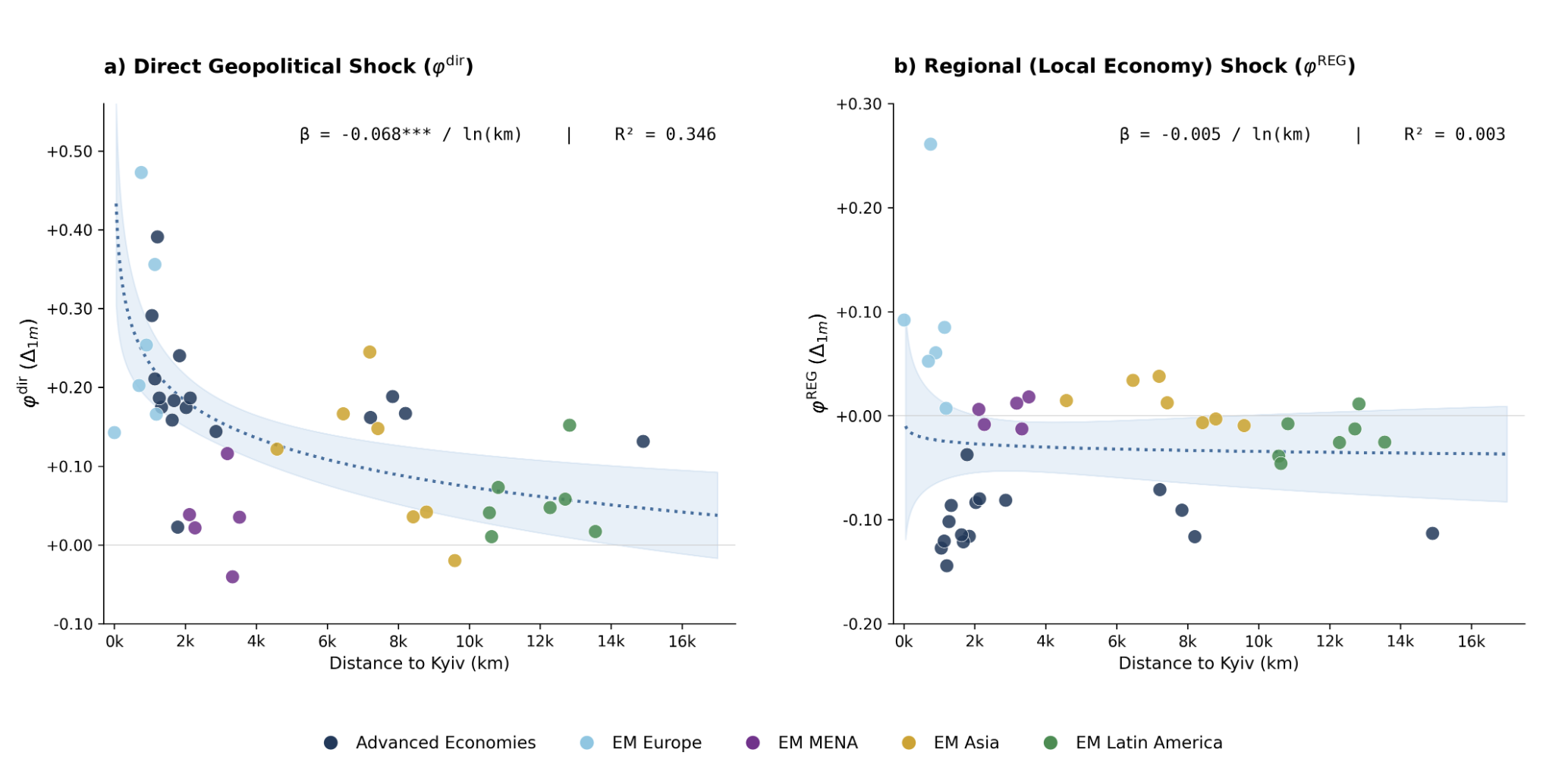}
    \label{fig:gravity_direct_vs_REG}
    \begin{flushleft}
       \scriptsize
        Notes: Panel~(a) plots the one-month Direct-channel response $\varphi^{\mathrm{dir}}_{i}(\Delta_{1m})$ and panel~(b) the REG-within-LOC Shapley value $\varphi^{\mathrm{REG}}_{i}(\Delta_{1m})$ against geodesic distance from each country's capital to Kyiv. The Direct channel exhibits significant gravity-like decay ($\hat{\beta} = -0.068^{***}$, $R^2 = 0.346$), consistent with geopolitical shocks repricing default risk in proportion to conflict proximity. The regional component---which captures the Shapley--Taylor attribution of the regional group indicators (Advanced, EM~Europe, EM~MENA, EM~Asia, EM~Latin America) absorbed into the Local channel---shows no distance relationship ($\hat{\beta} = -0.005$, $R^2 = 0.003$, $p = 0.72$), confirming that it does not proxy for geographic proximity. This contrast validates the absorption of REG into the Local channel rather than the creation of a fifth transmission channel. Dotted lines: OLS log-distance fit from eq.~(\ref{eq:gravity_ols}); shaded bands: 95\% confidence intervals. $^{***}\,p<0.01$, $^{**}\,p<0.05$, $^{*}\,p<0.10$. OLS includes all countries with available CDS data (excluding Ukraine at $d=0$).
    \end{flushleft}
\end{figure}

\clearpage
\section{Appendix. Country-Level Channel Decomposition Tables}\phantomsection\label{app:Transmission}

\begin{table}[ht]
\caption{\textbf{Geopolitical Episodes} --- Country-level channel decomposition ($\Delta_{1\text{m}}$)}
\label{tab:Geo_Country_combined}
\centering
\scriptsize
\setlength{\tabcolsep}{3pt}
\hspace*{20pt}
\begin{tabular}{p{2.8cm}
*{5}{>{\centering\arraybackslash}p{0.8cm}}p{0mm}*{5}{>{\centering\arraybackslash}p{0.8cm}}@{}}
\toprule
& \multicolumn{5}{c}{\textbf{Russia--Ukraine (Feb.\ 2022)}} & & \multicolumn{5}{c}{\textbf{Hamas--Israel (Oct.\ 2023)}} \\
\cmidrule(lr){2-6}\cmidrule(lr){8-12}
\textbf{Country}
  & $\varphi^{\rm dir}$ & $\varphi^{\rm GFC}$ & $\varphi^{\rm UNC}$ & $\varphi^{\rm LOC}$ & $\varphi^{\rm tot}$
  & & $\varphi^{\rm dir}$ & $\varphi^{\rm GFC}$ & $\varphi^{\rm UNC}$ & $\varphi^{\rm LOC}$ & $\varphi^{\rm tot}$ \\
\midrule
\multicolumn{12}{l}{\textit{Advanced Economies}} \\
\textit{Group avg.} & \cellcolor{red!15}+.15 & \cellcolor{green!15}$-$.15 & \cellcolor{red!8}+.01 & \cellcolor{green!15}$-$.17 & \cellcolor{green!15}$-$.16 & & \cellcolor{red!15}+.09 & \cellcolor{green!8}$-$.02 & \cellcolor{gray!10}.00 & \cellcolor{green!15}$-$.06 & \cellcolor{red!8}+.01 \\
USA & \cellcolor{red!15}+.14 & \cellcolor{green!15}$-$.12 & \cellcolor{gray!10}.00 & \cellcolor{green!15}$-$.17 & \cellcolor{green!15}$-$.14 & & \cellcolor{red!15}+.08 & \cellcolor{green!15}$-$.12 & \cellcolor{green!8}$-$.03 & \cellcolor{green!8}$-$.01 & \cellcolor{green!15}$-$.08 \\
Canada & \cellcolor{red!15}+.09 & \cellcolor{green!8}$-$.04 & \cellcolor{gray!10}.00 & \cellcolor{green!15}$-$.20 & \cellcolor{green!15}$-$.15 & & \cellcolor{red!15}+.10 & \cellcolor{green!15}$-$.09 & \cellcolor{green!8}$-$.02 & \cellcolor{green!15}$-$.15 & \cellcolor{green!15}$-$.15 \\
Australia & \cellcolor{red!15}+.10 & \cellcolor{green!15}$-$.18 & \cellcolor{red!8}+.02 & \cellcolor{green!15}$-$.21 & \cellcolor{green!15}$-$.27 & & \cellcolor{red!15}+.08 & \cellcolor{green!15}$-$.20 & \cellcolor{green!8}$-$.04 & \cellcolor{green!15}$-$.09 & \cellcolor{green!15}$-$.26 \\
UK & \cellcolor{red!15}+.15 & \cellcolor{green!15}$-$.14 & \cellcolor{gray!10}.00 & \cellcolor{green!15}$-$.19 & \cellcolor{green!15}$-$.17 & & \cellcolor{red!15}+.11 & \cellcolor{red!15}+.08 & \cellcolor{red!8}+.02 & \cellcolor{green!8}$-$.04 & \cellcolor{red!15}+.18 \\
Germany & \cellcolor{red!15}+.25 & \cellcolor{green!15}$-$.24 & \cellcolor{red!8}+.02 & \cellcolor{green!8}$-$.03 & \cellcolor{gray!10}.00 & & \cellcolor{red!15}+.09 & \cellcolor{green!15}$-$.15 & \cellcolor{green!8}$-$.02 & \cellcolor{gray!10}.00 & \cellcolor{green!15}$-$.08 \\
France & \cellcolor{red!15}+.14 & \cellcolor{green!15}$-$.08 & \cellcolor{red!8}+.02 & \cellcolor{green!15}$-$.16 & \cellcolor{green!15}$-$.07 & & \cellcolor{red!15}+.16 & \cellcolor{green!8}$-$.04 & \cellcolor{green!8}$-$.02 & \cellcolor{green!15}$-$.11 & \cellcolor{gray!10}.00 \\
Italy & \cellcolor{red!15}+.12 & \cellcolor{green!15}$-$.28 & \cellcolor{red!8}+.04 & \cellcolor{green!15}$-$.39 & \cellcolor{green!15}$-$.50 & & \cellcolor{red!15}+.12 & \cellcolor{gray!10}.00 & \cellcolor{green!8}$-$.02 & \cellcolor{green!15}$-$.12 & \cellcolor{green!8}$-$.01 \\
Spain & \cellcolor{red!15}+.11 & \cellcolor{green!15}$-$.11 & \cellcolor{red!8}+.01 & \cellcolor{green!15}$-$.08 & \cellcolor{green!15}$-$.07 & & \cellcolor{red!15}+.15 & \cellcolor{green!8}$-$.04 & \cellcolor{gray!10}.00 & \cellcolor{green!15}$-$.17 & \cellcolor{green!15}$-$.07 \\
Netherlands & \cellcolor{green!8}$-$.03 & \cellcolor{green!8}$-$.03 & \cellcolor{green!8}$-$.01 & \cellcolor{green!15}$-$.14 & \cellcolor{green!15}$-$.22 & & \cellcolor{red!15}+.06 & \cellcolor{red!15}+.10 & \cellcolor{red!8}+.03 & \cellcolor{green!8}$-$.04 & \cellcolor{red!15}+.15 \\
Belgium & \cellcolor{red!15}+.17 & \cellcolor{green!15}$-$.17 & \cellcolor{red!8}+.01 & \cellcolor{green!15}$-$.15 & \cellcolor{green!15}$-$.14 & & \cellcolor{red!15}+.14 & \cellcolor{green!8}$-$.01 & \cellcolor{red!8}+.02 & \cellcolor{green!15}$-$.12 & \cellcolor{red!8}+.04 \\
Austria & \cellcolor{red!15}+.23 & \cellcolor{green!15}$-$.26 & \cellcolor{red!8}+.01 & \cellcolor{green!15}$-$.12 & \cellcolor{green!15}$-$.14 & & \cellcolor{red!15}+.07 & \cellcolor{green!15}$-$.09 & \cellcolor{red!8}+.01 & \cellcolor{green!8}$-$.03 & \cellcolor{green!8}$-$.04 \\
Denmark & \cellcolor{red!15}+.17 & \cellcolor{green!15}$-$.12 & \cellcolor{red!8}+.01 & \cellcolor{green!15}$-$.08 & \cellcolor{green!8}$-$.02 & & \cellcolor{green!8}$-$.02 & \cellcolor{red!8}+.03 & \cellcolor{red!8}+.02 & \cellcolor{red!8}+.01 & \cellcolor{red!8}+.04 \\
Finland & \cellcolor{red!15}+.24 & \cellcolor{green!15}$-$.12 & \cellcolor{green!8}$-$.01 & \cellcolor{green!15}$-$.16 & \cellcolor{green!15}$-$.06 & & \cellcolor{red!15}+.08 & \cellcolor{red!8}+.05 & \cellcolor{red!8}+.02 & \cellcolor{green!8}$-$.01 & \cellcolor{red!15}+.14 \\
Norway & \cellcolor{red!15}+.16 & \cellcolor{green!15}$-$.18 & \cellcolor{green!8}$-$.02 & \cellcolor{green!15}$-$.17 & \cellcolor{green!15}$-$.20 & & \cellcolor{red!15}+.07 & \cellcolor{red!15}+.06 & \cellcolor{red!8}+.02 & \cellcolor{green!15}$-$.09 & \cellcolor{red!15}+.06 \\
Sweden & \cellcolor{red!15}+.20 & \cellcolor{green!15}$-$.12 & \cellcolor{red!8}+.01 & \cellcolor{green!15}$-$.19 & \cellcolor{green!15}$-$.10 & & \cellcolor{red!8}+.03 & \cellcolor{red!15}+.08 & \cellcolor{red!8}+.04 & \cellcolor{green!8}$-$.01 & \cellcolor{red!15}+.14 \\
Japan & \cellcolor{red!15}+.14 & \cellcolor{green!15}$-$.14 & \cellcolor{green!8}$-$.01 & \cellcolor{green!15}$-$.23 & \cellcolor{green!15}$-$.24 & & \cellcolor{red!15}+.08 & \cellcolor{red!8}+.04 & \cellcolor{green!8}$-$.01 & \cellcolor{green!8}$-$.01 & \cellcolor{red!15}+.10 \\
\addlinespace[2pt]
\multicolumn{12}{l}{\textit{Emerging Europe}} \\
\textit{Group avg.} & \cellcolor{red!15}+.24 & \cellcolor{red!15}+.22 & \cellcolor{red!8}+.05 & \cellcolor{red!15}+.52 & \cellcolor{red!15}+1.03 & & \cellcolor{red!8}+.05 & \cellcolor{green!8}$-$.02 & \cellcolor{gray!10}.00 & \cellcolor{green!8}$-$.03 & \cellcolor{gray!10}.00 \\
Russia & \cellcolor{red!15}+.52 & \cellcolor{red!15}+1.55 & \cellcolor{red!15}+.31 & \cellcolor{red!15}+2.23 & \cellcolor{red!15}+4.61 & & --- & --- & --- & --- & --- \\
Ukraine & \cellcolor{red!15}+.13 & \cellcolor{red!15}+.66 & \cellcolor{green!8}$-$.02 & \cellcolor{red!15}+.64 & \cellcolor{red!15}+1.41 & & --- & --- & --- & --- & --- \\
Poland & \cellcolor{red!15}+.15 & \cellcolor{green!15}$-$.10 & \cellcolor{red!8}+.04 & \cellcolor{red!15}+.11 & \cellcolor{red!15}+.19 & & \cellcolor{green!8}$-$.03 & \cellcolor{red!15}+.06 & \cellcolor{red!8}+.01 & \cellcolor{gray!10}.00 & \cellcolor{red!15}+.05 \\
Czech Rep. & \cellcolor{red!15}+.32 & \cellcolor{green!15}$-$.35 & \cellcolor{green!8}$-$.01 & \cellcolor{red!15}+.12 & \cellcolor{red!15}+.08 & & \cellcolor{red!8}+.05 & \cellcolor{red!8}+.04 & \cellcolor{gray!10}.00 & \cellcolor{red!8}+.02 & \cellcolor{red!15}+.11 \\
Hungary & \cellcolor{red!15}+.21 & \cellcolor{green!15}$-$.27 & \cellcolor{green!8}$-$.02 & \cellcolor{red!15}+.07 & \cellcolor{green!8}$-$.01 & & \cellcolor{red!15}+.11 & \cellcolor{green!8}$-$.03 & \cellcolor{green!8}$-$.01 & \cellcolor{green!15}$-$.09 & \cellcolor{green!8}$-$.02 \\
Turkey & \cellcolor{red!15}+.13 & \cellcolor{green!15}$-$.20 & \cellcolor{green!8}$-$.01 & \cellcolor{green!8}$-$.03 & \cellcolor{green!15}$-$.12 & & \cellcolor{red!15}+.05 & \cellcolor{green!15}$-$.14 & \cellcolor{gray!10}.00 & \cellcolor{green!15}$-$.06 & \cellcolor{green!15}$-$.15 \\
\addlinespace[2pt]
\multicolumn{12}{l}{\textit{Middle East \& North Africa}} \\
\textit{Group avg.} & \cellcolor{red!8}+.03 & \cellcolor{green!15}$-$.08 & \cellcolor{green!8}$-$.02 & \cellcolor{green!15}$-$.14 & \cellcolor{green!15}$-$.22 & & \cellcolor{red!15}+.17 & \cellcolor{red!15}+.12 & \cellcolor{red!15}+.07 & \cellcolor{red!15}+.13 & \cellcolor{red!15}+.50 \\
Egypt & \cellcolor{red!8}+.01 & \cellcolor{green!15}$-$.09 & \cellcolor{green!8}$-$.02 & \cellcolor{green!15}$-$.17 & \cellcolor{green!15}$-$.26 & & \cellcolor{red!15}+.27 & \cellcolor{red!15}+.08 & \cellcolor{red!15}+.08 & \cellcolor{red!15}+.16 & \cellcolor{red!15}+.59 \\
Israel & \cellcolor{red!15}+.08 & \cellcolor{green!15}$-$.06 & \cellcolor{green!8}$-$.02 & \cellcolor{green!15}$-$.10 & \cellcolor{green!15}$-$.10 & & \cellcolor{red!15}+.21 & \cellcolor{red!15}+.32 & \cellcolor{red!15}+.19 & \cellcolor{red!15}+.45 & \cellcolor{red!15}+1.17 \\
Morocco & \cellcolor{green!8}$-$.01 & \cellcolor{green!15}$-$.09 & \cellcolor{green!8}$-$.03 & \cellcolor{green!15}$-$.10 & \cellcolor{green!15}$-$.24 & & \cellcolor{red!15}+.06 & \cellcolor{red!15}+.17 & \cellcolor{red!8}+.04 & \cellcolor{green!15}$-$.06 & \cellcolor{red!15}+.20 \\
Qatar & \cellcolor{green!8}$-$.04 & \cellcolor{green!15}$-$.06 & \cellcolor{gray!10}.00 & \cellcolor{green!15}$-$.19 & \cellcolor{green!15}$-$.28 & & \cellcolor{red!15}+.20 & \cellcolor{red!15}+.16 & \cellcolor{red!8}+.04 & \cellcolor{gray!10}.00 & \cellcolor{red!15}+.39 \\
Saudi Arabia & \cellcolor{red!15}+.08 & \cellcolor{green!15}$-$.12 & \cellcolor{green!8}$-$.03 & \cellcolor{green!15}$-$.17 & \cellcolor{green!15}$-$.23 & & \cellcolor{red!15}+.14 & \cellcolor{green!15}$-$.11 & \cellcolor{gray!10}.00 & \cellcolor{red!15}+.12 & \cellcolor{red!15}+.14 \\
\addlinespace[2pt]
\multicolumn{12}{l}{\textit{Asia-Pacific}} \\
\textit{Group avg.} & \cellcolor{red!15}+.08 & \cellcolor{green!15}$-$.14 & \cellcolor{green!8}$-$.01 & \cellcolor{green!8}$-$.01 & \cellcolor{green!15}$-$.07 & & \cellcolor{red!15}+.06 & \cellcolor{green!15}$-$.09 & \cellcolor{gray!10}.00 & \cellcolor{green!15}$-$.06 & \cellcolor{green!15}$-$.10 \\
China & \cellcolor{red!15}+.12 & \cellcolor{green!15}$-$.17 & \cellcolor{gray!10}.00 & \cellcolor{red!15}+.06 & \cellcolor{red!8}+.01 & & \cellcolor{green!8}$-$.02 & \cellcolor{green!15}$-$.46 & \cellcolor{green!15}$-$.05 & \cellcolor{green!15}$-$.13 & \cellcolor{green!15}$-$.66 \\
India & \cellcolor{red!15}+.14 & \cellcolor{green!15}$-$.16 & \cellcolor{green!8}$-$.01 & \cellcolor{green!8}$-$.04 & \cellcolor{green!15}$-$.07 & & \cellcolor{green!8}$-$.03 & \cellcolor{green!15}$-$.08 & \cellcolor{gray!10}.00 & \cellcolor{red!8}+.02 & \cellcolor{green!15}$-$.10 \\
Indonesia & \cellcolor{green!8}$-$.04 & \cellcolor{green!8}$-$.02 & \cellcolor{red!8}+.01 & \cellcolor{green!15}$-$.07 & \cellcolor{green!15}$-$.13 & & \cellcolor{red!15}+.09 & \cellcolor{green!8}$-$.02 & \cellcolor{gray!10}.00 & \cellcolor{green!15}$-$.09 & \cellcolor{green!8}$-$.02 \\
Malaysia & \cellcolor{gray!10}.00 & \cellcolor{green!15}$-$.12 & \cellcolor{gray!10}.00 & \cellcolor{green!15}$-$.09 & \cellcolor{green!15}$-$.20 & & \cellcolor{gray!10}.00 & \cellcolor{green!15}$-$.07 & \cellcolor{red!8}+.01 & \cellcolor{gray!10}.00 & \cellcolor{green!15}$-$.07 \\
Philippines & \cellcolor{gray!10}.00 & \cellcolor{green!8}$-$.02 & \cellcolor{red!8}+.03 & \cellcolor{green!8}$-$.03 & \cellcolor{green!8}$-$.03 & & \cellcolor{red!15}+.20 & \cellcolor{red!8}+.02 & \cellcolor{red!8}+.02 & \cellcolor{green!15}$-$.19 & \cellcolor{red!8}+.04 \\
Thailand & \cellcolor{red!15}+.14 & \cellcolor{green!15}$-$.14 & \cellcolor{green!8}$-$.01 & \cellcolor{red!15}+.07 & \cellcolor{red!15}+.06 & & \cellcolor{red!15}+.10 & \cellcolor{red!15}+.06 & \cellcolor{red!8}+.01 & \cellcolor{green!15}$-$.05 & \cellcolor{red!15}+.11 \\
Vietnam & \cellcolor{red!15}+.19 & \cellcolor{green!15}$-$.33 & \cellcolor{green!8}$-$.05 & \cellcolor{red!8}+.04 & \cellcolor{green!15}$-$.15 & & \cellcolor{red!15}+.05 & \cellcolor{green!15}$-$.10 & \cellcolor{red!8}+.02 & \cellcolor{green!8}$-$.01 & \cellcolor{green!8}$-$.04 \\
\addlinespace[2pt]
\multicolumn{12}{l}{\textit{Latin America}} \\
\textit{Group avg.} & \cellcolor{red!8}+.03 & \cellcolor{green!15}$-$.11 & \cellcolor{green!8}$-$.01 & \cellcolor{green!8}$-$.04 & \cellcolor{green!15}$-$.13 & & \cellcolor{red!8}+.02 & \cellcolor{green!8}$-$.01 & \cellcolor{green!8}$-$.01 & \cellcolor{green!8}$-$.01 & \cellcolor{green!8}$-$.01 \\
Argentina & \cellcolor{red!15}+.10 & \cellcolor{green!15}$-$.14 & \cellcolor{green!8}$-$.03 & \cellcolor{green!8}$-$.03 & \cellcolor{green!15}$-$.10 & & \cellcolor{red!8}+.04 & \cellcolor{green!8}$-$.02 & \cellcolor{green!8}$-$.01 & \cellcolor{green!15}$-$.05 & \cellcolor{green!8}$-$.04 \\
Brazil & \cellcolor{green!8}$-$.03 & \cellcolor{green!15}$-$.14 & \cellcolor{green!8}$-$.03 & \cellcolor{green!15}$-$.11 & \cellcolor{green!15}$-$.31 & & \cellcolor{green!8}$-$.04 & \cellcolor{green!8}$-$.04 & \cellcolor{green!8}$-$.03 & \cellcolor{red!8}+.03 & \cellcolor{green!15}$-$.08 \\
Chile & \cellcolor{gray!10}.00 & \cellcolor{green!8}$-$.04 & \cellcolor{gray!10}.00 & \cellcolor{green!15}$-$.06 & \cellcolor{green!15}$-$.11 & & \cellcolor{red!15}+.08 & \cellcolor{red!8}+.05 & \cellcolor{gray!10}.00 & \cellcolor{green!8}$-$.03 & \cellcolor{red!15}+.09 \\
Colombia & \cellcolor{gray!10}.00 & \cellcolor{green!15}$-$.12 & \cellcolor{green!8}$-$.04 & \cellcolor{green!8}$-$.05 & \cellcolor{green!15}$-$.21 & & \cellcolor{gray!10}.00 & \cellcolor{gray!10}.00 & \cellcolor{green!8}$-$.02 & \cellcolor{green!8}$-$.03 & \cellcolor{green!15}$-$.05 \\
Mexico & \cellcolor{red!8}+.05 & \cellcolor{green!15}$-$.09 & \cellcolor{red!8}+.01 & \cellcolor{green!15}$-$.11 & \cellcolor{green!15}$-$.14 & & \cellcolor{red!15}+.06 & \cellcolor{green!15}$-$.10 & \cellcolor{green!8}$-$.03 & \cellcolor{green!8}$-$.01 & \cellcolor{green!15}$-$.09 \\
Peru & \cellcolor{red!15}+.06 & \cellcolor{green!15}$-$.09 & \cellcolor{gray!10}.00 & \cellcolor{red!15}+.05 & \cellcolor{red!8}+.03 & & \cellcolor{red!8}+.05 & \cellcolor{green!8}$-$.02 & \cellcolor{green!8}$-$.02 & \cellcolor{gray!10}.00 & \cellcolor{red!8}+.01 \\
Uruguay & \cellcolor{red!8}+.03 & \cellcolor{green!15}$-$.11 & \cellcolor{green!8}$-$.02 & \cellcolor{red!8}+.03 & \cellcolor{green!15}$-$.08 & & \cellcolor{green!8}$-$.01 & \cellcolor{red!15}+.07 & \cellcolor{red!8}+.01 & \cellcolor{green!8}$-$.01 & \cellcolor{red!15}+.06 \\
\addlinespace[2pt]
\bottomrule
\end{tabular}
\vspace{4pt}\newline{\scriptsize
\colorbox{red!15}{\phantom{X}} $\Delta>+0.05$\quad
\colorbox{red!8}{\phantom{X}} $+0.005<\Delta\leq+0.05$\quad
\colorbox{gray!10}{\phantom{X}} $|\Delta|\leq0.005$\quad
\colorbox{green!8}{\phantom{X}} $-0.05\leq\Delta<-0.005$\quad
\colorbox{green!15}{\phantom{X}} $\Delta<-0.05$.
Russia/Ukraine excl.\ from Hamas--Israel (no pre-event data).
$\varphi^{\rm LOC}$ includes Region cross-SHAP.}
\end{table}

% ---- TABLE 2: GEOECONOMIC EPISODES ----
\begin{table}[ht]
\caption{\textbf{Geoeconomic Episodes} --- Country-level channel decomposition ($\Delta_{1\text{m}}$)}
\label{tab:Econ_Country_combined}
\centering
\scriptsize
\setlength{\tabcolsep}{3pt}
\hspace*{20pt}
\begin{tabular}{p{2.8cm}*{5}{>{\centering\arraybackslash}p{0.8cm}}p{0mm}*{5}{>{\centering\arraybackslash}p{0.8cm}}@{}}
\toprule
& \multicolumn{5}{c}{\textbf{Trump Election (Nov.\ 2024)}} & & \multicolumn{5}{c}{\textbf{Liberation Day (Apr.\ 2025)}} \\
\cmidrule(lr){2-6}\cmidrule(lr){8-12}
\textbf{Country}
  & $\varphi^{\rm dir}$ & $\varphi^{\rm GFC}$ & $\varphi^{\rm UNC}$ & $\varphi^{\rm LOC}$ & $\varphi^{\rm tot}$
  & & $\varphi^{\rm dir}$ & $\varphi^{\rm GFC}$ & $\varphi^{\rm UNC}$ & $\varphi^{\rm LOC}$ & $\varphi^{\rm tot}$ \\
\midrule
\multicolumn{12}{l}{\textit{USMCA}} \\
\textit{Group avg.} & \cellcolor{red!8}+.04 & \cellcolor{green!15}$-$.07 & \cellcolor{gray!10}.00 & \cellcolor{red!8}+.02 & \cellcolor{green!8}$-$.01 & & --- & --- & --- & --- & --- \\
USA & \cellcolor{red!8}+.04 & \cellcolor{green!15}$-$.08 & \cellcolor{red!8}+.01 & \cellcolor{red!15}+.06 & \cellcolor{red!8}+.03 & & \cellcolor{gray!10}.00 & \cellcolor{green!8}$-$.02 & \cellcolor{gray!10}.00 & \cellcolor{red!15}+.07 & \cellcolor{red!15}+.05 \\
Canada & \cellcolor{red!8}+.05 & \cellcolor{green!15}$-$.10 & \cellcolor{green!8}$-$.01 & \cellcolor{gray!10}.00 & \cellcolor{green!15}$-$.07 & & \cellcolor{red!8}+.01 & \cellcolor{red!15}+.11 & \cellcolor{red!15}+.07 & \cellcolor{red!15}+.22 & \cellcolor{red!15}+.42 \\
Mexico & \cellcolor{red!8}+.03 & \cellcolor{green!8}$-$.03 & \cellcolor{red!8}+.01 & \cellcolor{red!8}+.01 & \cellcolor{red!8}+.01 & & \cellcolor{red!8}+.02 & \cellcolor{red!15}+.06 & \cellcolor{red!15}+.10 & \cellcolor{red!15}+.14 & \cellcolor{red!15}+.32 \\
\addlinespace[2pt]
\multicolumn{12}{l}{\textit{Advanced Economies}} \\
\textit{Group avg.} & \cellcolor{red!8}+.04 & \cellcolor{green!15}$-$.05 & \cellcolor{red!8}+.04 & \cellcolor{red!8}+.05 & \cellcolor{red!15}+.08 & & \cellcolor{gray!10}.00 & \cellcolor{green!8}$-$.05 & \cellcolor{gray!10}.00 & \cellcolor{gray!10}.00 & \cellcolor{green!8}$-$.05 \\
Australia & \cellcolor{red!8}+.04 & \cellcolor{green!15}$-$.12 & \cellcolor{red!8}+.05 & \cellcolor{red!8}+.03 & \cellcolor{green!8}$-$.01 & & \cellcolor{green!8}$-$.02 & \cellcolor{green!15}$-$.07 & \cellcolor{green!8}$-$.03 & \cellcolor{red!8}+.03 & \cellcolor{green!15}$-$.10 \\
UK & \cellcolor{red!8}+.04 & \cellcolor{green!15}$-$.05 & \cellcolor{red!8}+.03 & \cellcolor{red!8}+.01 & \cellcolor{red!8}+.02 & & \cellcolor{gray!10}.00 & \cellcolor{green!15}$-$.07 & \cellcolor{red!8}+.01 & \cellcolor{red!8}+.03 & \cellcolor{green!8}$-$.03 \\
Germany & \cellcolor{red!8}+.04 & \cellcolor{gray!10}.00 & \cellcolor{red!15}+.06 & \cellcolor{red!15}+.08 & \cellcolor{red!15}+.19 & & \cellcolor{gray!10}.00 & \cellcolor{green!15}$-$.05 & \cellcolor{red!8}+.01 & \cellcolor{green!8}$-$.01 & \cellcolor{green!15}$-$.05 \\
France & \cellcolor{red!8}+.03 & \cellcolor{red!15}+.14 & \cellcolor{red!8}+.05 & \cellcolor{red!15}+.16 & \cellcolor{red!15}+.38 & & \cellcolor{red!8}+.01 & \cellcolor{gray!10}.00 & \cellcolor{red!8}+.01 & \cellcolor{gray!10}.00 & \cellcolor{red!8}+.02 \\
Italy & \cellcolor{red!8}+.04 & \cellcolor{green!15}$-$.15 & \cellcolor{red!8}+.05 & \cellcolor{green!8}$-$.01 & \cellcolor{green!15}$-$.07 & & \cellcolor{red!8}+.01 & \cellcolor{green!15}$-$.05 & \cellcolor{red!8}+.01 & \cellcolor{green!8}$-$.03 & \cellcolor{green!15}$-$.07 \\
Spain & \cellcolor{red!8}+.02 & \cellcolor{green!15}$-$.11 & \cellcolor{red!8}+.01 & \cellcolor{red!8}+.02 & \cellcolor{green!15}$-$.06 & & \cellcolor{red!8}+.01 & \cellcolor{green!8}$-$.02 & \cellcolor{green!8}$-$.01 & \cellcolor{red!8}+.02 & \cellcolor{gray!10}.00 \\
Netherlands & \cellcolor{red!15}+.05 & \cellcolor{green!15}$-$.10 & \cellcolor{red!15}+.07 & \cellcolor{red!15}+.06 & \cellcolor{red!15}+.08 & & \cellcolor{gray!10}.00 & \cellcolor{green!8}$-$.01 & \cellcolor{green!8}$-$.02 & \cellcolor{green!8}$-$.02 & \cellcolor{green!8}$-$.05 \\
Japan & \cellcolor{red!8}+.02 & \cellcolor{green!8}$-$.01 & \cellcolor{red!8}+.03 & \cellcolor{red!8}+.03 & \cellcolor{red!15}+.08 & & \cellcolor{green!8}$-$.01 & \cellcolor{green!15}$-$.09 & \cellcolor{green!8}$-$.01 & \cellcolor{red!8}+.02 & \cellcolor{green!15}$-$.09 \\
\addlinespace[2pt]
\multicolumn{12}{l}{\textit{Emerging Europe}} \\
\textit{Group avg.} & \cellcolor{red!8}+.04 & \cellcolor{green!8}$-$.05 & \cellcolor{red!8}+.02 & \cellcolor{green!8}$-$.01 & \cellcolor{gray!10}.00 & & \cellcolor{gray!10}.00 & \cellcolor{gray!10}.00 & \cellcolor{green!8}$-$.01 & \cellcolor{gray!10}.00 & \cellcolor{gray!10}.00 \\
Poland & \cellcolor{red!8}+.02 & \cellcolor{red!8}+.02 & \cellcolor{red!8}+.02 & \cellcolor{green!8}$-$.01 & \cellcolor{red!8}+.05 & & \cellcolor{gray!10}.00 & \cellcolor{green!8}$-$.01 & \cellcolor{green!8}$-$.01 & \cellcolor{green!8}$-$.01 & \cellcolor{green!8}$-$.03 \\
Czech Rep. & \cellcolor{red!8}+.03 & \cellcolor{green!15}$-$.06 & \cellcolor{red!15}+.05 & \cellcolor{red!8}+.03 & \cellcolor{red!8}+.05 & & \cellcolor{red!8}+.01 & \cellcolor{green!8}$-$.01 & \cellcolor{red!8}+.01 & \cellcolor{green!8}$-$.03 & \cellcolor{green!8}$-$.02 \\
Hungary & \cellcolor{red!8}+.04 & \cellcolor{green!15}$-$.09 & \cellcolor{red!8}+.01 & \cellcolor{green!8}$-$.02 & \cellcolor{green!15}$-$.06 & & \cellcolor{green!8}$-$.02 & \cellcolor{red!8}+.02 & \cellcolor{green!8}$-$.01 & \cellcolor{red!15}+.07 & \cellcolor{red!15}+.05 \\
Turkey & \cellcolor{red!15}+.05 & \cellcolor{green!15}$-$.05 & \cellcolor{gray!10}.00 & \cellcolor{green!8}$-$.04 & \cellcolor{green!8}$-$.04 & & \cellcolor{gray!10}.00 & \cellcolor{red!8}+.02 & \cellcolor{gray!10}.00 & \cellcolor{green!8}$-$.02 & \cellcolor{green!8}$-$.01 \\
\addlinespace[2pt]
\multicolumn{12}{l}{\textit{Asia-Pacific}} \\
\textit{Group avg.} & \cellcolor{red!8}+.03 & \cellcolor{green!15}$-$.05 & \cellcolor{red!8}+.04 & \cellcolor{red!8}+.03 & \cellcolor{red!8}+.04 & & \cellcolor{green!8}$-$.01 & \cellcolor{green!8}$-$.03 & \cellcolor{green!8}$-$.01 & \cellcolor{red!8}+.04 & \cellcolor{green!8}$-$.01 \\
China & \cellcolor{red!8}+.03 & \cellcolor{green!8}$-$.03 & \cellcolor{red!15}+.05 & \cellcolor{red!15}+.07 & \cellcolor{red!15}+.12 & & \cellcolor{gray!10}.00 & \cellcolor{green!8}$-$.05 & \cellcolor{green!8}$-$.01 & \cellcolor{green!8}$-$.01 & \cellcolor{green!15}$-$.06 \\
India & \cellcolor{red!8}+.03 & \cellcolor{green!15}$-$.09 & \cellcolor{gray!10}.00 & \cellcolor{red!8}+.01 & \cellcolor{green!8}$-$.04 & & \cellcolor{gray!10}.00 & \cellcolor{green!15}$-$.09 & \cellcolor{green!8}$-$.02 & \cellcolor{red!8}+.01 & \cellcolor{green!15}$-$.10 \\
Indonesia & \cellcolor{red!8}+.03 & \cellcolor{green!15}$-$.06 & \cellcolor{red!8}+.03 & \cellcolor{red!8}+.03 & \cellcolor{red!8}+.03 & & \cellcolor{red!8}+.01 & \cellcolor{green!8}$-$.02 & \cellcolor{gray!10}.00 & \cellcolor{red!8}+.02 & \cellcolor{red!8}+.01 \\
Malaysia & \cellcolor{gray!10}.00 & \cellcolor{green!8}$-$.01 & \cellcolor{red!8}+.04 & \cellcolor{red!8}+.01 & \cellcolor{red!8}+.04 & & \cellcolor{green!8}$-$.03 & \cellcolor{green!8}$-$.03 & \cellcolor{green!8}$-$.02 & \cellcolor{red!15}+.07 & \cellcolor{gray!10}.00 \\
Philippines & \cellcolor{red!8}+.03 & \cellcolor{green!15}$-$.08 & \cellcolor{red!8}+.04 & \cellcolor{red!8}+.01 & \cellcolor{red!8}+.01 & & \cellcolor{green!8}$-$.02 & \cellcolor{red!8}+.02 & \cellcolor{gray!10}.00 & \cellcolor{gray!10}.00 & \cellcolor{red!8}+.01 \\
Thailand & \cellcolor{red!8}+.02 & \cellcolor{green!8}$-$.03 & \cellcolor{red!15}+.05 & \cellcolor{red!8}+.04 & \cellcolor{red!15}+.09 & & \cellcolor{green!8}$-$.03 & \cellcolor{red!8}+.05 & \cellcolor{red!8}+.01 & \cellcolor{red!15}+.14 & \cellcolor{red!15}+.18 \\
Vietnam & \cellcolor{red!8}+.04 & \cellcolor{green!15}$-$.07 & \cellcolor{red!8}+.03 & \cellcolor{red!8}+.03 & \cellcolor{red!8}+.02 & & \cellcolor{green!8}$-$.01 & \cellcolor{green!15}$-$.11 & \cellcolor{green!8}$-$.02 & \cellcolor{red!8}+.03 & \cellcolor{green!15}$-$.12 \\
\addlinespace[2pt]
\multicolumn{12}{l}{\textit{Latin America}} \\
\textit{Group avg.} & \cellcolor{red!8}+.03 & \cellcolor{green!15}$-$.07 & \cellcolor{red!8}+.04 & \cellcolor{red!8}+.02 & \cellcolor{red!8}+.01 & & \cellcolor{gray!10}.00 & \cellcolor{gray!10}.00 & \cellcolor{red!8}+.01 & \cellcolor{green!8}$-$.02 & \cellcolor{gray!10}.00 \\
Argentina & \cellcolor{red!8}+.03 & \cellcolor{green!15}$-$.05 & \cellcolor{red!8}+.05 & \cellcolor{green!8}$-$.04 & \cellcolor{green!8}$-$.02 & & \cellcolor{gray!10}.00 & \cellcolor{red!8}+.02 & \cellcolor{green!8}$-$.01 & \cellcolor{gray!10}.00 & \cellcolor{red!8}+.02 \\
Brazil & \cellcolor{red!8}+.05 & \cellcolor{green!15}$-$.10 & \cellcolor{red!15}+.06 & \cellcolor{red!15}+.07 & \cellcolor{red!15}+.09 & & \cellcolor{red!8}+.02 & \cellcolor{green!8}$-$.01 & \cellcolor{gray!10}.00 & \cellcolor{green!15}$-$.08 & \cellcolor{green!15}$-$.06 \\
Chile & \cellcolor{red!8}+.03 & \cellcolor{green!15}$-$.07 & \cellcolor{red!8}+.04 & \cellcolor{red!8}+.03 & \cellcolor{red!8}+.03 & & \cellcolor{green!8}$-$.03 & \cellcolor{gray!10}.00 & \cellcolor{red!8}+.01 & \cellcolor{green!8}$-$.01 & \cellcolor{green!8}$-$.02 \\
Colombia & \cellcolor{red!8}+.02 & \cellcolor{green!15}$-$.07 & \cellcolor{red!8}+.01 & \cellcolor{red!8}+.03 & \cellcolor{green!8}$-$.02 & & \cellcolor{red!8}+.01 & \cellcolor{green!8}$-$.01 & \cellcolor{red!8}+.02 & \cellcolor{green!8}$-$.03 & \cellcolor{gray!10}.00 \\
Peru & \cellcolor{red!8}+.04 & \cellcolor{green!15}$-$.08 & \cellcolor{red!8}+.04 & \cellcolor{green!8}$-$.01 & \cellcolor{gray!10}.00 & & \cellcolor{green!8}$-$.01 & \cellcolor{gray!10}.00 & \cellcolor{red!8}+.02 & \cellcolor{gray!10}.00 & \cellcolor{red!8}+.01 \\
Uruguay & \cellcolor{red!8}+.01 & \cellcolor{green!15}$-$.06 & \cellcolor{red!8}+.03 & \cellcolor{red!8}+.01 & \cellcolor{green!8}$-$.01 & & \cellcolor{green!8}$-$.01 & \cellcolor{red!8}+.02 & \cellcolor{red!8}+.02 & \cellcolor{red!8}+.01 & \cellcolor{red!8}+.04 \\
\addlinespace[2pt]
\multicolumn{12}{l}{\textit{Middle East \& North Africa}} \\
\textit{Group avg.} & \cellcolor{red!8}+.03 & \cellcolor{green!15}$-$.08 & \cellcolor{red!8}+.03 & \cellcolor{red!8}+.01 & \cellcolor{green!8}$-$.02 & & \cellcolor{green!8}$-$.01 & \cellcolor{gray!10}.00 & \cellcolor{red!8}+.01 & \cellcolor{red!8}+.01 & \cellcolor{red!8}+.01 \\
Egypt & \cellcolor{red!8}+.02 & \cellcolor{green!15}$-$.08 & \cellcolor{gray!10}.00 & \cellcolor{red!8}+.01 & \cellcolor{green!15}$-$.05 & & \cellcolor{gray!10}.00 & \cellcolor{green!8}$-$.05 & \cellcolor{red!8}+.01 & \cellcolor{green!8}$-$.05 & \cellcolor{green!15}$-$.08 \\
Israel & \cellcolor{red!8}+.04 & \cellcolor{green!15}$-$.08 & \cellcolor{red!8}+.02 & \cellcolor{green!15}$-$.06 & \cellcolor{green!15}$-$.08 & & \cellcolor{green!8}$-$.01 & \cellcolor{red!15}+.06 & \cellcolor{gray!10}.00 & \cellcolor{red!8}+.04 & \cellcolor{red!15}+.10 \\
Qatar & \cellcolor{red!8}+.03 & \cellcolor{green!15}$-$.12 & \cellcolor{red!15}+.05 & \cellcolor{red!8}+.04 & \cellcolor{gray!10}.00 & & \cellcolor{green!8}$-$.02 & \cellcolor{green!8}$-$.02 & \cellcolor{green!8}$-$.01 & \cellcolor{gray!10}.00 & \cellcolor{green!8}$-$.04 \\
Saudi Arabia & \cellcolor{red!8}+.03 & \cellcolor{green!8}$-$.05 & \cellcolor{red!8}+.05 & \cellcolor{red!8}+.04 & \cellcolor{red!15}+.07 & & \cellcolor{green!8}$-$.02 & \cellcolor{red!8}+.01 & \cellcolor{red!8}+.04 & \cellcolor{red!8}+.03 & \cellcolor{red!15}+.06 \\
\addlinespace[2pt]
\bottomrule
\end{tabular}
\vspace{4pt}\newline{\scriptsize
\colorbox{red!15}{\phantom{X}} $\Delta>+0.05$\quad
\colorbox{red!8}{\phantom{X}} $+0.005<\Delta\leq+0.05$\quad
\colorbox{gray!10}{\phantom{X}} $|\Delta|\leq0.005$\quad
\colorbox{green!8}{\phantom{X}} $-0.05\leq\Delta<-0.005$\quad
\colorbox{green!15}{\phantom{X}} $\Delta<-0.05$.
Liberation Day USMCA group avg.\ not reported separately; see individual countries.
$\varphi^{\rm LOC}$ includes Region cross-SHAP.}
\end{table}
\clearpage

% APPENDIX D — TWO TABLES ON ONE PAGE
% Key moves: \scriptsize tables, compressed vertical spacing,
% shorter notes, negative \vspace between tables
% % ============================================================
% APPENDIX D — TWO TABLES ON ONE PAGE (enlarged, no underlines)
% Key fix: \parbox replaces \makecell (which causes underlines)
% Enlarged: \footnotesize instead of \scriptsize
% ============================================================

\section{Appendix. Local Projection Robustness}
\label{app:LP_FD}

\vspace{-8pt}

\begin{table}[htbp]
\centering
\caption{Average Innovation LP : Estimated Channel Impulse Responses at $h = 5$, $30$, and $60$ Days}
\label{tab:appendix_ar5}
\scriptsize\setlength{\tabcolsep}{5pt}
\begin{threeparttable}
\begin{tabular}{l l r c r c r c}
\toprule
\multirow{2}{*}{Event} & \multirow{2}{*}{Channel} & \multicolumn{2}{c}{$h=5$} & \multicolumn{2}{c}{$h=30$} & \multicolumn{2}{c}{$h=60$} \\
\cmidrule(lr){3-4}\cmidrule(lr){5-6}\cmidrule(lr){7-8}
 &  & $\hat{\beta}$ & Sig. & $\hat{\beta}$ & Sig. & $\hat{\beta}$ & Sig. \\
\midrule
\multirow{4}{*}{\makecell[l]{\textbf{Russia--Ukraine} \\ {\footnotesize GPR, Feb.\ 2022} \\ {\footnotesize \textit{Geopolitical}}}} & $\varphi^{dir}$ & 0.0130 & -- & 0.0050 & -- & 0.0034 & -- \\
 & $\varphi^{GFC}$ & -0.0039 & -- & -0.0007 & -- & -0.0002 & -- \\
 & $\varphi^{UNC}$ & -0.0012 & -- & -0.0003 & -- & -0.0002 & -- \\
 & $\varphi^{LOC}$ & -0.0077 & -- & -0.0037 & -- & -0.0023 & -- \\
\midrule
\multirow{4}{*}{\makecell[l]{\textbf{Hamas--Israel} \\ {\footnotesize GPR, Oct.\ 2023} \\ {\footnotesize \textit{Geopolitical}}}} & $\varphi^{dir}$ & 0.0130 & -- & 0.0050 & -- & 0.0034 & -- \\
 & $\varphi^{GFC}$ & -0.0039 & -- & -0.0007 & -- & -0.0002 & -- \\
 & $\varphi^{UNC}$ & -0.0012 & -- & -0.0003 & -- & -0.0002 & -- \\
 & $\varphi^{LOC}$ & -0.0077 & -- & -0.0037 & -- & -0.0023 & -- \\
\midrule
\multirow{4}{*}{\makecell[l]{\textbf{U.S. Election} \\ {\footnotesize EPU, Nov.\ 2024} \\ {\footnotesize \textit{Geoeconomic}}}} & $\varphi^{dir}$ & 0.0034 & -- & 0.0017 & -- & 0.0014 & -- \\
 & $\varphi^{GFC}$ & 0.0032 & -- & 0.0017 & -- & 0.0010 & -- \\
 & $\varphi^{UNC}$ & 0.0008 & -- & 0.0006 & -- & 0.0004 & -- \\
 & $\varphi^{LOC}$ & -0.0006 & -- & 0.0003 & -- & 0.0003 & -- \\
\midrule
\multirow{4}{*}{\makecell[l]{\textbf{``Liberation Day''} \\ {\footnotesize TPU, Apr.\ 2025} \\ {\footnotesize \textit{Geoeconomic}}}} & $\varphi^{dir}$ & -0.0020 & -- & -0.0008 & -- & -0.0005 & -- \\
 & $\varphi^{GFC}$ & -0.0016 & -- & -0.0010 & -- & -0.0000 & -- \\
 & $\varphi^{UNC}$ & 0.0015 & -- & 0.0009 & -- & 0.0005 & -- \\
 & $\varphi^{LOC}$ & -0.0009 & -- & 0.0000 & -- & 0.0002 & -- \\
\bottomrule
\end{tabular}
\begin{tablenotes}[flushleft]
\scriptsize
\item \textit{Notes:} Estimated impulse responses from panel local projections \citep{Jorda2005} using AR(5) residual innovations as structural shocks. Standard errors follow \citet{DriscollKraay1998} with bandwidth $= \max(20,\, h)$, with country and time fixed effects and 5 lags of $\Delta y$. Contemporaneous controls: global VIX (moving average) and sovereign yield spread (moving average). 
\end{tablenotes}
\end{threeparttable}
\end{table}

\begin{table}[htbp]
\centering
\caption{Narrative LP: Estimated Channel Impulse Responses at $h = 5$, $30$, and $60$ Days}
\label{tab:appendix_nar}
\scriptsize\setlength{\tabcolsep}{5pt}
\begin{threeparttable}
\begin{tabular}{l l r c r c r c}
\toprule
\multirow{2}{*}{Event} & \multirow{2}{*}{Channel} & \multicolumn{2}{c}{$h=5$} & \multicolumn{2}{c}{$h=30$} & \multicolumn{2}{c}{$h=60$} \\
\cmidrule(lr){3-4}\cmidrule(lr){5-6}\cmidrule(lr){7-8}
 &  & $\hat{\beta}$ & Sig. & $\hat{\beta}$ & Sig. & $\hat{\beta}$ & Sig. \\
\midrule
\multirow{4}{*}{\makecell[l]{\textbf{Russia--Ukraine} \\ {\footnotesize GPR, Feb.\ 2022} \\ {\footnotesize \textit{Geopolitical}}}} & $\varphi^{dir}$ & 0.1031 & *** & 0.1213 & *** & 0.0948 & *** \\
 & $\varphi^{GFC}$ & -0.0782 & *** & -0.1035 & *** & -0.0808 & *** \\
 & $\varphi^{UNC}$ & -0.0110 & *** & -0.0096 & * & -0.0013 & -- \\
 & $\varphi^{LOC}$ & -0.0728 & *** & -0.0234 & -- & -0.0540 & *** \\
\midrule
\multirow{4}{*}{\makecell[l]{\textbf{Hamas--Israel} \\ {\footnotesize GPR, Oct.\ 2023} \\ {\footnotesize \textit{Geopolitical}}}} & $\varphi^{dir}$ & 0.0710 & *** & 0.0722 & *** & 0.0448 & *** \\
 & $\varphi^{GFC}$ & 0.0323 & ** & 0.0003 & -- & 0.0263 & + \\
 & $\varphi^{UNC}$ & -0.0164 & *** & 0.0060 & -- & 0.0014 & -- \\
 & $\varphi^{LOC}$ & -0.0275 & + & -0.0179 & -- & -0.0258 & + \\
\midrule
\multirow{4}{*}{\makecell[l]{\textbf{U.S. Election} \\ {\footnotesize EPU, Nov.\ 2024} \\ {\footnotesize \textit{Geoeconomic}}}} & $\varphi^{dir}$ & 0.0073 & *** & 0.0132 & *** & 0.0084 & *** \\
 & $\varphi^{GFC}$ & -0.0351 & *** & -0.0463 & *** & -0.0331 & *** \\
 & $\varphi^{UNC}$ & 0.0180 & *** & 0.0214 & *** & 0.0316 & *** \\
 & $\varphi^{LOC}$ & 0.0174 & *** & 0.0151 & *** & 0.0337 & *** \\
\midrule
\multirow{4}{*}{\makecell[l]{\textbf{``Liberation Day''} \\ {\footnotesize TPU, Apr.\ 2025} \\ {\footnotesize \textit{Geoeconomic}}}} & $\varphi^{dir}$ & -0.0300 & *** & -0.0090 & *** & -- & -- \\
 & $\varphi^{GFC}$ & -0.1386 & *** & -0.0194 & * & -- & -- \\
 & $\varphi^{UNC}$ & -0.0206 & *** & 0.0264 & *** & -- & -- \\
 & $\varphi^{LOC}$ & -0.0529 & *** & 0.0319 & ** & -- & -- \\
\bottomrule
\end{tabular}
\begin{tablenotes}[flushleft]
\scriptsize
\item \textit{Notes:} Estimated impulse responses from narrative local projections using dated event dummies within a $\pm 3$-day window around each crisis episode. Standard errors are clustered by country; country fixed effects are included. Contemporaneous controls: global VIX (moving average) and sovereign yield spread (moving average).
\item $^{***}p<0.01$, $^{**}p<0.05$, $^{*}p<0.10$, $^{+}p<0.32$ (68\% CI); $--$ = not significant at the 10\% level.
\end{tablenotes}
\end{threeparttable}
\end{table}

\subsection{Placebo Falsification: Channel-Level Evidence}
\label{app:placebo}

\begin{figure}[ht!]
    \centering
    \caption{\textbf{Placebo Test (GPR): Russia--Ukraine, Feb.\ 24, 2022}}
   
    \includegraphics[width=0.7\textwidth]{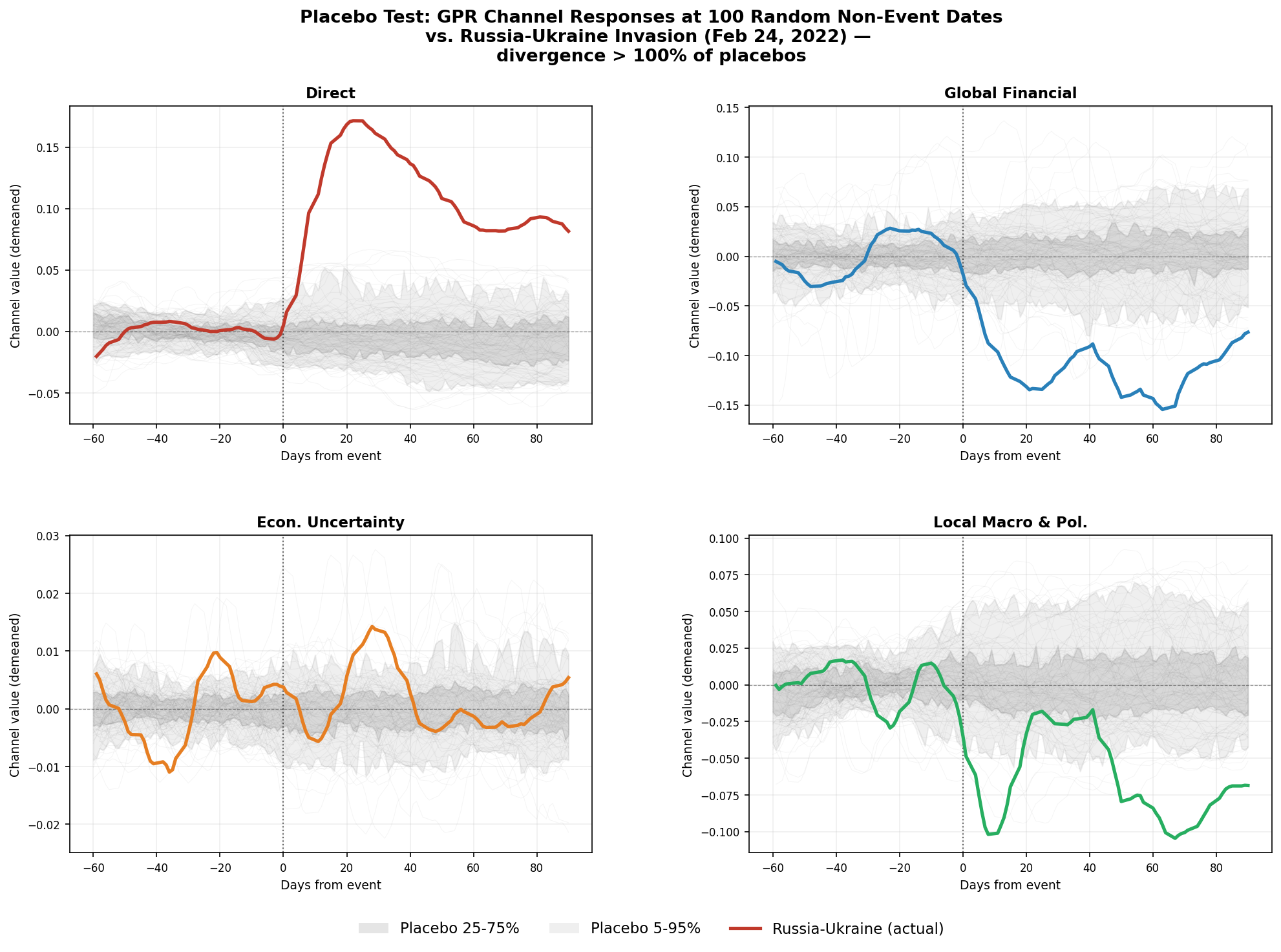}
    \label{fig:placebo_RU}
    \begin{flushleft}
       \scriptsize
        Notes: Channel-level cross-country mean responses vs.\ 25--75\% (dark) and 5--95\% (light) placebo bands from 100 random non-event dates. The Direct channel exits the 95th percentile band upward; the GFC channel exits downward. The Uncertainty channel remains within the placebo range, consistent with geopolitical---rather than geoeconomic---transmission. 7-day trailing MA; values demeaned over the pre-event window.
    \end{flushleft}
\end{figure}

\begin{figure}[ht!]
    \centering
    \caption{\textbf{Placebo Test (GPR): Hamas--Israel, Oct.\ 7, 2023 --- Divergence $>$ 86\% of Placebos}}
   
    \includegraphics[width=0.7\textwidth]{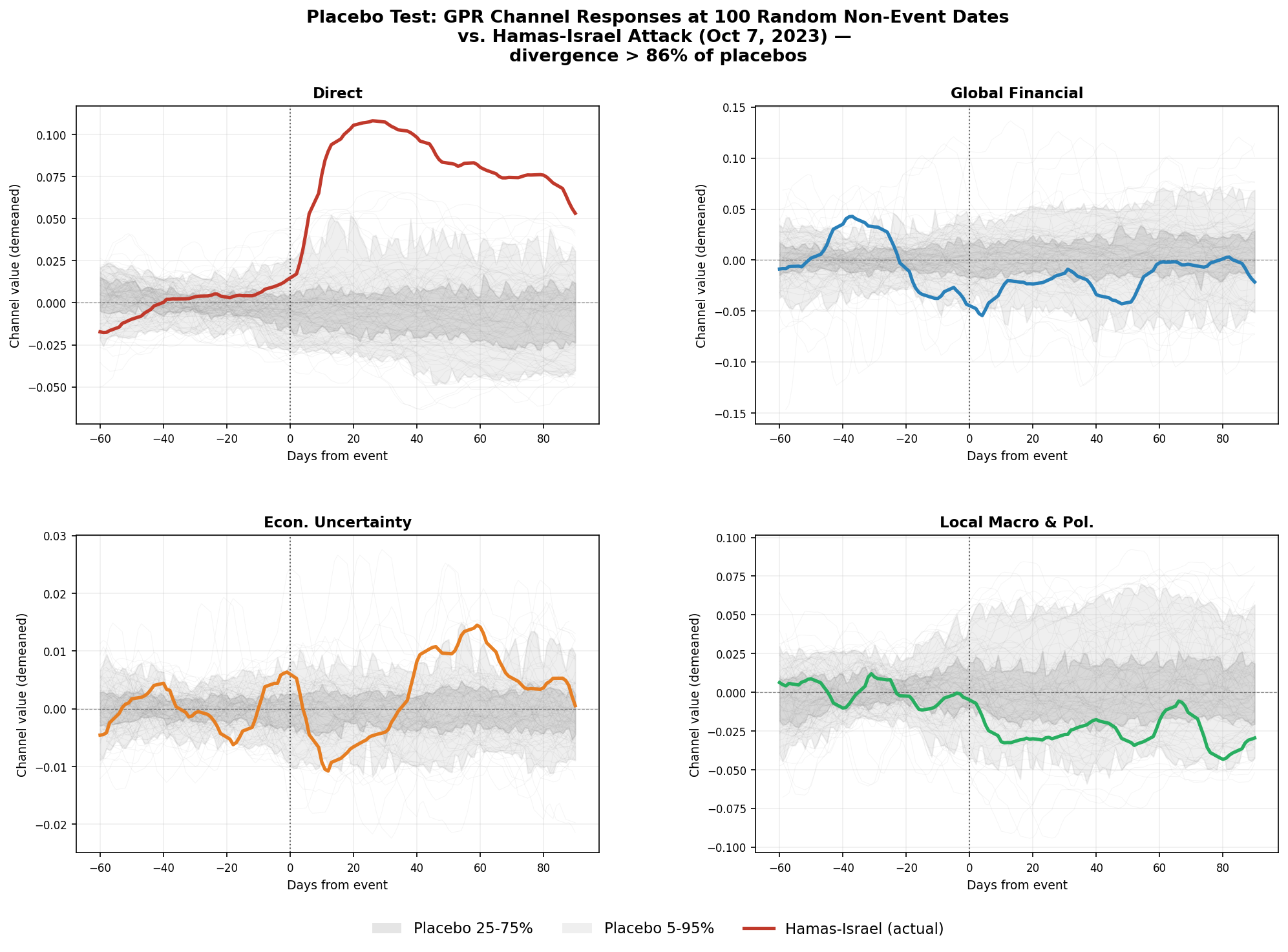}
    \label{fig:placebo_HI}
    \begin{flushleft}
       \scriptsize
        Notes: The Direct channel exits the 95th percentile band upward immediately after day~0, replicating the geopolitical signature observed for Russia--Ukraine. The GFC channel remains largely within the placebo range, consistent with the more regionally contained nature of this conflict. The Uncertainty and Local channels stay inside the bands. 25--75\% (dark) and 5--95\% (light) placebo bands from 100 random non-event dates. 7-day trailing MA; values demeaned over the pre-event window.
    \end{flushleft}
\end{figure}

\begin{figure}[ht!]
    \centering
    \caption{\textbf{Placebo Test (EPU): U.S.\ Presidential Election, Nov.\ 5, 2024 --- Divergence $>$ 83\% of Placebos}}
   
    \includegraphics[width=0.7\textwidth]{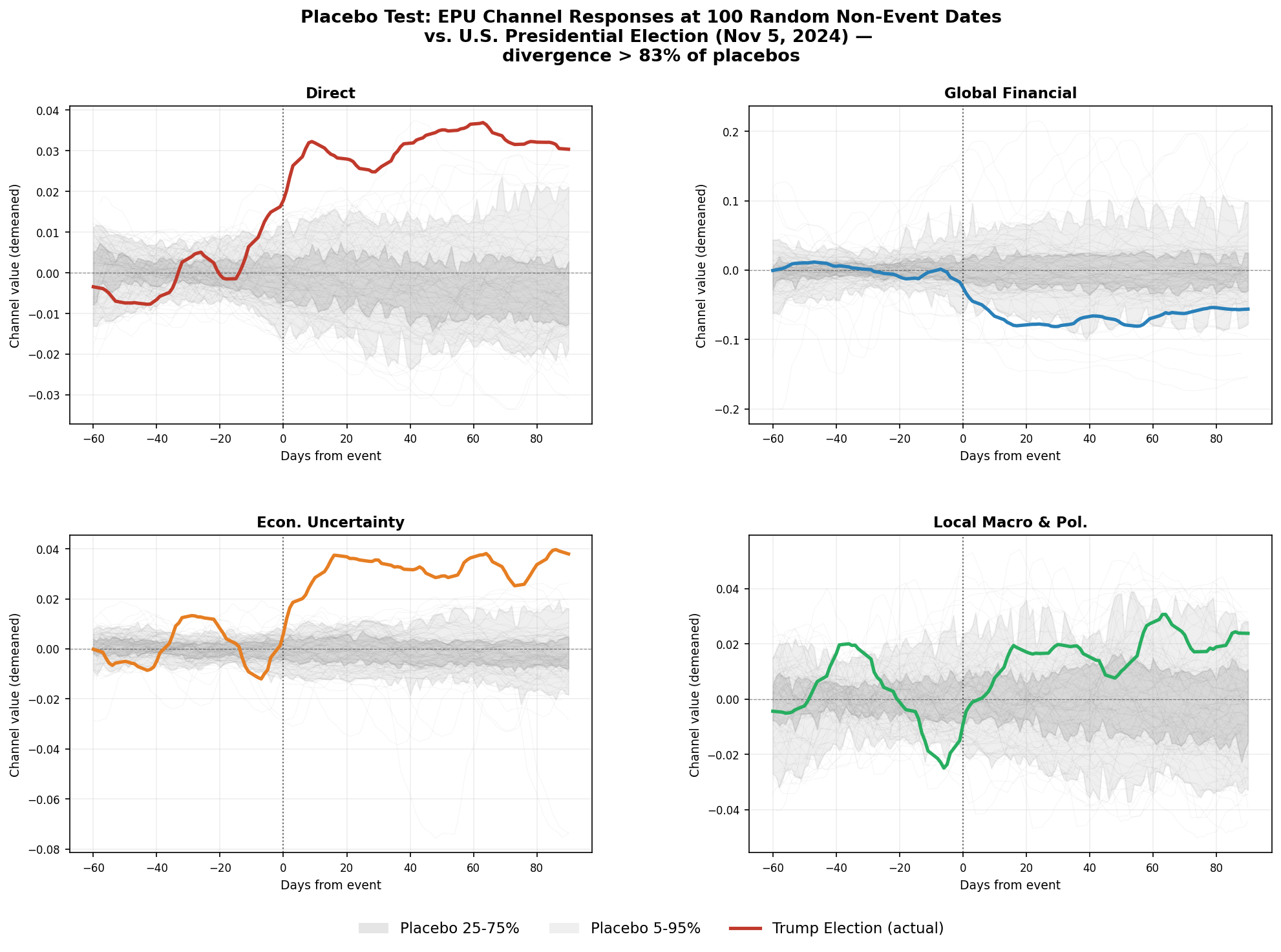}
    \label{fig:placebo_EL}
    \begin{flushleft}
       \scriptsize
        Notes: The Uncertainty channel exits the envelope upward after day~0, consistent with EPU-driven repricing. The GFC channel drops below the band. The Direct channel rises modestly above the envelope, reflecting tariff-related repricing accompanying the election. 25--75\% (dark) and 5--95\% (light) placebo bands from 100 random non-event dates. 7-day trailing MA; values demeaned over the pre-event window.
    \end{flushleft}
\end{figure}

\begin{figure}[ht!]
    \centering
    \caption{\textbf{Placebo Test (TPU): Liberation Day Tariff Shock, Apr.\ 2, 2025 --- Divergence $>$ 99\% of Placebos}}
   
    \includegraphics[width=0.7\textwidth]{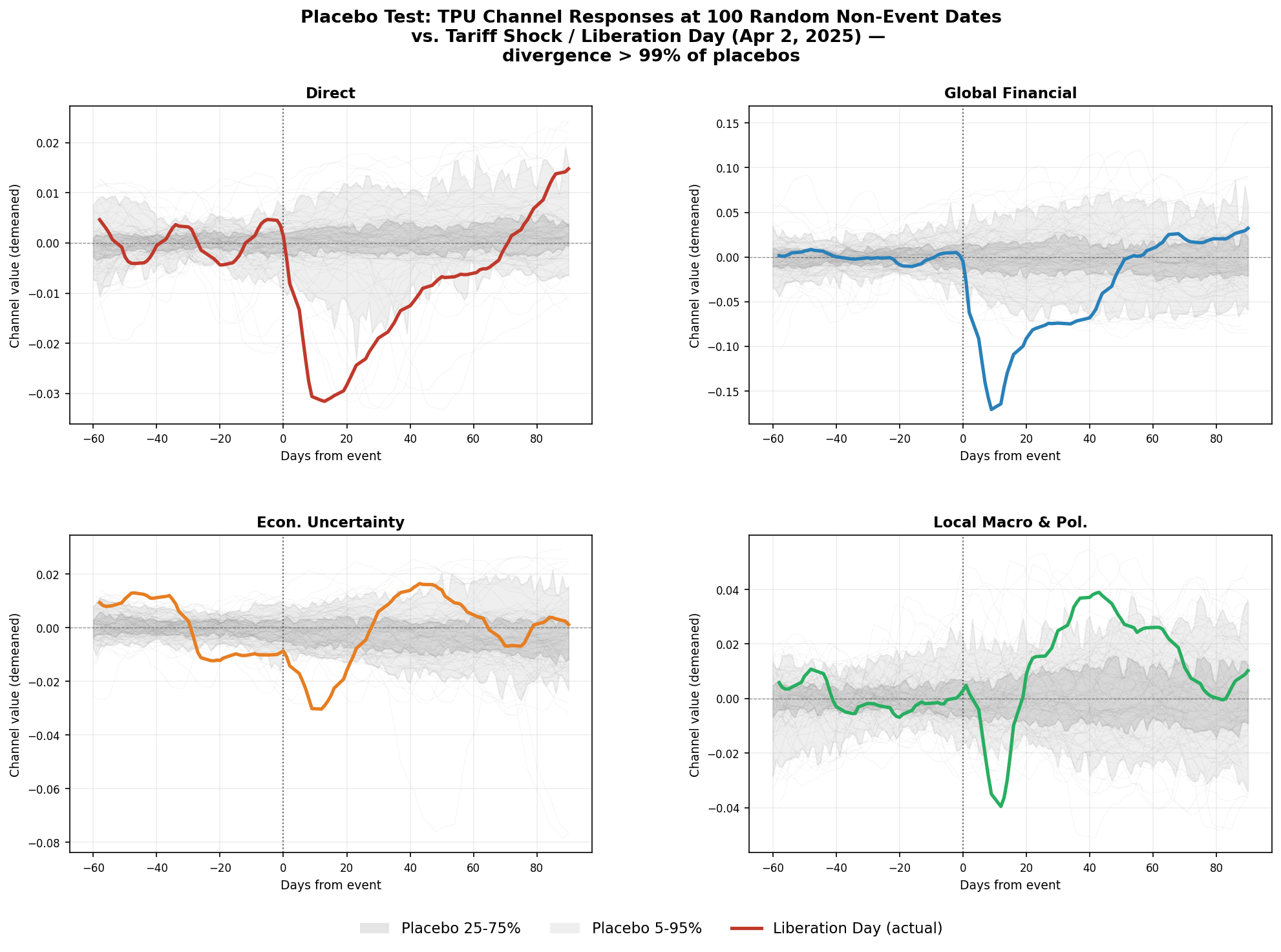}
    \label{fig:placebo_LD}
    \begin{flushleft}
       \scriptsize
        Notes: The GFC channel collapses below the envelope immediately after day~0---the largest single-channel departure across all four episodes. The Direct channel moves \emph{negative}, consistent with the taxonomy's prediction that tariff shocks bypass conflict-proximity repricing. The Local channel exits upward, reflecting heterogeneous tariff exposure. 25--75\% (dark) and 5--95\% (light) placebo bands from 100 random non-event dates. 7-day trailing MA; values demeaned over the pre-event window.
    \end{flushleft}
\end{figure}

Figures~\ref{fig:placebo_RU}--\ref{fig:placebo_LD} present the channel-level
placebo tests for all four episodes, using the driver-specific Shapley--Taylor
decomposition (GPR for geopolitical events; EPU and TPU for geoeconomic
events). Each panel plots a single channel's cross-country mean response
against the placebo envelope constructed from 100 randomly drawn non-event
dates.

\clearpage

\subsection{Block Bootstrap Validation of Channel Decomposition}
\label{app:bootstrap}
\begin{figure}[ht!]
    \centering
    \caption{\textbf{Block Bootstrap Validation of Narrative LP Impulse Responses
         ($B=500$, 4 narrative events)}}

    \includegraphics[width=0.7\textwidth]{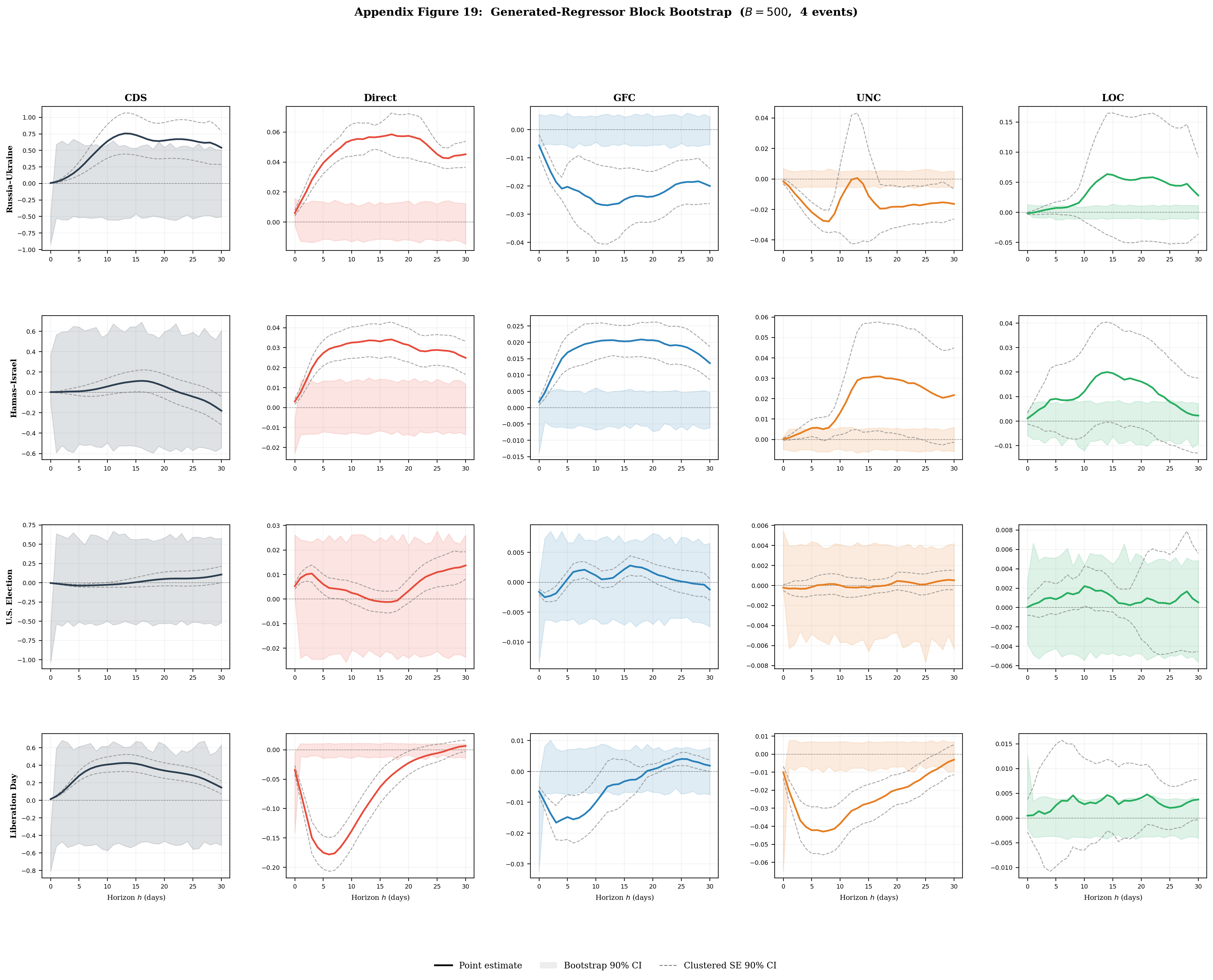}
    \label{fig:bootstrap_CI}
    \begin{flushleft}
       \scriptsize
        Notes: Each panel plots the narrative LP impulse-response function
for the episode indicated in the row label on the variable indicated
in the column header: the raw CDS spread and the four Shapley--Taylor
transmission channels ($\varphi^{\mathrm{DIR}}$,
$\varphi^{\mathrm{GFC}}$, $\varphi^{\mathrm{UNC}}$,
$\varphi^{\mathrm{LOC}}$) over horizons $h = 0, \ldots, 30$ days.
Solid coloured lines: point estimates from the baseline narrative
local projection (Table~\ref{tab:causal_LP}). Dashed grey lines:
clustered 90\% confidence intervals, robust to
heteroskedasticity and cross-sectional
dependence. Coloured shaded bands: 90\% confidence intervals from
$B = 500$ block-bootstrap replications that jointly re-estimate the
Multilayer Random Forest, recompute the Shapley--Taylor decomposition
via \textsc{TreeSHAP}, and re-run the panel local projection. For
the raw CDS spread (first column), bootstrap and clustered intervals are
comparable in width, confirming that the two methods agree when the
dependent variable is directly observed. For the four channel series,
bootstrap bands are substantially tighter because the channel
decomposition is highly stable across re-estimations of the random
forest trained on approximately 75,000 observations; the clustered
estimator is wider because it additionally corrects for cross-sectional
dependence across 42 sovereigns---a source of uncertainty the block
bootstrap does not fully replicate. No channel that is significant
under clustered inference loses significance under the bootstrap. Clustered
standard errors are used as the conservative baseline throughout.
    \end{flushleft}
\end{figure}

% ============================================================
%  APPENDIX H: Complementary Validation via Narrative Sign-Restricted SVARs
% ============================================================

\section{Appendix F. SVAR Specification and Additional Results}
\label{app:svar}

This appendix provides the technical specification underlying the sign-restricted SVAR results summarised in Section~\ref{sec:svar_validation}, together with additional robustness exercises.  The key results---the 8/8 scorecard on raw observables (Figure~\ref{fig:svar_raw_irfs}) and the cross-validation summary (Table~\ref{tab:svar_summary})---are reported in the main text.

% ------------------------------------------
\subsection*{F.1 \; Specification}
\label{app:svar_spec}

\paragraph{Observable mapping.}
For each country $i$, we estimate a five-variable SVAR on observables
that map directly into the four transmission channels without passing
through the machine-learning model:
\begin{equation}
\mathbf{Y}_{i,t}
=
\bigl(\,
\text{CDS}_{i,t},\;
\text{GPR}_{i,t},\;
\text{GFC}_{t}^{\text{obs}},\;
\text{UNC}_{i,t}^{\text{obs}},\;
\text{LOC}_{i,t}^{\text{obs}}
\,\bigr)' .
\label{eq:svar_y_revised}
\end{equation}
Here, $\text{GPR}_{i,t}$ is the country-level geopolitical risk index
that proxies the Direct channel. $\text{GFC}_{t}^{\text{obs}}$ is a
two-variable global financial-conditions composite based on the
standardised VIX and U.S. Treasury yield, normalised so that lower
values correspond to the flight-to-safety configuration that offsets
the Direct channel in the main text. $\text{UNC}_{i,t}^{\text{obs}}$
combines economic and trade policy uncertainty measures and proxies the
Uncertainty channel. $\text{LOC}_{i,t}^{\text{obs}}$ combines domestic
economic, institutional, and political indicators and proxies the
Local channel.\footnote{Results are robust to alternative normalisations
of the global financial composite, to using the underlying components
individually, and to principal-component extraction within each
composite.}

All series are transformed into 28-day moving averages for
comparability with the main text. Country-specific variables are
standardised within country; global variables are standardised once
over the full sample and then stacked across countries in the panel.

\paragraph{Reduced form.}
For each country $i$, we estimate a reduced-form VAR($p$) with
$p=5$, matching the lag structure used in the main text:
\begin{equation}
\mathbf{Y}_{i,t}
=
\boldsymbol{\mu}_i
+
\sum_{\ell=1}^{5}
\mathbf{A}_{i,\ell}\mathbf{Y}_{i,t-\ell}
+
\mathbf{u}_{i,t},
\qquad
\mathbf{u}_{i,t}\sim\mathcal{N}(\mathbf{0},\boldsymbol{\Sigma}_i).
\label{eq:svar_rf_revised}
\end{equation}

\paragraph{Structural representation.}
Let
$\mathbf{u}_{i,t}=\mathbf{B}_i\boldsymbol{\varepsilon}_{i,t}$,
where $\boldsymbol{\varepsilon}_{i,t}$ contains five orthogonal
structural shocks and $\mathbf{B}_i\mathbf{B}_i'=\boldsymbol{\Sigma}_i$.
We label two shocks of interest: a geopolitical shock
$\varepsilon^G_{i,t}$ and a geoeconomic shock $\varepsilon^E_{i,t}$.
The remaining three orthogonal shocks are left unlabeled. Identification
combines sign restrictions, relative-magnitude restrictions, and
narrative restrictions derived from the semistructural benchmarks in
Section~\ref{sec:benchmarks}.

% ------------------------------------------
\subsection*{F.2 \; Restriction Design}
\label{app:svar_restrictions}

\paragraph{Layer 1: Sign restrictions.}
The semistructural framework implies sign predictions for the initial
response of each observable system. We impose sign restrictions at
$h=0$ and $h=5$ (approximately one trading week in the daily system):

\begin{table}[h!]
\centering
\small
\caption{SVAR Sign Restrictions Implied by the Semistructural Framework}
\label{tab:svar_signs}
\begin{tabular}{l cc cc}
\toprule
& \multicolumn{2}{c}{Geopolitical shock ($\varepsilon^G$)}
& \multicolumn{2}{c}{Geoeconomic shock ($\varepsilon^E$)} \\
\cmidrule(lr){2-3}\cmidrule(lr){4-5}
Variable & $h{=}0$ & $h{=}5$ & $h{=}0$ & $h{=}5$ \\
\midrule
CDS$_{i,t}$                      & $+$ & $+$ & -- & -- \\[2pt]
GPR$_{i,t}$ (Direct)             & $+$ & $+$ &
$\approx 0$ & $\approx 0$ \\[2pt]
GFC$_{t}^{\text{obs}}$           & $-$ & $-$ & $-$ & $-$ \\[2pt]
UNC$_{i,t}^{\text{obs}}$         & --  & --  & $+$ & $+$ \\[2pt]
LOC$_{i,t}^{\text{obs}}$         & --  & --  & $+$ & $+$ \\
\bottomrule
\end{tabular}

\vspace{4pt}
{\footnotesize\textit{Notes:} ``$+$'' and ``$-$'' denote sign restrictions;
``--'' denotes unrestricted. ``$\approx 0$'' for the geoeconomic
response of GPR is implemented as a relative-magnitude restriction:
$|\Theta^{h}_{\text{GPR},\varepsilon^E}| 
0.5\cdot|\Theta^{h}_{\text{GPR},\varepsilon^G}|$ for
$h\in\{0,5\}$. CDS is left unrestricted under $\varepsilon^E$
because the identifying content of the taxonomy lies in the relative
channel configuration rather than in a universally positive immediate
spread response at all horizons.}
\end{table}

\paragraph{Layer 2: Relative-magnitude restrictions.}
The key empirical content of the taxonomy is not only the sign of each
response but also the ranking of the dominant channel. We therefore
impose two relative-magnitude inequalities over the initial week:
\begin{align}
\text{Geopolitical shock:}\quad & \max_{h\in\{0,\dots,5\}} |\Theta^{h}_{\text{GFC},\varepsilon^G}| < \max_{h\in\{0,\dots,5\}} |\Theta^{h}_{\text{GPR},\varepsilon^G}|
\label{eq:dominance_geo_revised} \\[4pt]
\text{Geoeconomic shock:}\quad & \max_{h\in\{0,\dots,5\}} |\Theta^{h}_{\text{GFC},\varepsilon^E}| > \max_{h\in\{0,\dots,5\}} |\Theta^{h}_{\text{GPR},\varepsilon^E}|.
\label{eq:dominance_geoeco_revised}
\end{align}
The first inequality encodes the scissors logic: under a geopolitical
shock, the GFC response moves in the opposite direction but remains
secondary to the direct repricing channel. The second inequality
captures the geoeconomic benchmark: the financial-conditions channel
dominates the muted direct channel.

\paragraph{Layer 3: Narrative restrictions.}
We impose narrative restrictions on the four crisis dates used in the
main text:
\begin{align*}
t_{E1} &= \text{24 Feb 2022} \qquad \text{(Russia--Ukraine)} \\
t_{E2} &= \text{7 Oct 2023}  \qquad \text{(Hamas--Israel)} \\
t_{E3} &= \text{5 Nov 2024}  \qquad \text{(U.S.\ election)} \\
t_{E4} &= \text{2 Apr 2025}  \qquad \text{(Liberation Day / tariff shock)} .
\end{align*}

Following \citet{AntolinDiazRubioRamirez2018}, the restrictions are
imposed country by country at these common calendar dates. We use two
types of narrative conditions.

\textit{Type 1 (shock sign).} The target shock must be positive on the
corresponding narrative date:
\begin{align}
\varepsilon^G_{i,t_{E1}} > 0, \qquad
\varepsilon^G_{i,t_{E2}} > 0, \qquad
\varepsilon^E_{i,t_{E3}} > 0, \qquad
\varepsilon^E_{i,t_{E4}} > 0 .
\label{eq:narrative_signs_revised}
\end{align}

\textit{Type 2 (shock ranking among labeled shocks).} On each narrative
date, the target shock must be larger in absolute value than the other
labeled shock:
\begin{align}
|\varepsilon^G_{i,t_{E1}}| > |\varepsilon^E_{i,t_{E1}}|,
\qquad
|\varepsilon^G_{i,t_{E2}}| > |\varepsilon^E_{i,t_{E2}}|,
\label{eq:narrative_rank_g_revised}
\\[2pt]
|\varepsilon^E_{i,t_{E3}}| > |\varepsilon^G_{i,t_{E3}}|,
\qquad
|\varepsilon^E_{i,t_{E4}}| > |\varepsilon^G_{i,t_{E4}}|.
\label{eq:narrative_rank_e_revised}
\end{align}

These narrative conditions anchor the sign-restricted system on the
same dated episodes used in the local-projection validation, while
leaving the remaining three structural shocks unlabeled.

% ------------------------------------------
\subsection*{F.3 \; Estimation Algorithm and Aggregation}
\label{app:svar_estimation}

We follow the Bayesian accept--reject procedure in
\citet{AntolinDiazRubioRamirez2018}. For each country, we:
(i) draw reduced-form VAR parameters from the Normal--inverse-Wishart
posterior;
(ii) draw an orthogonal rotation matrix $\mathbf{Q}$ uniformly from
the Haar measure on $O(5)$;
(iii) compute structural impulse responses at horizons
$h=0,\ldots,60$; and
(iv) retain the draw if all sign, relative-magnitude, and narrative
restrictions are satisfied.

The baseline raw-observable system yields 639 accepted draws. For each
accepted draw and each horizon $h$, we compute a mean-group response as
the unweighted cross-country average of the corresponding country-level
IRF. Pointwise posterior medians and 68\%/90\% credible bands are then
constructed from the distribution of these mean-group IRFs across
accepted draws. This aggregation preserves country-level heterogeneity
in the reduced form while reporting a single cross-country response
profile.

% ------------------------------------------
% ------------------------------------------
\subsection*{F.4 \; Internal Coherence Check: SVAR on Shapley--Taylor Channel Series}
\label{app:svar_coherence}

\begin{figure}[ht!]
    \centering
    \caption{\textbf{Narrative Sign-Restricted SVAR: Structural Impulse Responses on Shapley--Taylor Channel Series}}
    \label{fig:svar_channel_irfs}
    \includegraphics[width=0.95\textwidth]{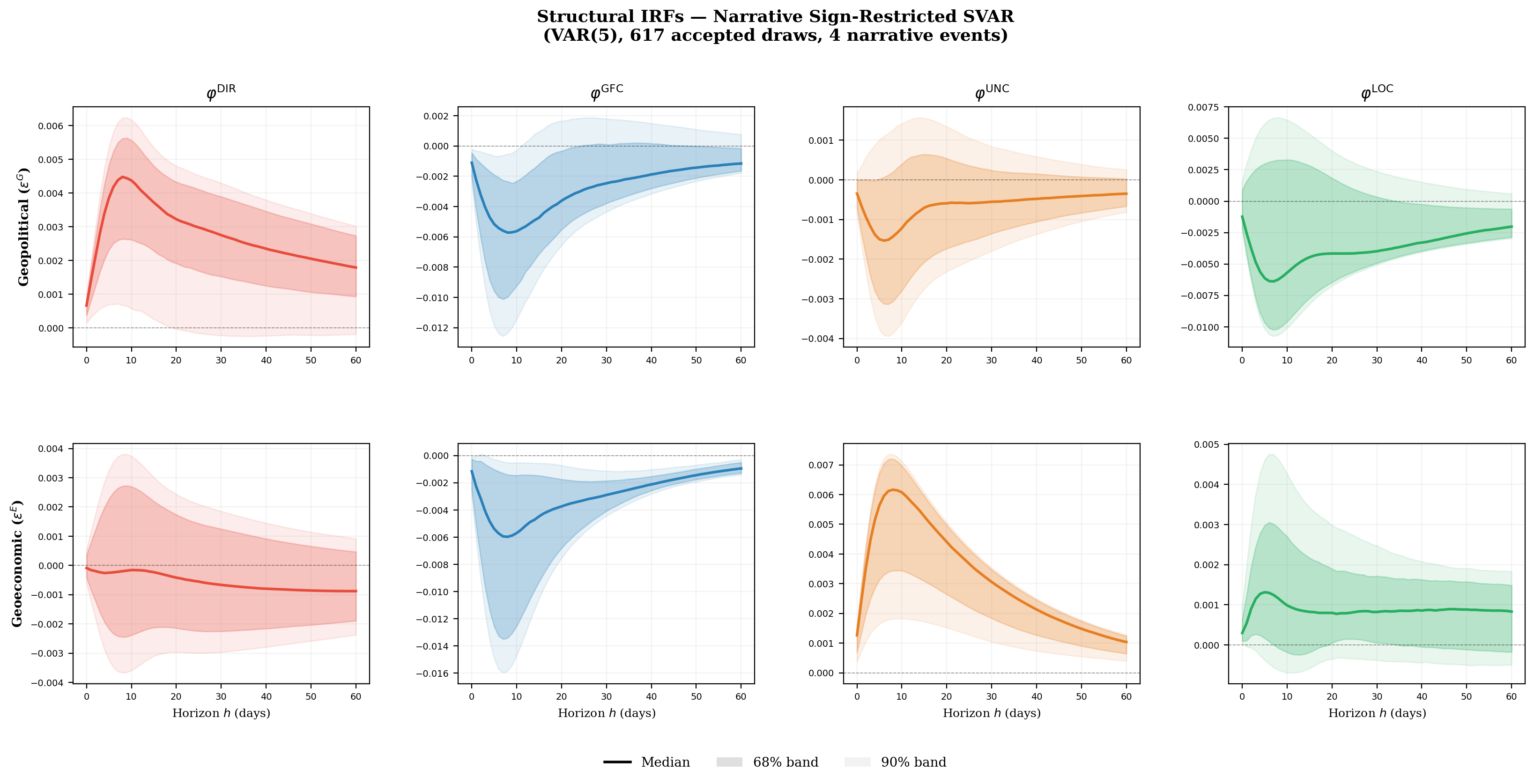}
    \begin{flushleft}
       \scriptsize
        Notes: Mean-group structural impulse responses from country-level VAR(5) models estimated on the four Shapley--Taylor channel series ($\varphi^{\textsc{dir}}$, $\varphi^{\textsc{gfc}}$, $\varphi^{\textsc{unc}}$, $\varphi^{\textsc{loc}}$), identified with sign restrictions (Table~\ref{tab:svar_signs}), relative-magnitude restrictions, and narrative restrictions on four crisis dates following Antol\'in-D\'iaz and Rubio-Ram\'irez (2018). Top row: geopolitical shock~($\varepsilon^G$); bottom row: geoeconomic shock~($\varepsilon^E$). Solid lines: posterior median across 617 accepted draws. Dark shaded bands: 68\% pointwise credible intervals; light bands: 90\%. The geopolitical shock raises $\varphi^{\textsc{dir}}$ while $\varphi^{\textsc{gfc}}$ falls, reproducing the scissors pattern. The geoeconomic shock produces a muted and eventually negative $\varphi^{\textsc{dir}}$, with the GFC, Uncertainty, and Local channels carrying transmission. All eight sign and dominance predictions are confirmed at the posterior median. Because this system uses Shapley--Taylor values as inputs, it serves as an internal coherence check rather than an independent validation; Figure~\ref{fig:svar_raw_irfs} provides the independent test on raw observables.
    \end{flushleft}
\end{figure}

As an internal coherence check, we estimate the same sign-restricted
system on the Shapley--Taylor channel series
$(\varphi^{\text{DIR}},\varphi^{\text{GFC}},
\varphi^{\text{UNC}},\varphi^{\text{LOC}})$ together with CDS.
This exercise is not independent of the machine-learning decomposition,
but it tests whether the channel series admit a structural
representation aligned with the semistructural benchmarks. The channel
SVAR yields the same qualitative 8/8 scorecard, with somewhat tighter
credible bands and 617 accepted draws (Figure~\ref{fig:svar_channel_irfs}).

% ------------------------------------------
\subsection*{F.5 \; Full-Sample LP with SVAR-Identified Shocks}
\label{app:svar_fullsample}

We use the identified structural shocks
$\hat{\varepsilon}^{G}_{i,t}$ and $\hat{\varepsilon}^{E}_{i,t}$ as
regressors in full-sample panel local projections with country fixed
effects and Driscoll--Kraay standard errors. In these regressions, the
responses of both the raw observables and CDS are economically small
and statistically indistinguishable from zero across horizons. This
finding is stable across three control sets:
(A) lags only,
(B) lags plus global financial controls, and
(C) the full contemporaneous control set shown in
Figure~\ref{fig:svar_robustness}.

\begin{figure}[ht!]
    \centering
    \caption{\textbf{Full-Sample Local Projections with 
    SVAR-Identified Shocks: Attenuation Toward Zero}}
    \label{fig:lp_svar_shocks}
    \includegraphics[width=0.95\textwidth]{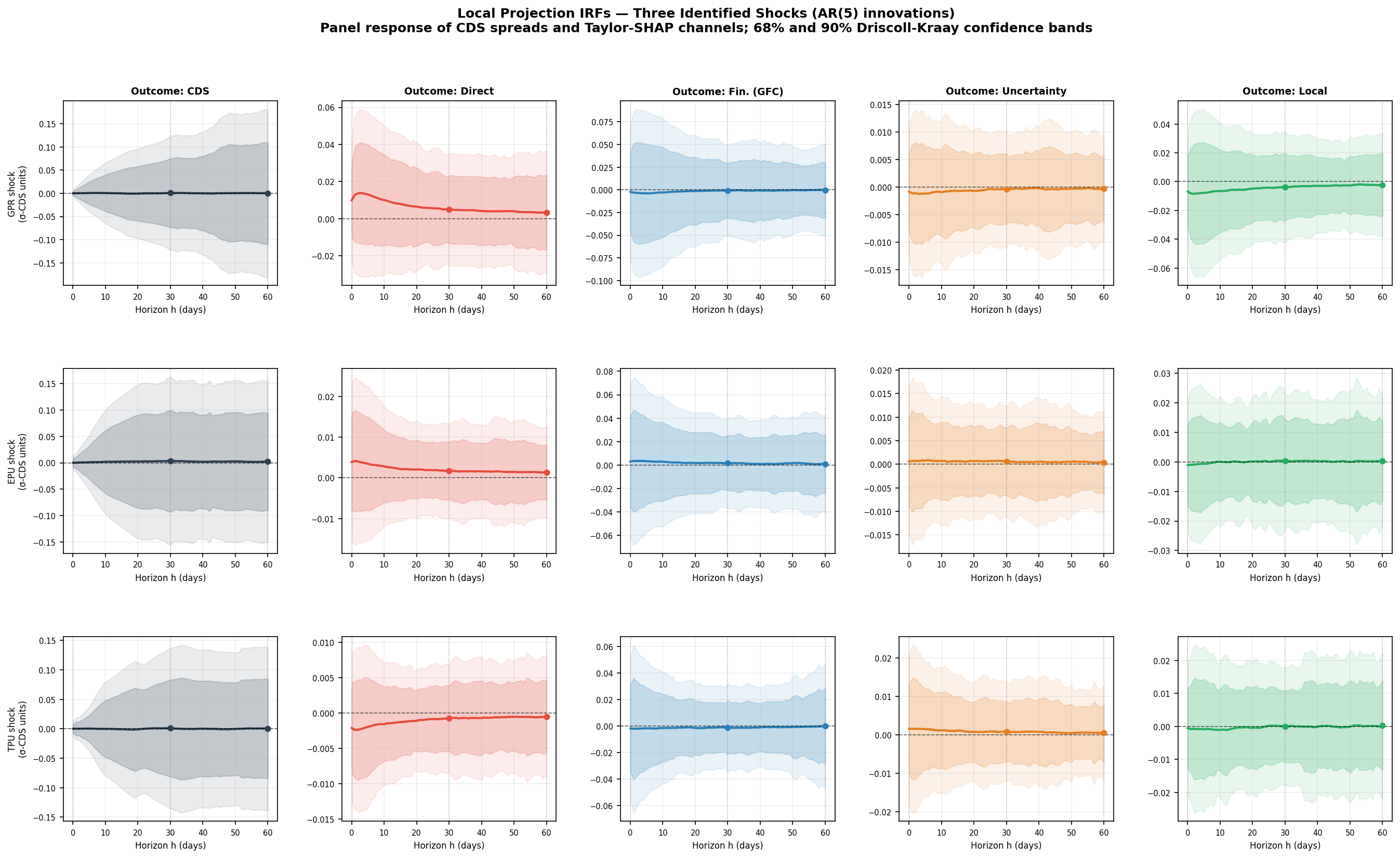}
    \begin{flushleft}
       \scriptsize
        Notes: Panel local-projection impulse responses of sovereign 
        CDS spreads and the four Shapley--Taylor channels 
        ($\varphi^{\textsc{dir}}$, $\varphi^{\textsc{gfc}}$, 
        $\varphi^{\textsc{unc}}$, $\varphi^{\textsc{loc}}$) to 
        SVAR-identified structural shocks used as regressors over 
        the full sample. Rows: GPR shock (top), EPU shock (middle), 
        TPU shock (bottom). Columns: raw CDS spread and the four 
        channel series. Structural shocks are extracted from 
        country-level sign-restricted SVARs 
        ($\hat{\varepsilon}_{i,t} = B_i^{-1} u_{i,t}$; see 
        Appendix~\ref{app:svar} for identification details); 
        standard errors follow \citet{DriscollKraay1998} with 
        bandwidth $\max(20,h)$, country and time fixed effects, 
        and 5 lags. Dark shaded bands: 68\% confidence intervals; 
        light bands: 90\%. Dots at $h = 30$ mark the horizon 
        reported in Table~\ref{tab:causal_LP}. All responses are 
        economically small and statistically indistinguishable from 
        zero at conventional levels, consistent with state-dependent 
        transmission concentrated in discrete crisis episodes rather 
        than a pervasive linear relation in the full sample. This 
        attenuation mirrors the innovation-based null 
        documented in Section~\ref{sec:LP_AR5} and contrasts with 
        the large and significant responses recovered by the
        narrative design (Figure~\ref{fig:channels_taxonomy}).
    \end{flushleft}
\end{figure}

\begin{figure}[ht!]
    \centering
    \caption{\textbf{Robustness of Full-Sample LP Attenuation Across Three Control Sets}}
    \label{fig:svar_robustness}
    \includegraphics[width=0.95\textwidth]{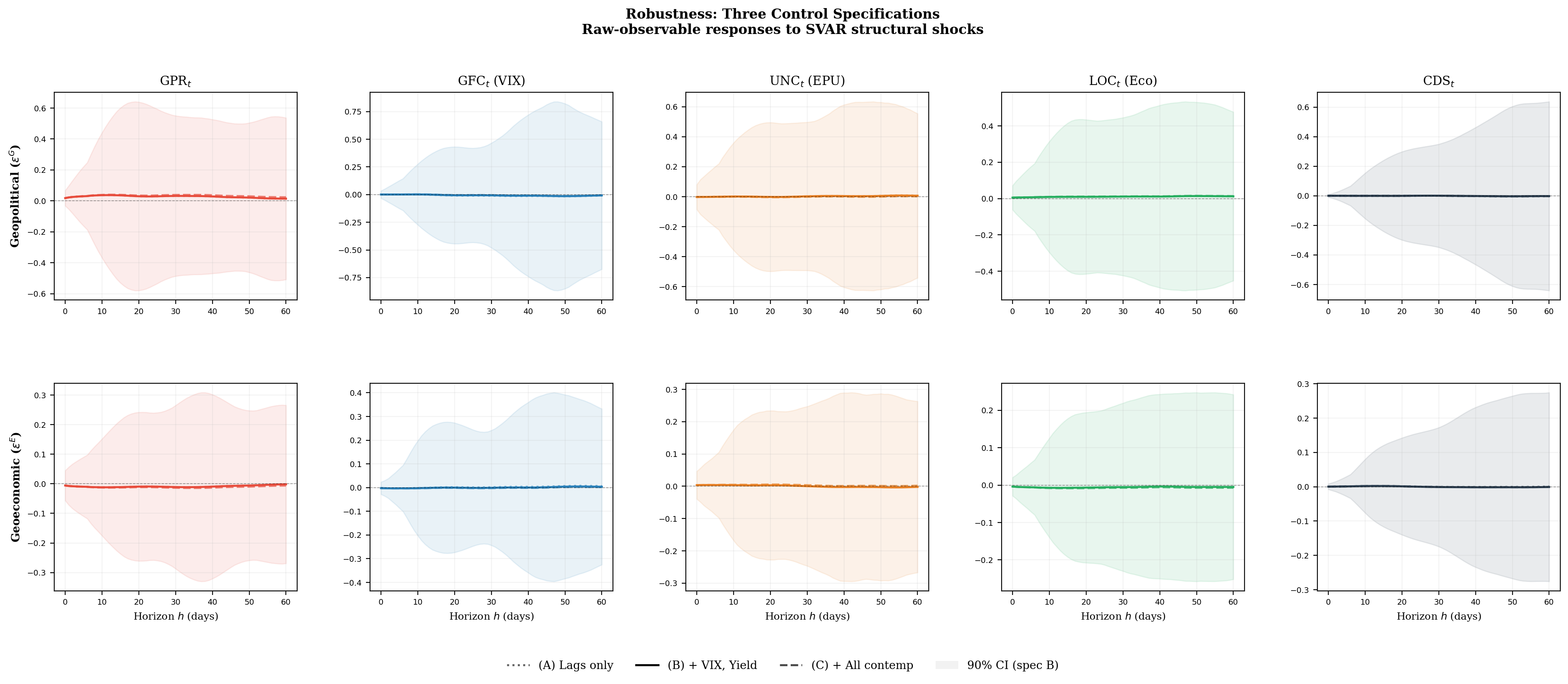}
    \begin{flushleft}
       \scriptsize
        Notes: Raw-observable impulse responses to SVAR-identified structural shocks under three control specifications: (A)~lags only, (B)~lags plus VIX and U.S.\ Treasury yield, and (C)~the full contemporaneous control set. Top row: geopolitical shock~($\varepsilon^G$); bottom row: geoeconomic shock~($\varepsilon^E$). Columns: GPR (Direct proxy), GFC composite, Uncertainty composite, Local composite, and sovereign CDS spread. Shaded bands: 90\% Driscoll--Kraay confidence intervals (specification~B). All responses remain economically small and statistically indistinguishable from zero across control sets, confirming that the full-sample attenuation documented in Figure~\ref{fig:lp_svar_shocks} is not driven by the choice of controls.
    \end{flushleft}
\end{figure}

Taken together, Figures~\ref{fig:svar_raw_irfs}
and~\ref{fig:lp_svar_shocks} mirror the main-text contrast between sharp
responses in narrative episode designs and weak average responses in
full-sample innovation designs. The SVAR evidence therefore
corroborates, rather than mechanically restates, the state-dependent
interpretation of sovereign-risk transmission.

% ------------------------------------------
\subsection*{F.6 \; Interpretation: Why the Full-Sample LP Attenuates}
\label{app:svar_interpretation}

The identified SVAR shocks are linear combinations of reduced-form
innovations with time-invariant weights:
\begin{equation}
\hat{\boldsymbol{\varepsilon}}_{i,t}
=
\mathbf{B}_i^{-1}\mathbf{u}_{i,t}.
\label{eq:linear_shocks_revised}
\end{equation}
Because these weights do not vary with the state of the financial system, a full-sample LP averages over states in which transmission is active and states in which it is dormant, attenuating the average response toward zero. The SVAR structural IRFs are less affected because the narrative restrictions concentrate identification power on tail events, and the VAR propagates shocks through a cross-equation dynamic system \citep{PlagborgMollerWolf2021}. The combined pattern---sharp structural IRFs in the narrative SVAR but weak average responses in full-sample LPs---is consistent with state-dependent transmission.

\end{document}